\documentclass{article}

\PassOptionsToPackage{numbers, compress}{natbib}

\usepackage[final]{neurips_2024}

\usepackage[utf8]{inputenc} 
\usepackage[T1]{fontenc}    
\usepackage{hyperref}       
\usepackage{url}            
\usepackage{booktabs}       
\usepackage{amsfonts}       
\usepackage{nicefrac}       
\usepackage{microtype}      
\usepackage[svgnames]{xcolor}         
\usepackage{comment}
\usepackage{amsmath, pifont}
\usepackage{dsfont}

\usepackage{graphicx}  
\usepackage{colortbl}  
\usepackage{multirow}  
\usepackage{multicol}  
\usepackage{wrapfig}  
\usepackage{tablefootnote} 

\usepackage{caption}  
\usepackage{subcaption}  

\arrayrulecolor{Teal} 
\setcitestyle{square} 

\newcommand{\NDVI}{\text{NDVI}}  
\newcommand{\BT}{\text{BT}}  
\newcommand{\VCI}{\text{VCI}}  
\newcommand{\TCI}{\text{TCI}}  
\newcommand{\VHI}{\text{VHI}}  

\newcommand{\X}{\textbf{X}}  
\newcommand{\Z}{\textbf{Z}}  
\newcommand{\B}{\textbf{B}}  
\newcommand{\E}{\textbf{E}}  
\newcommand{\Snow}{\textbf{S}}
\newcommand{\N}{\textbf{N}}  
\newcommand{\Q}{\textbf{Q}}  

\newcommand{\cmark}{\ding{51}} 
\newcommand{\xmark}{\ding{55}} 

\DeclareMathOperator*{\argmin}{arg\,min}
\DeclareMathOperator{\sign}{sign}
\def\1{\mathbbm{1}}

\newcommand{\RR}{\mathbb{R}}
\newcommand{\ZZ}{\mathbb{Z}}

\newcommand{\QQ}{\mathbb{Q}}
\newcommand{\EE}{\mathbb{E}}

\newcommand{\bepsilon}{{\boldsymbol{\epsilon}}}

\newcommand{\gray}[1]{\textcolor{gray}{#1}}
\newcommand{\teal}[1]{\textcolor{teal}{#1}}

\newcommand{\olive}[1]{\textcolor{olive}{#1}}

\newcommand{\legendbox}[1]{%
  \textcolor{#1}{\rule{\fontcharht\font`X}{\fontcharht\font`X}}%
}

\newcommand{\draft}{false}
\newcommand{\vis}{false}

\def\logo{\makebox[12pt][l]{\raisebox{-0.9ex}{\includegraphics[draft=false, height=12pt]{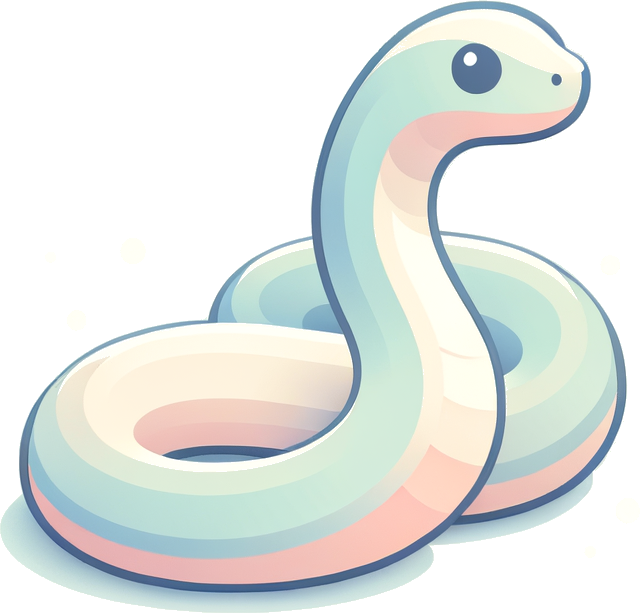}}}\hspace{6pt}}

\title{Identifying Spatio-Temporal Drivers of Extreme Events}

\author{%
  Mohamad Hakam Shams Eddin \qquad Juergen Gall\\\\
  Institute of Computer Science, University of Bonn\\
  Lamarr Institute for Machine Learning and Artificial Intelligence \\
  \texttt{\{shams, gall\}@iai.uni-bonn.de} \\
}

\begin{document}

\maketitle

\begin{abstract}
The spatio-temporal relations of impacts of extreme events and their drivers in climate data are not fully understood and there is a need of machine learning approaches to identify such spatio-temporal relations from data. The task, however, is very challenging since there are time delays between extremes and their drivers, and the spatial response of such drivers is inhomogeneous. In this work, we propose a first approach and benchmarks to tackle this challenge. Our approach is trained end-to-end to predict spatio-temporally extremes and spatio-temporally drivers in the physical input variables jointly. By enforcing the network to predict extremes from spatio-temporal binary masks of identified drivers, the network successfully identifies drivers that are correlated with extremes. We evaluate our approach on three newly created synthetic benchmarks, where two of them are based on remote sensing or reanalysis climate data, and on two real-world reanalysis datasets.
The source code and datasets are publicly available at the project page \teal{\url{https://hakamshams.github.io/IDE}}.

\end{abstract}
\section{Introduction}
\label{sec:1}
A frontier research challenge is to understand the affects of global change on the magnitude and probability of extreme weather events \cite{Herring_2022}.
Overall, the evolution of extreme events such as agricultural droughts results from stochastic processes \cite{SILLMANN201765}, conditions at ecosystem scales \cite{EK_2004, Miralles_2019}, and the interaction between the Earth land and atmospheric variables as a part of a complex system of feedbacks \cite{geirinhas2023combined}.
However, the relative impacts of these factors differ depending on the event \cite{SILLMANN201765}.
The time delays between extremes and their drivers vary seasonally \cite{HUANG201545, WANG2022107301, DAI2022127897, CAO2022153270}, and the spatial response of these drivers is inhomogeneous \cite{Miralles_2019}.
A major challenge is therefore to model the spatio-temporal relations between extremes and their drivers during the development of these events \cite{Miralles_2019}.
The overarching goal of this modelling is to improve our understanding of the patterns and impacts of such events. This would improve our ability to project duration and intensity of extreme events and hence assisting in adaptation planning \cite{Stott_2016, Hao_2018}.

\begin{figure}[!t]
  \centering
  \includegraphics[draft=\vis, width=.9\textwidth]{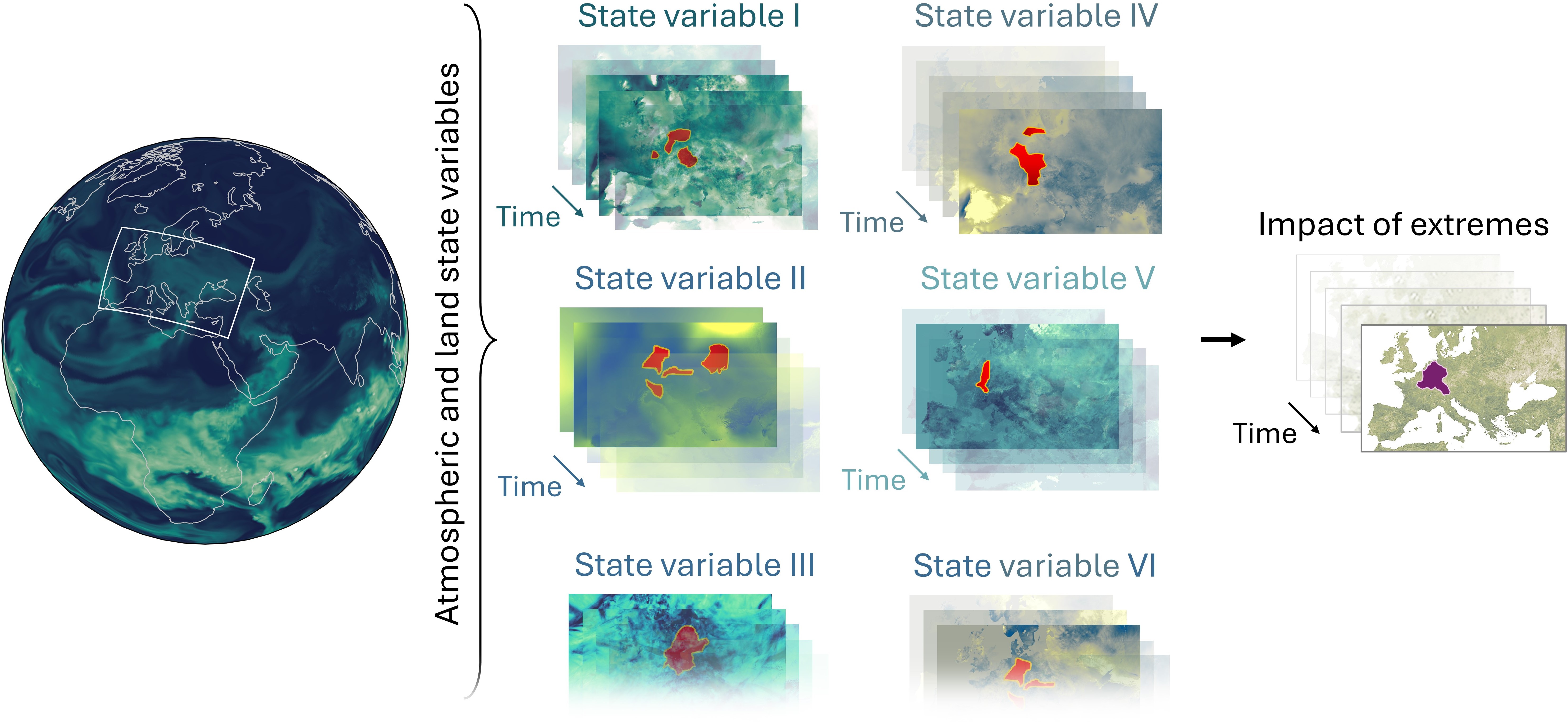}
  \caption{
  Overview of the objective of this work. We are interested in identifying spatio-temporal relations between the measurable impacts of extremes like the vegetation health index \legendbox{Purple} and their drivers \legendbox{red}. As drivers, we focus on anomalies in state variables of the land-atmosphere and hydrological cycle. The task is very challenging since the drivers can occur at a different region than the extreme event and earlier in time.
  }\label{fig:0}
\end{figure}

In this work, we propose an approach that identifies spatio-temporal drivers in multivariate climate data that are correlated with the impact of extreme events. For the extreme events, we focus on agricultural droughts as an example, which can be measured by extremely low values of the vegetation health index (VHI). As drivers for such measurable extreme events, we consider anomalies in atmospheric and hydrological state variables like temperature or soil moisture, as well as land-atmosphere fluxes like evaporation. The task of identifying end-to-end spatio-temporal drivers for measurable impacts of extremes has not been addressed before, and it is very challenging since the drivers can occur earlier in time and at a different location than the measured extreme event as illustrated in Fig.~\ref{fig:0}. 

To address this challenging task, we propose a network that is trained to predict spatio-temporally extremes. Instead of simply predicting the extremes, the network quantizes the spatio-temporal input variables into binary states and predicts the extremes only from the spatio-temporal binary maps for each time series of input variables. In this way, the network is enforced to identify only drivers in the input variables that are spatio-temporally correlated with extreme events. While the network is trained using annotations of impacts of extreme events, which can be derived from remote sensing or reported data, we do not have any annotations of drivers or anomalies in the input variables.

Since drivers of extreme events are not fully understood, we cannot quantitatively measure the accuracy of the identified drivers on real-world data. We therefore propose a framework for generating synthetic data that can be used to assess the performance of our model as well as other baselines quantitatively. We evaluate our approach on three synthetic datasets where two of them are based on remote sensing or reanalysis climate data. Our evaluation shows that our approach outperforms approaches for interpretable forecasting, spatio-temporal anomaly detection, out-of-distribution detection, and multiple instance learning. Furthermore, we conduct empirical studies on two real-world reanalysis climate data.
Our contributions can be summarized as follow:
\begin{itemize}
    \item We introduce the new task of identifying spatio-temporal drivers of extreme
events and three benchmarks for evaluating this highly important task.
    \item We propose a novel approach for identifying spatio-temporal drivers in climate data that are spatio-temporally correlated with the impacts of extreme events.
    \item We further verify our approach on two long-term real-world reanalysis datasets including various physical variables from five biogeographical diverse regions.
\end{itemize}
\section{Related works}
\label{sec:2}

\textbf{Anomalies and extremes detection in climate data.}~
The identification of climatic changes and extreme weather has been a subject of many studies \cite{Das_2009, salcedo2024analysis}.
Typical algorithms for extreme events detection are built upon domain knowledge in setting usually empirical thresholds for the physical variables through sensitivity experiments \cite{HUANG201545, liu2016application}.
Many works applied multivariate and statistical methods to detect extreme events such as droughts \cite{HUANG201545, CAO2022153270, nhess-12-3519-2012, ZHANG2021107028, ZHANG2017141}.
However, individual events are difficult to generalized across multiple events \cite{SILLMANN201765} and predefined indicators become less effective with changing climate \cite{yihdego2019drought}.
Thus, machine and deep learning methods have been proposed as an alternative to classical methods, i.e., for supervised anomaly detection \cite{Dahoui_ecmwf, ECMWF_2021} and for the detection of extremes in climate data \cite{liu2016application, Flach, racah2017extremeweather, Prabhat_2021, Weirich_2023}. 

While methods for forecasting vegetation indices \cite{BARRETT2020111886, EarthNet_2021, Benson_2024_CVPR, focal_tsmp} do not focus on extremes, future impacts of extremes like agricultural droughts can be derived from forecast vegetation indices like the vegetation health index (VHI). For instance, the work \cite{focal_tsmp} uses a climate simulation as input and forecasts the vegetation health index. Since predicting VHI directly is difficult, the approach predicts the normalized difference vegetation index and the brightness temperature instead. Both indices are then normalized and used to estimate VHI. Although we obtain the impacts of extreme events in our study from vegetation indices, our approach is not limited to such extremes. Since we use a binary representation of extremes, our approach can also be applied to other extremes that cannot be derived from satellite products, but that are stored in a binary format in databases.        

While we aim to learn the relations between the impacts of extreme events and their relevant drivers from a data-driven perspective, spatio-temporal relations within the Earth system can also be inferred by causal inference and causal representation learning \cite{runge2019inferring, Schoelkopf_2021, brehmer2022weakly, Tibau_2022, boussard2023towards, runge2023causal, rohekar23a, TARRAGA2024110628}. In contrast to statistical methods \cite{esd-12-151-2021}, data-driven methods do not require a prior hypothesis about drivers for extremes. Instead, they generate hypotheses that can be verified by statistical methods in a second step. We believe that this is an important direction since climate reanalysis provides huge amount of data and it is infeasible to test all combinations. This is also known as a curse of dimensionality in causal discovery problems \cite{runge_2019} and data-driven approaches are therefore needed to generate potential candidates.

\textbf{Anomaly detection algorithms.}~
Since we focus on anomalies in land and meteorological data as drivers, we give an overview of approaches for anomaly detection and discuss their applicability to our task, which has not been previously addressed. 
\textit{One-class:}~
The main stream in one-class anomaly detection is to model the distribution of the normal data during training and consider the deviation from the learned features as anomalies. This includes distance-based \cite{OCSVM, liznerski2021explainable, Yi_2020_ACCV}, patch-based \cite{Yi_2020_ACCV, 9577292, PaDim}, student-teacher \cite{Deng_2022_CVPR, salehi2021multiresolution, bergmann2020uninformed, wang2021student}, and embedding-based approaches \cite{PaDim, Reiss_2021_CVPR, 9493210, 9879738, SimpleNet}.
In general, the alignment between the anomaly type and the assumptions of the methods is the critical factor for their performance \cite{han2022adbench}. One of the limiting factors to apply these methods to climate data is that they assume priori knowledge about what is considered as normal. Furthermore, not all detected anomalies are drivers of an extreme event.  
%
\textit{Reconstruction-based:}~ These methods assume that a trained model to reconstruct normal data will be unsuccessful in reconstructing anomalies, while it will reconstruct normal data well. Despite being widely applied for anomaly detection problems \cite{Ribana_2019, mishra21, lee2022anovit, Ribana_2022, AVID, venkataramanan2020attention, STEALNet, liu2023diversity, yang2023video, 10273635, 9857019, he2024mambaad}, these methods face the same problem as the one-class methods. In addition, many studies showed that anomalies can still be reconstructed by the trained model \cite{UniAD}. 
\textit{Self-supervised learning:}~ These methods rely on the hypothesis that a model trained for a pretext task on normal data will be successful only on similar normal data during inference \cite{9779083, luo2020video, NSA, wang2022video}.
In addition to the above discussed limitations, finding a suitable pretext task for anomaly detection is challenging. For instance, common tasks such as solving a jigsaw puzzle \cite{wang2022video} will fail in homogeneous regions. 
%
\textit{Pseudo-anomaly:}~
The intuition in pseudo-based anomaly detection is to convert the problem of unsupervised learning into a supervised one by synthesizing abnormal data during training \cite{zavrtanik2021draem, collin2021improved, 9578875, pourreza2021g2d, zaheer2020old, thakare2023dyannet, li2023unified, zhang2024realnet}. Since these methods depend partially on the degree to which the proxy anomalies correspond to the unknown true anomalies \cite{Deng_2022_CVPR}, applying these approaches to our task would require some knowledge about the coupling between the variables and extremes.
\textit{Multivariate anomaly detection:}~
Multivariate approaches detect anomalies simultaneously in multiple data streams \cite{xu2022anomaly, lee2021weakly, zhong2024patchad, tuli2022tranad, 10316684}. The main difference to our task is that we aim to detect drivers across multiple data streams that do not necessary occur simultaneously. 
%
\textit{Multiple instance learning:}~
Multiple instance learning (MIL) has been proposed for weakly supervised anomaly detection \cite{adgcn_cvpr19, zaheer2020claws, 9408419, feng2021mist, Park_2023_WACV, zhou2023dual, qi2022weakly, S3R, sapkota2022bayesian}. In MIL-based algorithms, the model is provided with labeled positive and negative bags where each bag includes a set of instances. The model is then trained to classify the instances inside the bags giving only the high level supervision, i.e., label of the bag \cite{jiang2023weakly}. A weakly supervised approach has been also applied for hyperspectral anomaly detection \cite{9444588}. Most algorithms such as \cite{DeepMIL, ARNet, RTFM, MGFN} choose the top-k potentially anomalous snippets within each video. This makes it challenging to apply them since the abnormality ratio varies in real-world applications \cite{zhou2023batchnorm}.
Furthermore, the MIL detector can be biased toward a specific class depending on the context \cite{lv2023unbiased}.

\section{Method}
\label{sec:3}
Our aim is to design a model that is capable of identifying spatio-temporal drivers of extremes in multivariate climate data, i.e., Earth observations or climate reanalysis. In particular, we want to identify anomalies that are spatio-temporally related to extreme events like agricultural droughts. This is different to standard anomaly detection since we are not interested in all anomalies, but in spatio-temporal configurations of variables that potentially cause an extreme event with some time delay and at a potentially different location as illustrated in Fig.~\ref{fig:0}.         
To achieve this, we propose a network that is trained end-to-end on observed extremes where we focus on agricultural droughts. The network classifies the input variables before an extreme event occurs into spatio-temporal drivers without additional annotations besides the annotated extremes. The network then aims to predict future extremes based on the identified drivers. 
It needs to be noted that we are interested in input variables that do not define an extreme event, but we aim to find anomalies in input variables that are correlated with the occurrence of an extreme event. We will thus denote potential drivers as anomalies.         

\begin{figure}[t]
  \centering
  \includegraphics[draft=false, width=.95\textwidth]{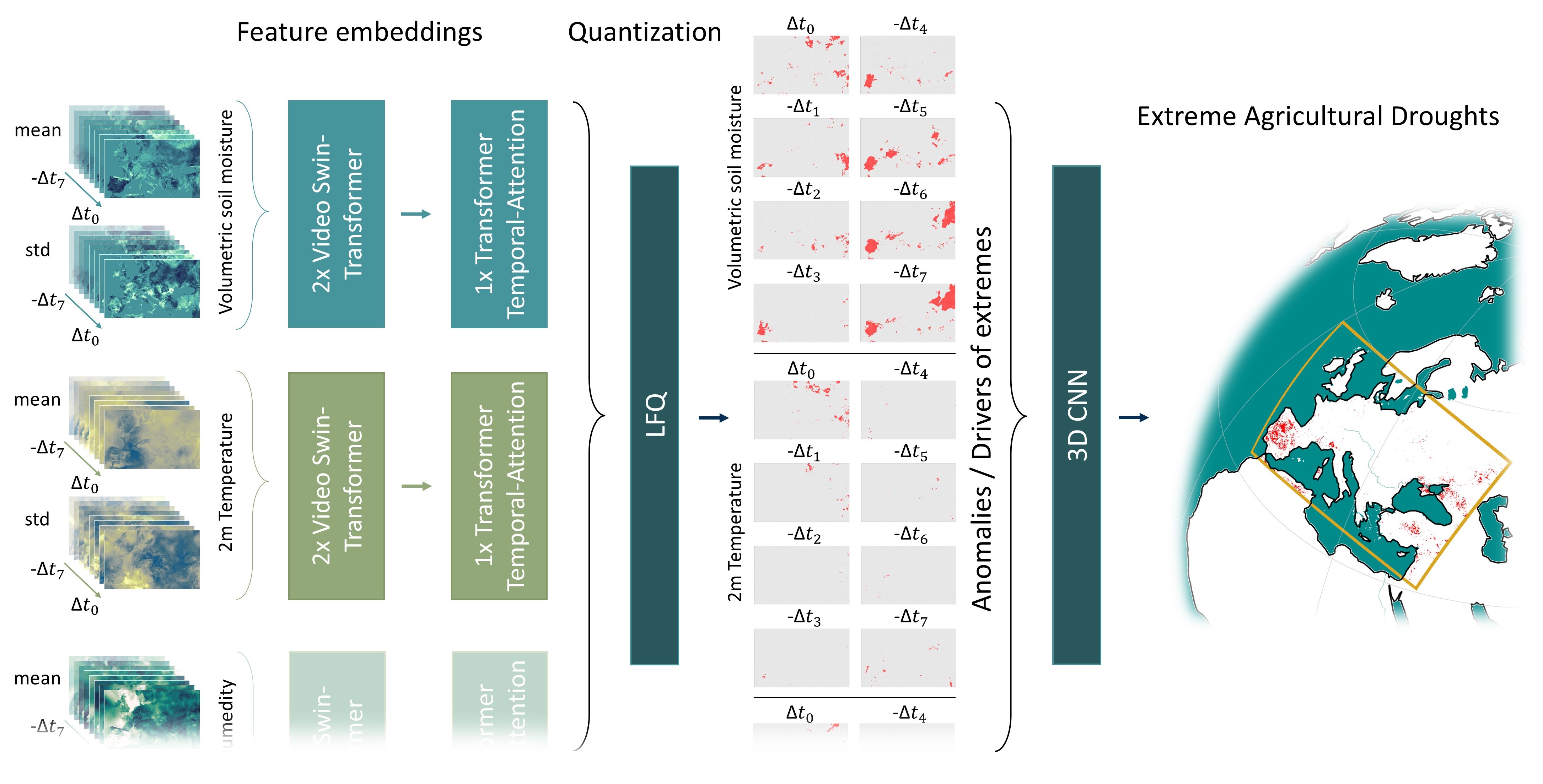}
  \caption{An overview of the proposed model to identify the spatio-temporal relations between extreme agricultural droughts and their drivers. The input variables are first encoded into features. In a subsequent step, a lockup free quantization layer (LFQ) takes the extracted features and classifies the variables into a binary representation of drivers, where we consider the drivers as anomalous events in the input variables. Finally, a classifier is used to predict impacts of extreme events from the identified drivers.}\label{fig:1}
\end{figure}

An overview of our approach is shown in Fig.~\ref{fig:1}.
The input are weekly climate variables at sequential time steps ($\Delta t_{-7},\dots,\Delta t_{0}$) and the model is composed of three main parts. First, a feature extractor extracts relevant features from each input variable independently. The second component is a quantization layer that takes the extracted features as input and classifies the input variables into drivers. The role of the quantization layer is to transform the input variable into a binary representation ($1$ = drivers and $0$ otherwise). 
This ensures that no additional information is encoded as an input to the subsequent classifier except that if the input variable at a specific location and time is a driver or not. The third component is a classifier that takes as input the variables, location, and time where drivers have been identified and predicts where extreme droughts occur at the time step $\Delta t_{0}$. All model components are trained jointly.

We denote the input data as $\X\in\RR^{V \times C \times T \times Lat \times Lon}$, where $V$ is the input variables, $C$ is the channel dimension for each variable (i.e., mean and standard deviation of the week), $T$ is the temporal resolution ($\Delta t_{-T+1},\dots,\Delta t_{0}$), and $Lat$ and $Lon$ are the spatial extensions. The model has two outputs, $\Q\in\ZZ_2^{V \times T \times Lat \times Lon}$ 
representing the binary classification of the input variables into potential drivers of the extreme events, and probabilities $\E\in\RR^{Lat \times Lon}$ 
to predict extreme events at the time step $\Delta t_{0}$.
In the following, we describe the model components:

\textbf{Feature extraction.}~ First, embedded features $f_{\theta}: \X \rightarrow \Z \in\RR^{V \times K \times T \times Lat \times Lon}$ are extracted independently for each input variable $v \in V$ with $K$ embedding dimensions and parameters $\theta$.
The rationale behind this independence is to prevent that drivers leak into other variables.
We use the Video Swin Transformer model \cite{liu2022video} as backbone to capture long-range interrelations across time and space.
The input $\X$ is projected into a higher feature dimension $K$ and followed by two Video Swin Transformer layers. The first layer has two consecutive 3D shifted window blocks for a spatio-temporal feature extraction. The second layer consists of one block for a temporal feature extraction. The later is useful to focus only on the temporal evolution of the variables. An ablation study regarding the backbone is provided in Sec.~\ref{sec:9.3.4}. 

\textbf{Quantization layer.}~ The role of this layer is to map $\Z$ from an embedded space into a compact binary representation $\Q$ suitable for detecting drivers.
Using vector quantization (VQ) \cite{VQ}, each embedded feature vector $z \in \Z$ 
is assigned into a learnable codebook feature vector $z_q \in \Z_q$ based on the Euclidean distance:\begin{align}
VQ : z \rightarrow z_q, \text{where~} q=\argmin_{q \in \{1, \dots, Q\}} \lVert z - z_q \rVert_2 \,,\label{eq:1}
\end{align}
where $Q$ is the size of the codebook. Recently, lookup-free quantization (LFQ) \cite{LFQ} substitutes the learnable codebook with a set of integers $\QQ$ with $|\QQ| = Q$ and represents the embedding space as a Cartesian product of binary numbers. 
This omits the need for a distance metric to do the nearest vector assignment and simplifies the quantization. Based on experimental results, we built the vector quantizer on LFQ with two integers $Q=2$~ ($q=1$ for drivers and $q=0$ otherwise).
Given a feature vector $z$, LFQ first maps $z$ into a scalar value $z_l \in  \Z_l \subset \RR^{V \times 1 \times T \times Lat \times Lon}$. For multi-modality, we use two sequential 3D CNNs on each input variable followed by a shared linear layer that maps $z$ to $z_l$ and reduces the dimensions from $K$ to $1$. The quantization is then given by the sign of $z_l$: \begin{align}
z_q = \text{Linear}\bigl(\sign(z_l)\bigr) = \text{Linear}(-\mathds{1}_{\{z_l \leq 0\}}+\mathds{1}_{\{z_l > 0\}}), \text{~~and~~} q = \mathds{1}_{\{z_l > 0\}} \,,\label{eq:2}
\end{align}
where $q \in \Q$ represents the class ($q=1$ or $q=0$), and Linear is a linear layer that converts $\text{sign}(z_l)$ back to the dimension $K$ of the input after the quantization. Note that $\Z_q$ has only two unique vectors $z_{q=1}$ for a driver and $z_{q=0}$ otherwise.




\textbf{Prediction of extreme events.}~ We use a classifier that predicts the probably of extreme events $\E$ at the time step $\Delta t_{0}$ from the identified drivers $\Z_q$. We use a 3D CNN classifier instead of a transformer to reduce the computations. For training, we only know the ground truth of extremes at time step $\Delta t_{0}$ denoted by $\hat{\E}\in\ZZ_2^{Lat \times Lon}$. While we could compute the cross-entropy between $\E$ and $\hat{\E}$, we found that a single 3D CNN is insufficient to detect all drivers that are correlated with an extreme event. Instead, we use $V{+}1$ 3D CNNs where each predicts $\E_{v}$. While the first $V$ 3D CNNs take the identified drivers for a single variable $v$ as input, the last one takes the identified drivers of all variables as input. The multiple CNNs are only used for training. During inference, $\E$ is only predicted by the multivariate CNN where all variables are jointly used. The loss is thus given by \begin{align}
\mathcal{L}_{(extreme)} ={}& - \sum_{v=1}^{V+1} \bigr(\hat{\E} \log(\E_{v}) + (1 - \hat{\E}) \log(1 - \E_v)\bigl) \Snow \,,\label{eq:3}
\end{align}
where $\Snow \in \ZZ_{2}^{T \times Lat \times Lon}$ is a mask for the valid regions.
We actually utilize a weighted version of $\mathcal{L}_{(extreme)}$ to mitigate the class imbalance issue (Sec.~\ref{sec:9.3.3}).
While the loss $\mathcal{L}_{(extreme)}$ ensures that extreme events can be predicted from the identified drivers, we need to add standard loss terms for the quantization to ensure that the learned codes and thus drivers are compact:
\begin{align}
\mathcal{L}_{(quantize)} ={}& \lambda_{(commit)}\lVert \Z_l - \text{sg}(\sign(\Z_l)) \rVert_{2}^{2} + \lambda_{(ent)} \EE[H\bigl(\sign(\Z_l)\bigr)] - \lambda_{(div)} H[\EE\bigl(\sign(\Z_l)\bigr)]\,.\label{eq:4}
\end{align}
The commitment loss $\lVert \Z_l - \text{sg}(\sign(\Z_l)) \rVert_{2}^{2}$ prevents the outputs of the encoder from growing and encourages $\Z_l$ to commit to the codes \cite{VQ}, where sg stands for the stopgradient operator with zero partial derivative. The term $\EE[H\bigl(\sign(\Z_l)\bigr)]$ encourages that the entropy per quantized code is low \cite{LFQ, jansen2020coincidence}, meaning that it provides more confident assignments. Whereas the term $H[\EE\bigl(\sign(\Z_l)\bigr)]$ increases the entropy inside the batch to encourage the utilization of all codes \cite{LFQ, jansen2020coincidence}. The last important ingredient is a loss that ensures that only spatio-temporal regions are identified that correlate with an extreme event. To this end, we look at regions and intervals where no extreme event occurred and use these examples without drivers. Formally, we use $\hat{\E}_t\in\ZZ_2^{Lat \times Lon}$ as the union of extreme ground truth at all time steps ($\Delta t_{-T+1},\dots,\Delta t_{0}$) and compute the loss by      
\begin{align}
\mathcal{L}_{(driver)} ={}& \lambda_{(driver)}| \Z_q - \text{sg}(z_{q=0}) |(1-\hat{\E}_t) \Snow\,,\label{eq:5}
\end{align}
where $z_{q=0}$ is the quantization code for normal data without drivers.
The model is trained end-to-end with the joint optimization of the loss function:    
\begin{align}
    \min_{\theta, \phi, \psi} \underbrace{\mathcal{L}_{(extreme)}\bigr(\E,\hat{\E}, \Snow\bigl)}_{\text{\teal{predicts extremes from drivers}}} + \underbrace{\mathcal{L}_{(quantize)}\bigr(\Z_l\bigl)}_{\text{\teal{encourages confident quantization}}} + \underbrace{\mathcal{L}_{(driver)}\bigr(\Z_q, \hat{\E}{_t},\Snow, \Z_{q=0}\bigl)}_{\text{\teal{assigns drivers to the same code in the codebook}}}\,,\label{eq:6}
\end{align}
where $\theta, \phi, \psi$ are the learnable parameters, and $\lambda_{(commit)}$, $\lambda_{(ent)}$, $\lambda_{(div)}$, and $\lambda_{(driver)}$ are weighting parameters.
Ablation studies are provided in Sec.~\ref{sec:5.1} and in Appendix Sec.~\ref{sec:9.3}.

\section{Dataset}
\label{sec:4}

\subsection{Defining extreme agricultural droughts from remote sensing}\label{sec:4.1}
We are interested in a specific impact of extreme events namely extreme agricultural drought. To define such extreme event, we rely on the observational satellite-based vegetation health index (VHI) obtained from NOAA \cite{Blended}.
This remote sensing product cannot be directly derived from the input reanalysis, which makes the task very challenging. 
VHI approximates the vegetation state based on a combination of the brightness temperature and normalized difference vegetation index ($\text{VHI}=0$ for unfavorable condition and $\text{VHI}=100$ for favorable condition).
Extreme agricultural droughts are usually defined as VHI < 26 \cite{Blended}. The dataset has a temporal coverage of 1981-onward and is provided globally on a weekly basis. We mapped this dataset into the same domains of the reanalysis data as described in Sec.~\ref{sec:4.2} and used this dataset as ground truth for extreme events. Note that VHI is a general vegetation index and should be interpreted carefully. Details about this index, the dataset and pre-processing are provided in the Appendix Sec.~\ref{sec:9.9}.

\subsection{Climate reanalysis}\label{sec:4.2}
Reanalysis data aim to provide a coherent and complete reconstruction of the historical Earth system state as close to reality as possible. During reanalysis, short-term forecasts from numerical climate models are refined with observations within the so called data assimilation framework \cite{ERA5}. We conducted the experiments on two real-world reanalysis datasets; CERRA reanalysis \cite{CERRA} and ERA5-Land \cite{ERA5-Land}. ERA5-Land is widely used for global climate research and it is provided hourly at $0.1^\circ \times 0.1^\circ$ on the regular latitude longitude grid. CERRA is a regional reanalysis for Europe and is provided originally at $5.5\text{km} \times 5.5 \text{km}$ on its Lambert conformal conical grid with a $3$-hourly temporal resolution. We aggregated these two datasets on a weekly basis and selected the years within the period overlapping with the remote sensing data. In addition, we mapped ERA5-Land into $6$ CORDEX domains \cite{CORDEX} over the globe and conduct experiments on each region separately.
We do the experiments with $6$ common variables from ERA5-Land and CERRA based on their connections to agricultural droughts.
For each variable and week, we computed the mean and standard deviation separately. More details regarding the variables and the domains along with the training/validation/test splits are provided in the Appendix Sec.~\ref{sec:9.8} and Tables \ref{table:17} and \ref{table:18}.

\subsection{Synthetic dataset}\label{sec:4.3}
Although ground truth for extreme droughts can be obtained from remote sensing, an important methodological question remains as how to reliably have a meaningful quantitative evaluation of the identified drivers and their relations to the extreme events. To solve this critical issue, we introduce a new synthetic dataset that mimics the properties of Earth observations including drivers and anomalies. We are aware that the dynamic of the synthetic data are simplified compared to real Earth observations. However, we rely on this generated dataset to perform the quantitative evaluation of the proposed approach. In a first step, we generate the normal data. For instance to generate synthetic data of 2m temperature from CERRA reanalysis, the normal signal at a specific time and location is generated based on the typical value of 2m temperature at that time and location (i.e., the mean or median value from a long-term climatology). The second step is to generate anomalies conditioned on the occurrence of extremes. 
To achieve this, we assign binary spatio-temporally connected flags as extreme events randomly within the datacube and track their precise spatio-temporal locations.
Then based on a predefined coupling matrix between the variables and the extreme event, we generate anomalous events only for the variables that are defined to be correlated with the extremes. We consider these anomalies as the drivers for the extreme events. Finally, we add additional random anomalous events for all variables. We synthesize overall $46$ years of data; $34$ years for training, $6$ subsequent years for validation and the last 6 years for testing. The challenge is to identify the drivers, i.e., the anomalous events that are correlated with extreme events. Examples of the synthetic data are shown in Fig.~\ref{fig:2} and in Appendix in Figs.~\ref{fig:6}-\ref{fig:11}. Technical details are explained in Appendix Sec.~\ref{sec:9.1}.

\section{Experimental results}
\label{sec:5}
First, we conducted experiments and ablation studies on the synthetic datasets (Sec.~\ref{sec:4.3}). We also empirically verified the effectiveness of the proposed design compared to baselines on this synthetic dataset. 
Then, we validate the model on two real-world datasets over five continents in Sec.~\ref{sec:5.2}.

\textbf{Setup and implementation details.}~
We set the hidden dimension $K$ to $16$ by default. The temporal resolution is $T=6$ for the synthetic data and $T=8$ for real-world data. Since, seasonal cycles are typical in climate data, we deseasonalize locally by subtracting the median seasonal cycle and normalizing by the seasonal variance for each pixel. Details regarding the model and implementation setup are given in Appendix Sec.~\ref{sec:9.7}. For evaluation, we use the F1-score, intersection over union (IoU), and overall accuracy on both classes (OA).

\subsection{Experiments on the synthetic datasets}
\label{sec:5.1}

\begin{table}[!b]
  \caption{Driver detection results on the synthetic CERRA reanalysis. The best performance on each metric is highlighted in a bold text. \gray{($\pm$)} denotes the standard deviation for 3 runs.}
  \label{table:1}
  \centering
  \small
  \tabcolsep=1.5pt\relax
  \setlength\extrarowheight{4.5pt}
  \begin{tabular}{c r *{6}l}
    \toprule
    & & \multicolumn{3}{c}{Validation} & \multicolumn{3}{c}{Testing} \\
    \cmidrule{3-5} \cmidrule{6-8}
    & Algorithm & F1-score ($\uparrow$) & IoU ($\uparrow$) & OA ($\uparrow$) & F1-score ($\uparrow$) & IoU ($\uparrow$) & OA ($\uparrow$) \\
    \midrule
    & Naive & 47.93 & 31.52 & 98.61 & 51.24 & 34.45 & 98.37\\
    \midrule
    & Integrated Gradients I~ \cite{Integrated_Gradient} & 24.15\gray{$\pm$9.94} & 14.12\gray{$\pm$6.71} & 92.18\gray{$\pm$2.94} & 23.14\gray{$\pm$7.05} & 13.27\gray{$\pm$4.65} & 91.58\gray{$\pm$2.25}\\
    & Integrated Gradients II \cite{Integrated_Gradient} & 31.23\gray{$\pm$4.40} & 18.58\gray{$\pm$3.05} & 95.26\gray{$\pm$1.03} & 30.34\gray{$\pm$4.27} & 17.95\gray{$\pm$2.94} & 94.19\gray{$\pm$1.22} \\
    \midrule
   \multirow{3}{*}{\rotatebox[origin=c]{90}{\small{One-Class}}} & OCSVM \cite{OCSVM} & 28.21\gray{$\pm$2.49} & 16.44\gray{$\pm$1.67} & 95.64\gray{$\pm$0.16} & 29.98\gray{$\pm$2.26} & 17.66\gray{$\pm$1.54} & 94.91\gray{$\pm$0.19} \\
    & IF \cite{IF} & 34.99\gray{$\pm$0.56} & 21.28\gray{$\pm$0.42} & 97.16\gray{$\pm$0.02} & 37.16\gray{$\pm$0.67} & 22.84\gray{$\pm$0.51} & 96.61\gray{$\pm$0.03} \\
   & SimpleNet \cite{SimpleNet} & 75.31\gray{$\pm$0.07} & 60.39\gray{$\pm$0.10} & 99.20\gray{$\pm$0.01} & 73.50\gray{$\pm$0.24} & 58.11\gray{$\pm$0.30} & 98.91\gray{$\pm$0.02}\\
    \midrule
    \multirow{2}{*}{\rotatebox[origin=c]{90}{\small{Rec.}}} & STEALNet \cite{STEALNet} & 55.98\gray{$\pm$0.90} & 38.87\gray{$\pm$0.86} & 98.47\gray{$\pm$0.03} & 57.74\gray{$\pm$0.95} & 40.60\gray{$\pm$0.93} & 98.22\gray{$\pm$0.03} \\
   & UniAD \cite{UniAD} & 47.53\gray{$\pm$0.17} & 31.18\gray{$\pm$0.14} & 97.44\gray{$\pm$0.02} & 49.23\gray{$\pm$0.41} & 32.65\gray{$\pm$0.36} & 97.18\gray{$\pm$0.05} \\
    \midrule
   \multirow{3}{*}{\rotatebox[origin=c]{90}{\small{MIL}}} & DeepMIL \cite{DeepMIL} & 70.68\gray{$\pm$1.61} & 54.68\gray{$\pm$1.91} & 99.22\gray{$\pm$0.03} & 71.54\gray{$\pm$1.60} & 55.72\gray{$\pm$1.92} & 99.09\gray{$\pm$0.04} \\
   & ARNet \cite{ARNet} & 72.92\gray{$\pm$0.85} & 57.39\gray{$\pm$1.06} & 99.26\gray{$\pm$0.01} & 73.68\gray{$\pm$0.86} & 58.34\gray{$\pm$1.08} & 99.13\gray{$\pm$0.02} \\
   & RTFM \cite{RTFM} & 60.09\gray{$\pm$0.31} & 42.95\gray{$\pm$0.31} & 98.34\gray{$\pm$0.03} & 61.88\gray{$\pm$0.28} & 44.81\gray{$\pm$0.30} & 98.12\gray{$\pm$0.02} \\
   \midrule
    & Ours$^*$~~~~~~~ & 82.78\gray{$\pm$0.53} & 70.63\gray{$\pm$0.78} & 99.51\gray{$\pm$0.02} & 80.44\gray{$\pm$0.70} & 67.28\gray{$\pm$0.97} & \textbf{99.29}\gray{$\pm$0.04} \\
    & \logo Ours$^\dag$~~~~~~~ & \textbf{83.45}\gray{$\pm$0.37} & \textbf{71.60}\gray{$\pm$0.54} & \textbf{99.52}\gray{$\pm$0.01} & \textbf{80.65}\gray{$\pm$0.24} & \textbf{67.58}\gray{$\pm$0.33} & \textbf{99.29}\gray{$\pm$0.01} \\
    \bottomrule
\multicolumn{3}{l}{\footnotesize{$^*$Video Swin Transformer backbone \cite{liu2022video}}} & \multicolumn{3}{l}{\footnotesize{$^\dag$Mamba backbone \cite{gu2023mamba}}}\\
\end{tabular}
\end{table}

\begin{figure}[!h]
  \centering
  \includegraphics[draft=\draft, width=.95\textwidth]{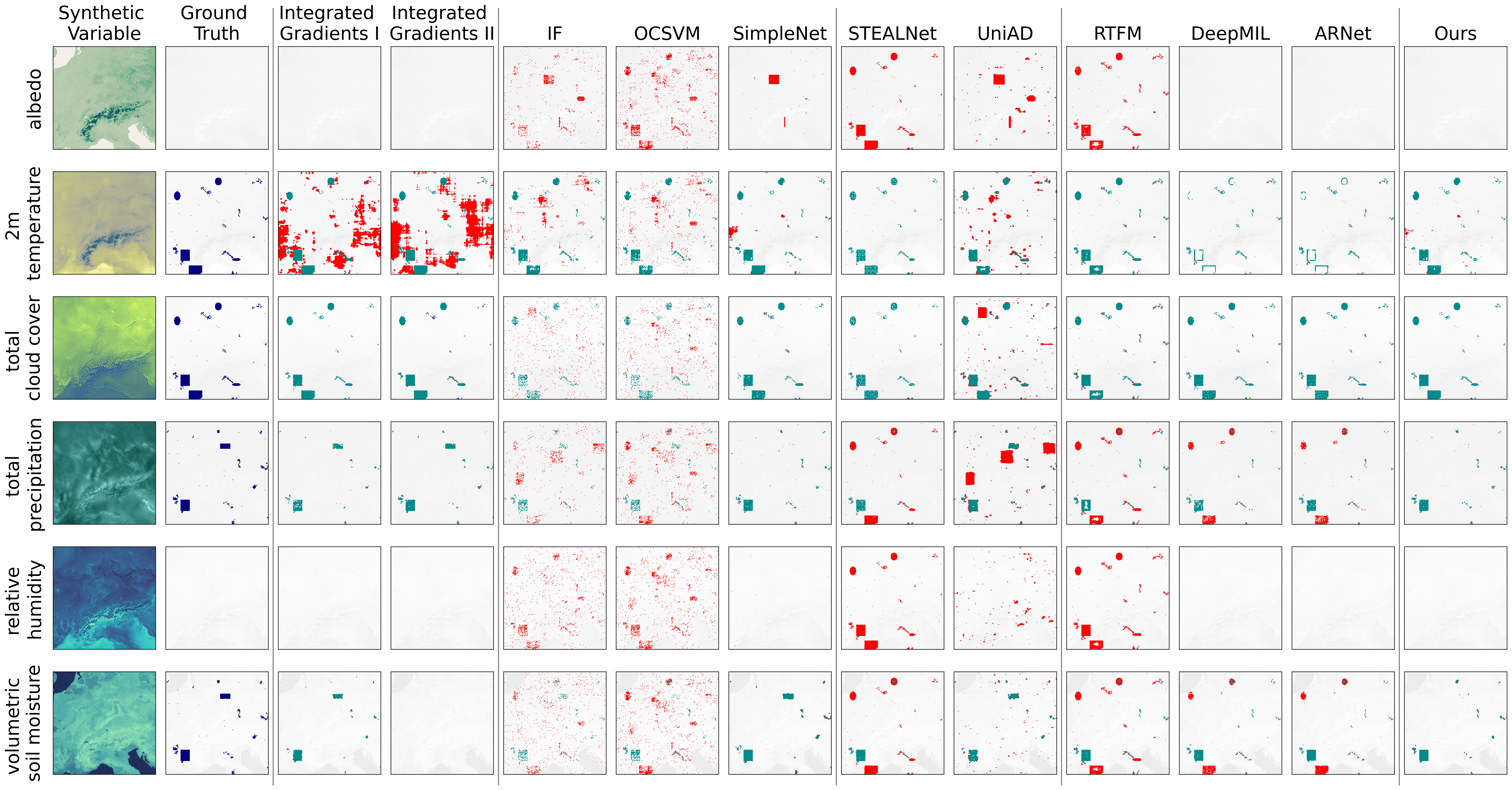}
  \caption{Qualitative results on the synthetic CERRA reanalysis from the test set at time step 2160. \legendbox{teal} is the prediction, \legendbox{NavyBlue} is the ground truth, and \legendbox{red} is the false positive. Albedo and relative humidity are not correlated with extremes, meaning that they do not contain drivers, but only random anomalies.}\label{fig:2}
\end{figure}

We show the results on the Synthetic CERRA described in Sec.~\ref{sec:4.3} and in Appendix Table \ref{table:3}. The generated dataset mimics a set of variables ($V=6$) using statistics from the real-world CERRA reanalysis \cite{CERRA}. We artificially correlated four variables with extremes (2m temperature, total cloud cover, total precipitation, and volumetric soil moisture) and kept two variables uncorrelated (albedo and relative humidity).

\textbf{Comparison to the baselines.}~
We compare the new approach to interpretable forecasting approaches using integrated gradients \cite{Integrated_Gradient} and to $8$ baselines from $3$ different categories of anomaly detection approaches; one-class unsupervised \cite{OCSVM, IF, SimpleNet}, reconstruction-based \cite{STEALNet, UniAD}, and multiple instance learning \cite{DeepMIL, ARNet, RTFM}. We also compare to a naive baseline which labels all variables as drivers for pixels where extreme events occur. The implementation details of these baselines are given in Appendix Sec.~\ref{sec:9.5}.

The quantitative results are shown in Table \ref{table:1}. The naive baseline is impacted by two main issues; first by the time delay between drivers and extreme events, and second not all variables are correlated with the extremes. The second issue affects the one-class and reconstruction-based baselines where they suffer mostly from false positives. 
In fact, both integrated gradients models achieve high F1-scores for detecting extremes (93.32 for Integrated Gradients I and 93.80 for Integrated Gradients II), but they have worse performance on identifying the drivers. SimpleNet is trained with our model as a feature extractor which explains its good performance. 
However, SimpleNet showed a drop of performance when it is tested on other datasets (see Appendix Sec.~\ref{sec:9.2} for results on two more synthetic datasets). Among the reconstruction-based approach, STEALNet outperforms UniAD. 
This is probability because STEALNet exploits more weakly supervision information during training by maximizing the reconstruction loss for locations with extreme flags. 
MIL-based baselines are more suitable for the task.
Finally, our model consistently outperforms the baselines on all metrics.
 \begin{wrapfigure}{r}{0.5\textwidth}
\includegraphics[draft=\draft, width=.98\linewidth]{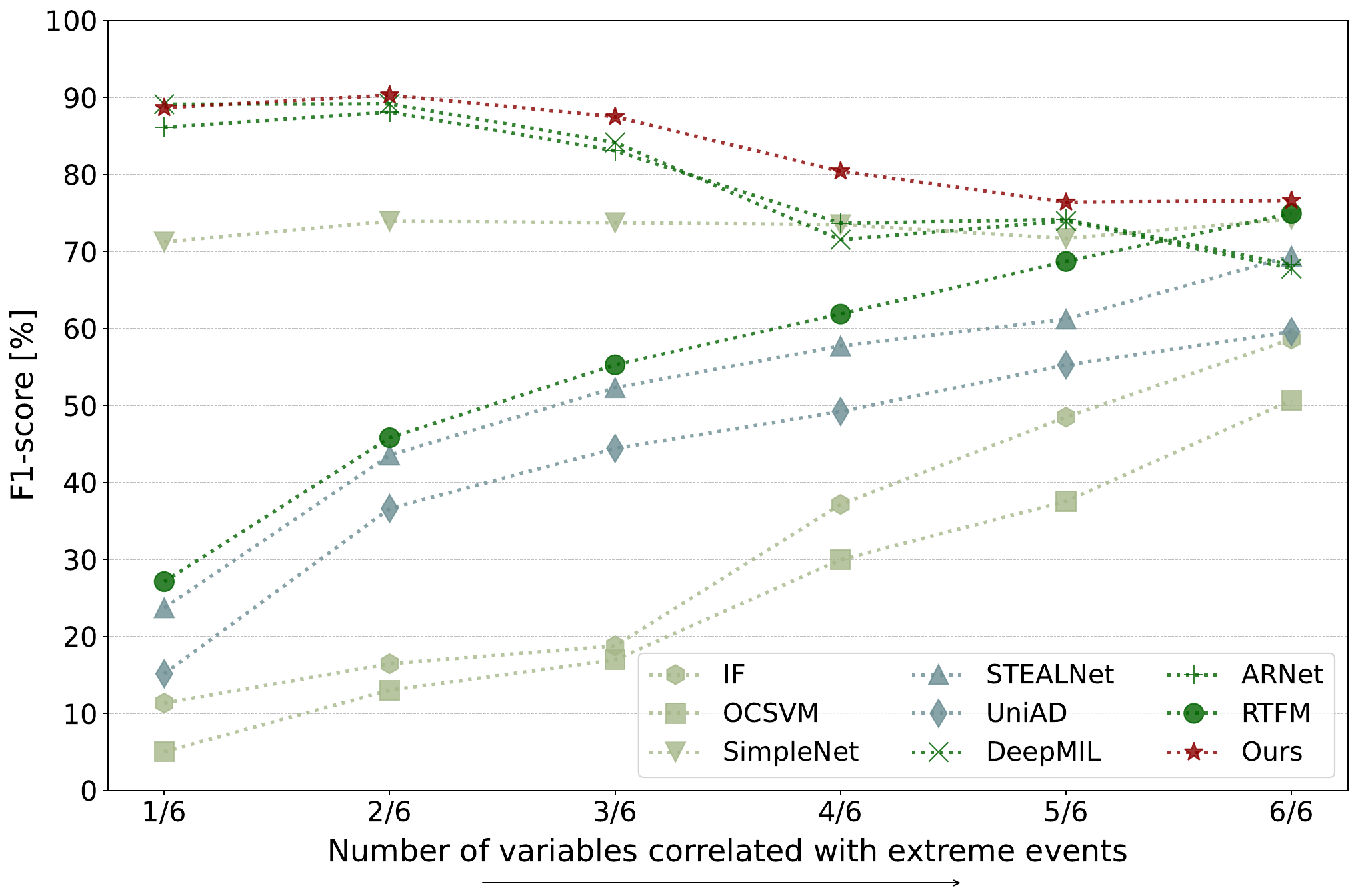}
\caption{F1-score with different correlation settings between the input variables and extremes.}\label{fig:3}
\end{wrapfigure}
Qualitative samples in comparison with baselines are presented in Fig.~\ref{fig:2}. The qualitative examples indicate that our model and the MIL-based baselines except RTFM are capable of learning which variables are correlated with the extremes. The main weakness of RTFM is the reliance on feature magnitudes and the cross attention module (see Appendix Sec.~\ref{sec:9.5} and Table \ref{table:2}), which make it more prone to produce false positives. Other baselines predict incorrect relations between the variables and extremes. Regarding the explainable AI methods, when we add more interactions between the variables (Integrated Gradients II), the gradients tend to omit some variables (soil moisture). Both integrated gradients models have also difficulties with the synthetic t2m, which includes red noise by design. These results demonstrate that networks that predict the extremes directly from the input variables utilize much more information even when it is not correlated with an extreme. It is thus beneficial to introduce a bottleneck into the network that enforces the network to explicitly identify drivers of extremes.


\textbf{Performance on easy-to-hard correlation settings.}~
We conduct an additional experiment to assess the model performance in relation to the correlation setup between the variables and extremes.
We generate a synthetic CERRA dataset starting with only one correlated variable with the target extreme.
We then generate different versions of the dataset by increasing the number of correlated variables with the extremes up to $100\%$.
This analysis allows us to point out the strengths and weaknesses of the comparative models for different scenarios and where our model becomes more effective, as well as where it could struggle most.
The results are shown in Fig.~\ref{fig:3}. One-class, reconstruction-based and RTFM baselines benefit with increasing the number of correlated variables. In case of 6/6, the task reduces to an anomaly detection task. The performance of our model and MIL-based baselines generally decreases when the number of correlated variables increases as the task of finding all correlated anomalies becomes harder. Nevertheless, our approach performs best in all settings. 
 
\textbf{Ablation study.}~
We conducted a set of ablation studies. This includes three main experiments:
\begin{table}[t]
  \caption{Ablation studies from the validation set. The metric is F1 on the driver/\olive{extreme} detection.}
  \label{table:2}
  \centering
  \small
  \tabcolsep=1.0pt\relax
  \setlength\extrarowheight{0pt}
  \begin{tabular}{*{12}l}
        \multicolumn{4}{c}{(a) Loss function} & & \multicolumn{4}{c}{(b) Key model architecture} & & \multicolumn{2}{c}{(c) $T$ dimension} \\
      \cmidrule[\heavyrulewidth]{1-4} \cmidrule[\heavyrulewidth]{6-9} \cmidrule[\heavyrulewidth]{11-12}
      $\mathcal{L}_{(quantize)}$ & $\mathcal{L}_{(driver)}$ & $\mathcal{L}_{(extreme)}$ & F1-score ($\uparrow$) & & Temporal- & Cross- & Shared- & F1-score ($\uparrow$) & & $T$ & F1-score ($\uparrow$)\\
       &  & multi-head & & & attention & attention & ($f_{\theta}$) & & & & \\
      \cmidrule{1-4} \cmidrule{6-9} \cmidrule{11-12}
      \cmark & \xmark & \xmark & 29.99 / \olive{48.94} & & \cmark & \xmark & \cmark & 69.59 / \olive{90.42} & & 1 & 83.38 / \olive{69.25}\\
      \cmark & \xmark & \cmark & 02.51 / \olive{92.88} & & \xmark & \cmark & \xmark & 67.18 / \olive{93.78} & & 4 & 81.36 / \olive{90.68} \\
      \cmark & \cmark & \xmark & 31.23 / \olive{66.40} & & \xmark & \xmark & \xmark & 82.39 / \olive{91.97} & & 6 & \textbf{82.78 / \olive{92.45}} \\
      \cmark & \cmark & \cmark & \textbf{82.78 / \olive{92.45}} & & \cmark & \xmark & \xmark & \textbf{82.78 / \olive{92.45}} & & 8 & 77.33 / \olive{90.30} \\
      \cmidrule[\heavyrulewidth]{1-4} \cmidrule[\heavyrulewidth]{6-9} \cmidrule[\heavyrulewidth]{11-12}
  \end{tabular}
  \vspace{-5mm}
\end{table}

\textbf{Loss functions.}~
As shown in Table \ref{table:2} (a), $\mathcal{L}_{(quantize)}$ and $\mathcal{L}_{(driver)}$ are essential for training. As other quantization models, ours can not be trained without $\mathcal{L}_{(quanitze)}$, which ensures that the outputs of $f_{\theta}$ do not grow and commit to the binary embedding. 
$\mathcal{L}_{(driver)}$ unifies the representation of drivers for all variable as $q=1$, which boosts the extreme detection. 
Moreover, the results demonstrate the impact of using $V{+}1$ 3D CNNs (multi-head) instead of one for $\mathcal{L}_{(extreme)}$. If a single 3D CNN is used, drivers are only identified in a small subset of variables. We discuss this more in detail in Sec.~\ref{sec:9.3.7}.

\textbf{Feature extractor.}~
In Table \ref{table:2} (b), we show the benefit of having independent feature extractors for driver detection. In a first experiment, we share the feature extractor $f_{\theta}$ among the variables. The performance is worst. Second, we replaced the temporal attention by a cross attention between the variables similar to \cite{nguyen2023climax} and \cite{lessig2023atmorep} where each variable performs a cross attention with the other variables. We can see a drop of performance for the second experiment. We noticed that anomalies propagate between variables when adding connections in the feature extraction stage. This also explains the poor performance of RTFM compared to other MIL baselines. The best performance is shown for the proposed setup (last row), which also shows the benefit for the temporal attention. 

\textbf{Temporal resolution $T$.}~
Table \ref{table:2} (c) evaluates the impact of the temporal resolution on driver and extreme detection. $T{=}6$ provides a good balance between driver and extreme detection.
More ablation studies on other aspects of the model design can be found in Appendix Sec.~\ref{sec:9.3}.

\subsection{Experiments on real-world datasets}
\label{sec:5.2}

We evaluate our model on two reanalysis data with diverse geographical and climate regions (Sec.~\ref{sec:4.2}). The input for the experiment is the normalized mean and standard deviation of each week. 
We exclude pixels over water surfaces, desert, and snow.
In addition to the quantitative evaluation on the synthetic data, we aim to verify our method considering the following aspects:

\textbf{Quantitative results.}~
We expect that the developed model can identify drivers in real-world scenarios. We demonstrate this by measuring how well the model can predict extreme agricultural droughts from the identified drivers.
The results verify that the model can predict the droughts across different regions and datasets (see Appendix Sec.~\ref{sec:9.4} and Table \ref{table:14}). Note that compared to the synthetic dataset, the real-world drought prediction is much more difficult. 

\begin{figure}[!h]
  \centering
  \includegraphics[width=1.\textwidth]{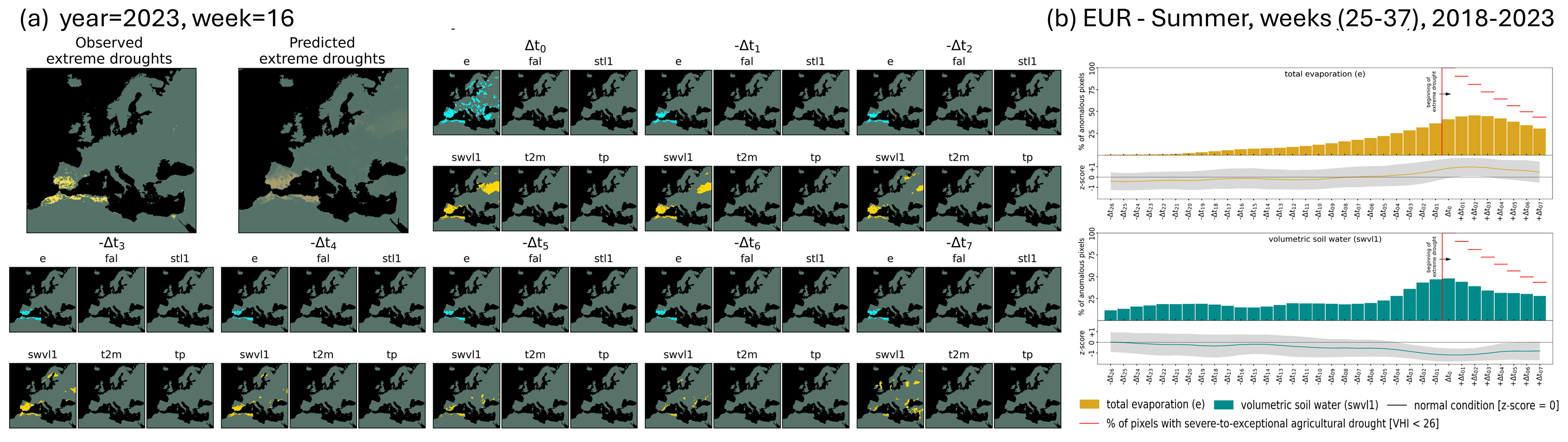}
  \caption{(a) Qualitative results on ERA5-Land over the EUR-11 domain. Shown are the identified drivers localized spatio-temporally $7$ weeks before the extreme agricultural drought events. (b) Temporal evolution of drivers during the extremes.}\label{fig:4}
\end{figure}

\textbf{Extreme detection without anomaly detection.}~
We trained the model without the quantization step, meaning without driver detection. This can be considered as an upper bound on the extreme detection accuracy since there is no information reduction by the quantization. We found that when we trained on the synthetic and real-world EUR-11 data, the F1-score for detecting extreme events increased only by $\sim0.96\%$ and $\sim1.93\%$, respectively, compared to the model with quantization (see Appendix Table \ref{table:12}). 
This verifies that the detected drivers are highly correlated with the extremes. 

\textbf{Qualitative results and spatial distribution.}~
In Fig.~\ref{fig:4} (a), we show the spatial distribution of the identified drivers at a specific time over EUR-11. Shown are the identified drivers up to 7 weeks ($\Delta t_{-7}$) before the extreme agricultural droughts at time $\Delta t_0$. We can see that the prediction of drivers and extremes are spatially correlated with the ground truth.

\textbf{Physical consistency.}~
In Fig.~\ref{fig:4} (b), we show the relation between the input reanalysis data, extreme droughts, and identified drivers. For this experiment, we selected pixels with extreme events during summer (weeks $25$-$38$) and visualize the average distribution of drivers with time. The red line at $\Delta t_0$ indicates the beginning of the extreme droughts. $Z_{score}$ in the underneath curve represents the deviation from the mean computed from the climatology. 
It is expected that evaporation reduces soil moisture, which dries out the  soil and vegetation \cite{Miralles_2019}.
Our model indicates that over Europe, the evaporation and soil moisture are the most informative variables to detect drivers related to extreme droughts.
All pixels experienced a pronounced decline in soil moisture and an increase in evaporation as the events evolved.
Please see Sec.~\ref{sec:9.4.3} for more discussion on the scientific validity.


\section{Discussions and conclusions}
\label{sec:6}
We introduced a model that can identify the spatio-temporal relations between impacts of extreme events and their drivers. For this, we assumed that there exist precursor drivers, primarily as anomalies in assimilated land surface and atmospheric data, for every observable impact of extremes.
We demonstrated the effectiveness of our approach by measuring to which degree the identified drivers can be used to predict extreme agricultural droughts.
Apart from experiments on two real-world datasets, we also presented a new framework to generate synthetic datasets that can be used for spatio-temporal anomaly detection and climate research.
The results on the synthetic datasets show that the approach is not limited to droughts and can be applied to other extremes.
While we have shown that our approach outperforms other approaches, the study has some limitations. 
First, evaluating ability to handle a very large number of climate variables in a unified model needs further examination. Similarly, performing an additional pre-processing of specific variables like accumulating precipitation over many weeks might also improve the results. 
Second, modelling the temporal relations is limited by the time window $T$.
Moreover, teleconnections of climatic anomalies can occur in distant regions on Earth, e.g., affects of El Ni{\~n}o and La Ni{\~n}a variability on drought and flood \cite{Das_2009}. Modelling and disentangle such large spatio-temporal relations across the globe is an open research problem. 
Third, it would be appealing to provide scores for drivers instead of a binary classification. This could be achieved by measuring the distance to the nearest code in the LFQ.
Forth, the prediction of the model depends on the capacity of reanalysis data to accurately represent the local environmental factors and land–atmospheric feedbacks. Most importantly, drawing conclusions on drivers from weak predictive models may lead to unreliable interpretations.
Finally, our model does not identify causal relationships.

Despite these limitations, our approach demonstrates a clear capability in identifying drivers and anomalies in climate data which would allow a more timely event attributions during and right after extreme events. The identified spatio-temporal relations between extreme events and their drivers could support the understanding and forecasting of extremes.

\newpage

\section{Acknowledgments and Disclosure of Funding}
We thank Petra Friederichs, Till Fohrmann, Sebastian Buschow, and Svenja Szemkus for the insightful discussions on detection and attribution of weather and climate extremes. We would also like to thank Olga Zatsarynna and Emad Bahrami for the technical discussion related to feature extraction. Finally, we thank the four anonymous reviewers for their comments and suggestions which improved the quality of this paper.

This work was supported by the Deutsche Forschungsgemeinschaft (DFG, German Research Foundation) – SFB 1502/1–2022 – project no. 450058266 within the Collaborative Research Center (CRC) for the project Regional Climate Change: Disentangling the Role of Land Use and Water Management (DETECT) and by the Federal Ministry of Education and Research (BMBF) under grant no. 01IS24075C RAINA.

We acknowledge EuroHPC Joint Undertaking for awarding us access to Leonardo at CINECA, Italy, through EuroHPC Regular Access Call - proposal No. EHPC-REG-2024R01-076. The authors also gratefully acknowledge the granted access to the Marvin cluster hosted by the University of Bonn.


\medskip

{
\small
\bibliographystyle{unsrt}
\bibliography{ref}
}

\newpage
\appendix
\label{sec:9}
\section{Synthetic data}
\label{sec:9.1}
To the best of our knowledge, ground truth labels of drivers or anomalies which are correlated with extreme events impacts barely exist. This is especially the case when it comes to drivers or anomalous events which are related to a specific definition of an extreme event within the Earth system (i.e., drought or flood). For this reason, we designed a framework to generate artificial datasets which can be adopted to the task. Our framework is inspired by and related to Flach, et al.~\cite{Flach}. However, we differ in the following two aspects:\\
(1) First, we aim to generate multivariate anomalous that are correlated with a specific extreme event, while their aim is to generate anomalous that can occur simultaneously in multivariate data streams similar to \cite{xu2022anomaly, lee2021weakly, zhong2024patchad, tuli2022tranad}. Think of an increasing/decreasing of temperature, existing approaches are interested in temperature anomalies regardless of the subsequent extreme events they might cause, while we focus on temperature anomalies that can cause a particular extreme events in the near future. (2) Second, we generate the synthetic data based on real-world data stream signals, while they use trigonometric functions (i.e., sine function) to mimic Earth observations across spacetime.\\

In the following, we explain our overall framework in more details:\\
(1) First, we generate the normal base signals $\B \in\RR^{V \times T \times Lat \times Lon}$ for a set of different variables $V$, where $T$ is the temporal extension, and $Lat$ and $Lon$ are the spatial extensions.
For instance, to synthesize CERRA \cite{CERRA} \textbf{B} signals, we take the mean values from the CERRA climatologoy pixel-wise. This represents the typical value of \textbf{B} at specific time (week) and location (lat, lon). By definition, \textbf{B} inherits the intrinsic properties of the simulated variables including the existence of seasonality and correlations among variables.\\
(2) In the next step, we induce binary extreme events $\mathbf{E}^{ex} \in\ZZ_2^{T\times Lat \times Lon}$ within the datacube and track the precise spatio-temporal location of these events. Similar to Flach, et.~al \cite{Flach}, the type and duration of extreme events vary within the datacube. For instance, we alter the duration of the events between long and short extreme events. The spatial distribution of the extreme event vary also between a local event at one pixel (LocalEvent), a rectangular event (CubeEvent), a Gaussian shape (GaussianEvent), an onset event that starts at specific time and lasts until the end of the series (OnsetEvent), and a random walk event that starts at a specific pixel and affecting neighboring pixels with time (RandomWalkEvent).
\begin{figure}[!h]
  \centering
  \includegraphics[width=.8\textwidth]{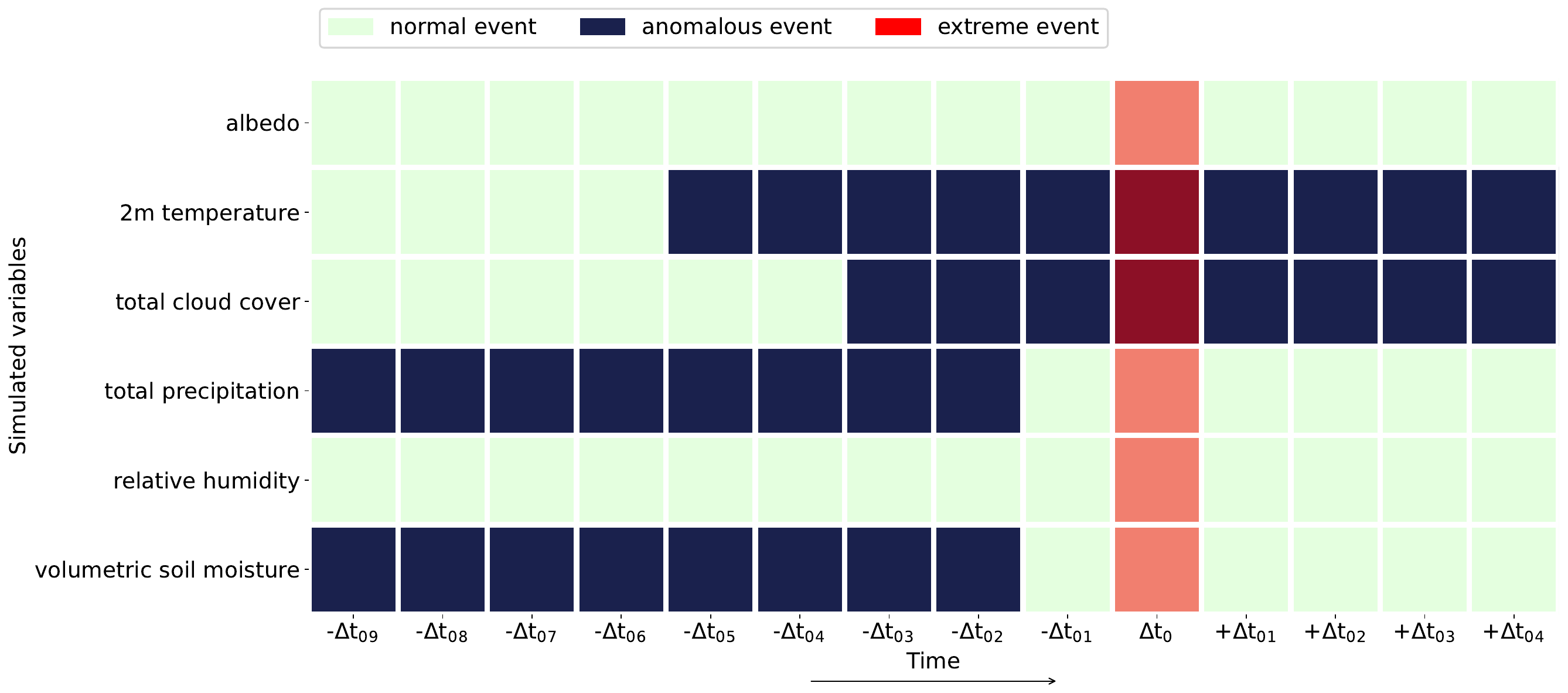}
  \caption{An example of the randomly generated coupling matrix between the synthetic variables and extremes for the synthetic CERRA dataset.}\label{fig:5}
\end{figure}

(3) We randomly define a coupling matrix \textbf{M} between the variables and the extreme event. An example is shown in Fig.~\ref{fig:5}, i.e., the anomalous events for 2m temperature start $5$ time steps ($-\Delta T_{a}$) before the extreme and last for $4$ time steps afterwards ($+\Delta T_{a}$). While albedo and relative humidity are not coupled with the extremes. Based on \textbf{M} and $\mathbf{E}^{ex}$, we generate the binary anomalous events matrix $\mathbf{E}^{a} \in\ZZ_2^{T\times Lat \times Lon}$.\\
(4) Similar to steps (2)-(3), we generate binary random anomalous events $\mathbf{E}^{r} \in\ZZ_2^{T\times Lat \times Lon}$. However, these events are uncorrelated with the extremes and generated randomlies for all variables.\\
(5) We sample noise signals $\N \in\RR^{V \times T \times Lat \times Lon}$ for each variable. The noise signals could be sampled from a normal Gaussian distribution (GaussianNoise), a standard Cauchy distribution (CauchyNoise), a double exponential distribution (LaplaceNoise), or from a spatiotemporal correlated noise across the datacube (RedNoise).
(6) Using the indices $v \in \{1, \dots, V\}$, $t \in \{1, \dots, T\}$, $lat \in \{1, \dots, Lat\}$, and $lon \in \{1, \dots, Lon\}$, the synthetic signal $\mathbf{\Phi}_{(v, t, lat, lon)}$ is generated with the following formula:
\begin{align}
\mathbf{\Phi}_{(v, t, lat, lon)} ={}& \textbf{B}_{(v, t, lat, lon)} + \mathbf{\Lambda}_{(v, t, lat, lon)} . \mathbf{\Theta}_{(v, t, lat, lon)}\,,\label{eq:7}\\
\mathbf{\Theta}_{(v, t, lat, lon)} ={}& \textbf{B}_{(v, t, lat, lon)}. (2^{(kb . ({\mathbf{E}_{(v, t, lat, lon)}^{r}} \vee {\mathbf{E}_{(v, t, lat, lon)}^{a}}))} - 1)\notag\\
&+ \mathbf{N}_{(v, t, lat, lon)} . 2^{(kn . ({\mathbf{E}_{(v, t, lat, lon)}^{r}} \vee {\mathbf{E}_{(v, t, lat, lon)}^{a}}))}\notag\\
&+ ks . ({\mathbf{E}_{(v, t, lat, lon)}^{r}} \vee {\mathbf{E}_{(v, t, lat, lon)}^{ex}}) . \sigma_N\,,\label{eq:8}\\
\mathbf{\Lambda}_{(v, t, lat, lon)} ={}&
    \begin{cases}
      \dfrac{\mathbf{\Delta}_{(v, t, lat, lon)}}{\delta}, & \text{if}\enspace \mathbf{E}_{(v, t, lat, lon)}^{a} = 1 ,\\
      +1, & \text{otherwise.}\enspace
    \end{cases}\,,\label{eq:9}\\
\mathbf{\Delta}_{(v, t, lat, lon)} ={}& 
    \begin{cases}
      -1, & \text{if}\enspace \mathbf{\Theta}_{(v, t, lat, lon)} \leq 0 ,\\
      +1, & \text{otherwise,}\enspace
    \end{cases}\,,\label{eq:10}
\end{align}
where $\mathbf{\Theta}_{(v, t, lat, lon)}$ is the induced anomaly, $kb, kn$, and $ks$ are control parameters for the events magnitudes, $\sigma_N$ is the standard deviation of the noise signal, $\delta\in\{-1, 1\}$ is the predefined coupling sign with extremes from \textbf{M}, and $\mathbf{\Lambda}_{(v, t, lat, lon)}$ controls the sign of the induced anomaly for extremes (i.e., a deficiency in soil moisture ($\delta=-1$) and an increased in temperature ($\delta=+1$) during an extreme drought event).

Using this framework, we generated 3 types of datasets; (1) synthetic CERRA reanalysis (Tables \ref{table:3}), (2) synthetic NOAA remote sensing (Tables \ref{table:4}), and (3) synthetic artificial data (Tables \ref{table:5}).

We generate the NOAA base signals from \cite{Blended}. For the NOAA and artificial datasets, we also considered artificial linear (LinearCoupling) and non-linear (QuadraticCoupling) dependencies among the variables as additional data properties. For instance to generate a new dependent base signal from independent bases:
\begin{align}
\mathbf{B}_{(t, lat, lon)} ={}&
    \begin{cases}
      \sum_{v=1}^{V} w_{(v)} . B_{(v)}, & \text{if}\enspace \text{LinearCoupling},\\
      \sum_{v=1}^{V} \frac{1}{\sqrt{2}} w_{(v)} . (B_{(v)}^2 - 1), & \text{otherwise,}\enspace
    \end{cases}\,,\label{eq:11}\\
\mathbf{E}_{(t, lat, lon)}^{r} ={}& \vee_{v=1}^{V} \mathbf{E}_{(v, t, lat, lon)}^{r}\,,\label{eq:12}
\end{align}
 where $w_{(v)}\in\RR^{V}$ are weighted coefficients sampled either from a normal (NormWeight) or a Lablace (LaplaceWeight) distribution. In addition, we add an option to generate disturbed weights where weights vary spatio-temporally based on the locations of anomalies. Anomalous events that occur in one of the independent base signals propagate to the new generated dependent signals.

\begin{figure}[!h]
  \centering
  \includegraphics[draft=\draft,width=.87\textwidth]{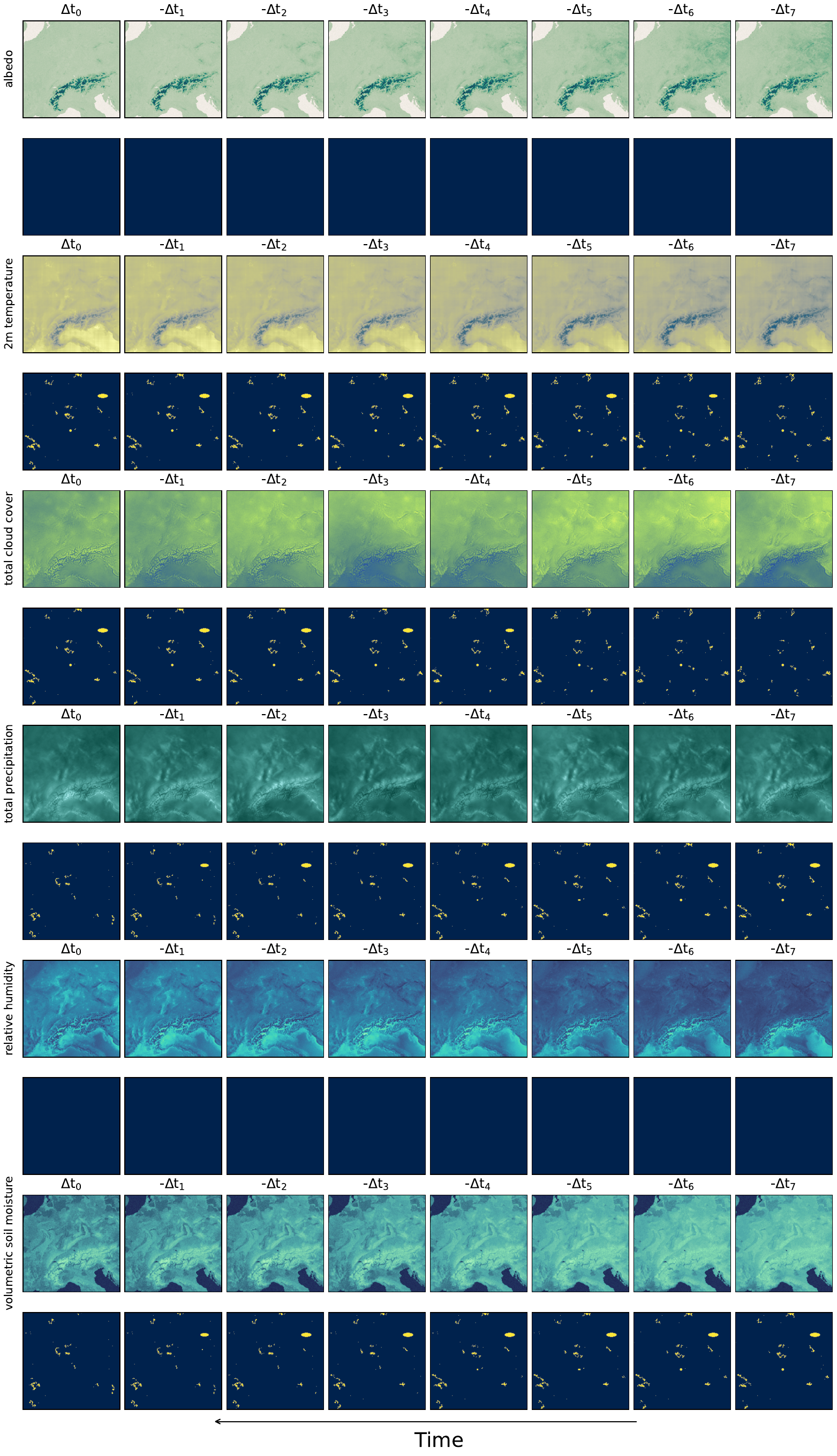}
  \caption{Examples of the synthetic CERRA reanalysis data described in Table \ref{table:3}. The drivers/anomalies \legendbox{Gold} are visualized under each variable directly. Here, albedo and relative humidity are not correlated with the extremes.}\label{fig:6}
\end{figure}

\begin{figure}[!h]
  \centering
  \includegraphics[draft=\draft,width=.9\textwidth]{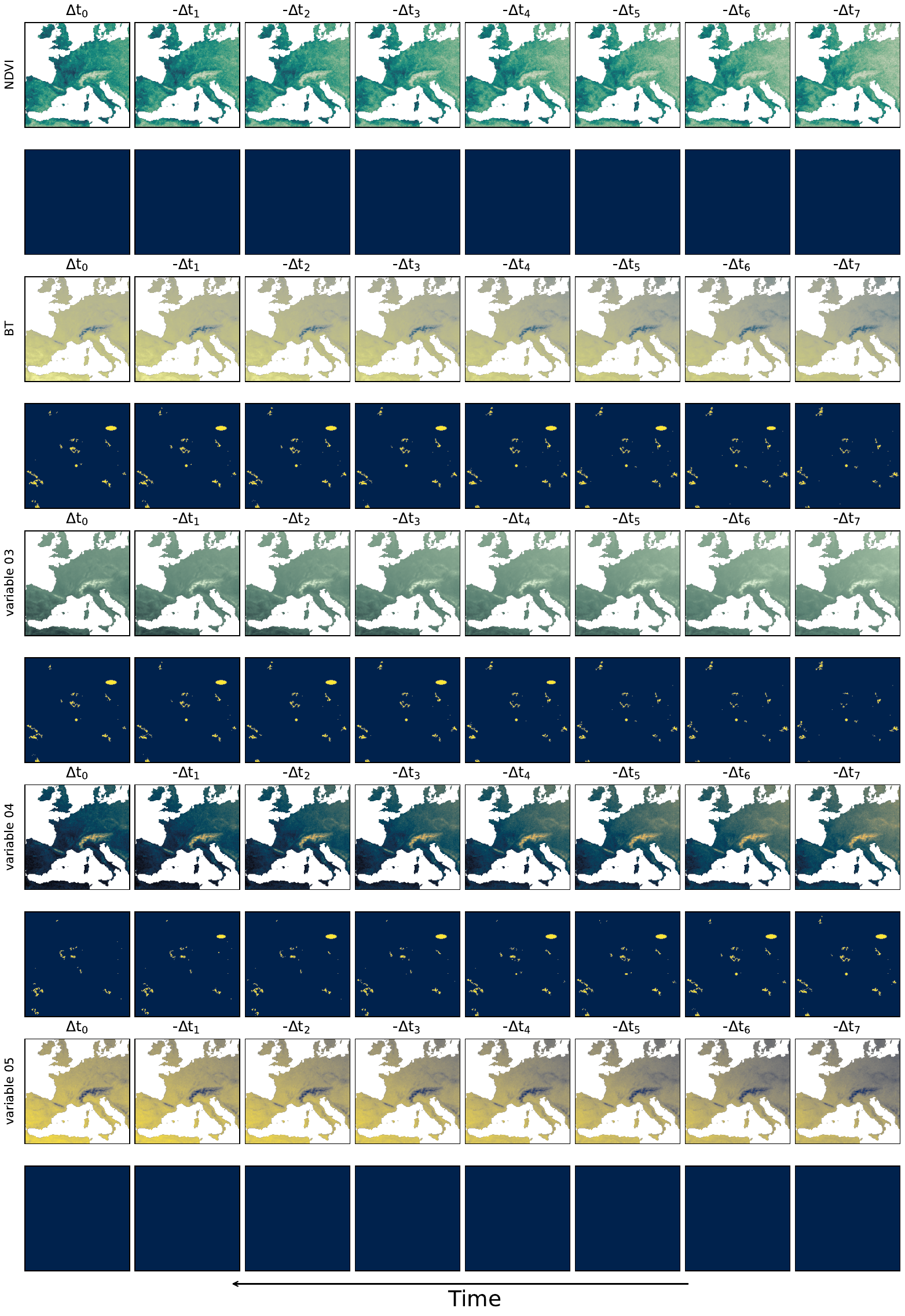}
  \caption{Examples of the synthetic NOAA remote sensing described in Table \ref{table:4}. The drivers/anomalies \legendbox{Gold} are visualized under each variable directly. Here, NDVI and variables 05 are not correlated with the extremes.}\label{fig:7}
\end{figure}

\begin{figure}[!h]
  \centering
  \includegraphics[draft=\draft,width=.87\textwidth]{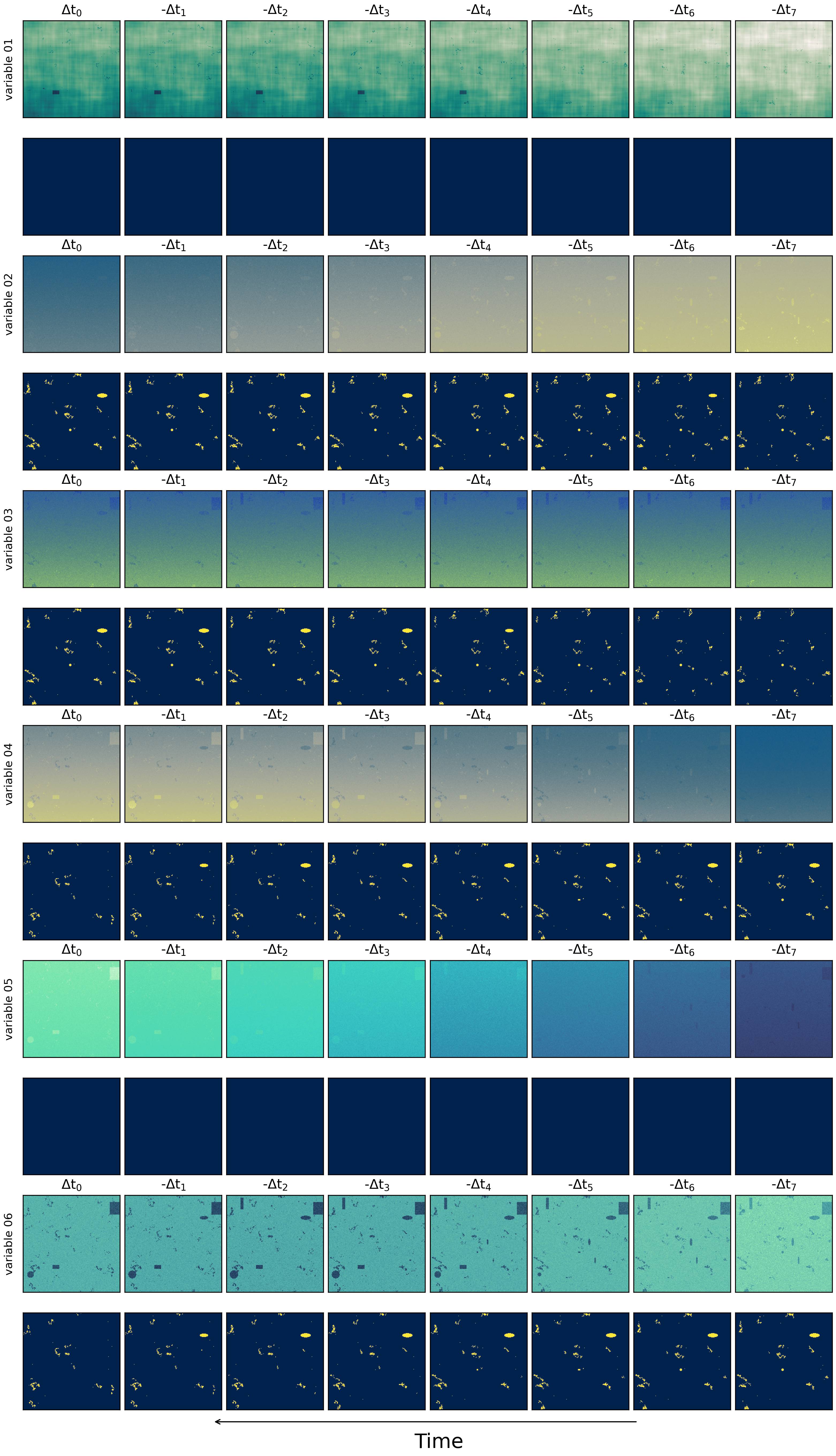}
  \caption{Examples of the synthetic artificial data described in Table \ref{table:5}. The drivers/anomalies \legendbox{Gold} are visualized under each variable directly. Here, variables 01 and 05 are not correlated with the extremes.}\label{fig:8}
\end{figure}

\begin{figure}[!h]
  \centering
  \includegraphics[draft=\draft,width=.7\textwidth]{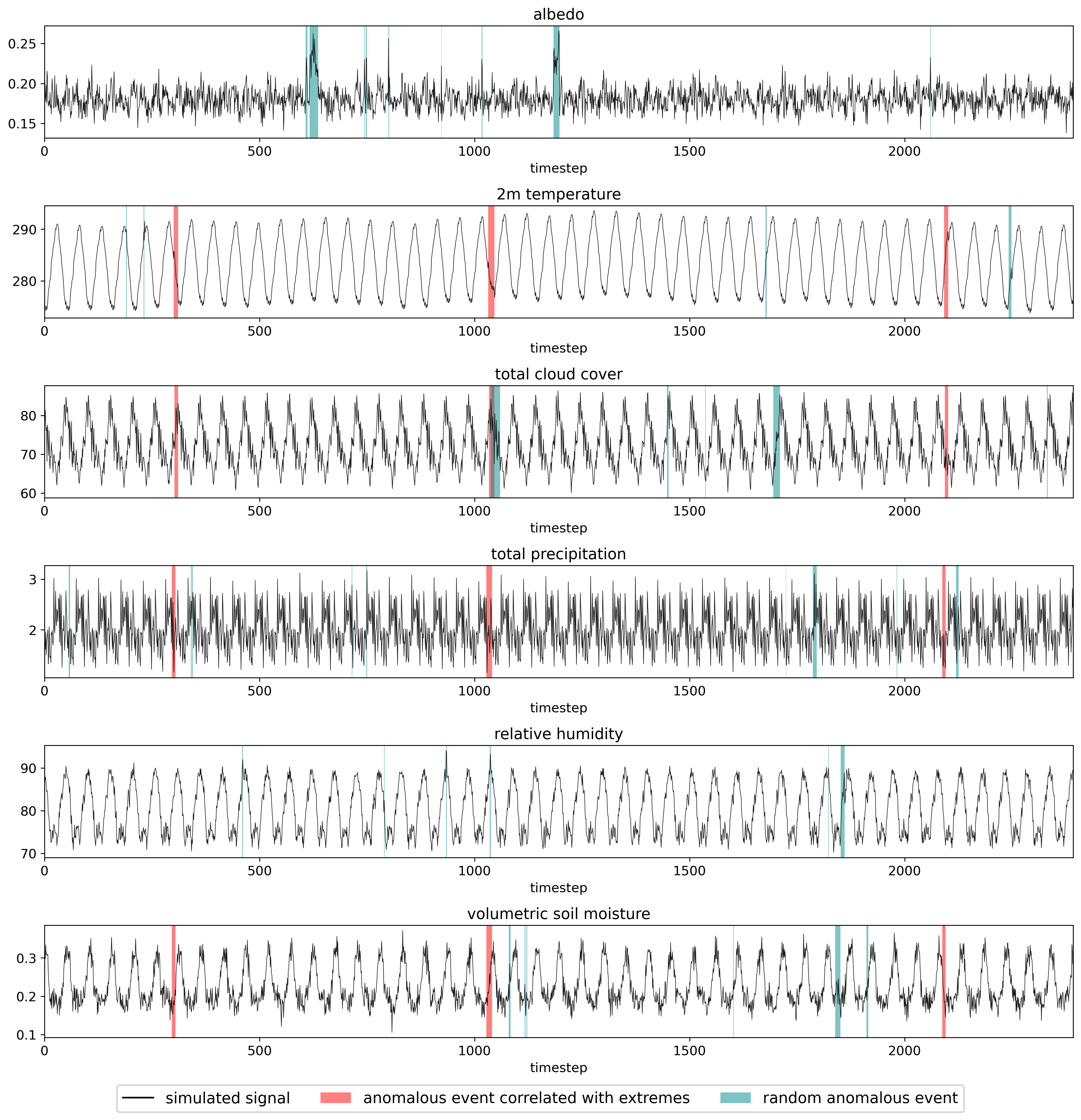}
  \caption{Visualization of the generated signals $\mathbf{\Phi}$ for $6$ different variables from the synthetic CERRA reanalysis described in Table \ref{table:3}. The time series are shown for the location ($\text{lat}=50, \text{lon}=50$). \legendbox{red} are the drivers/anomalies which are correlated with extremes, and \legendbox{teal} are random anomalies.}\label{fig:9}
\end{figure}

\begin{figure}[!h]
  \centering
  \includegraphics[draft=\draft,width=.7\textwidth]{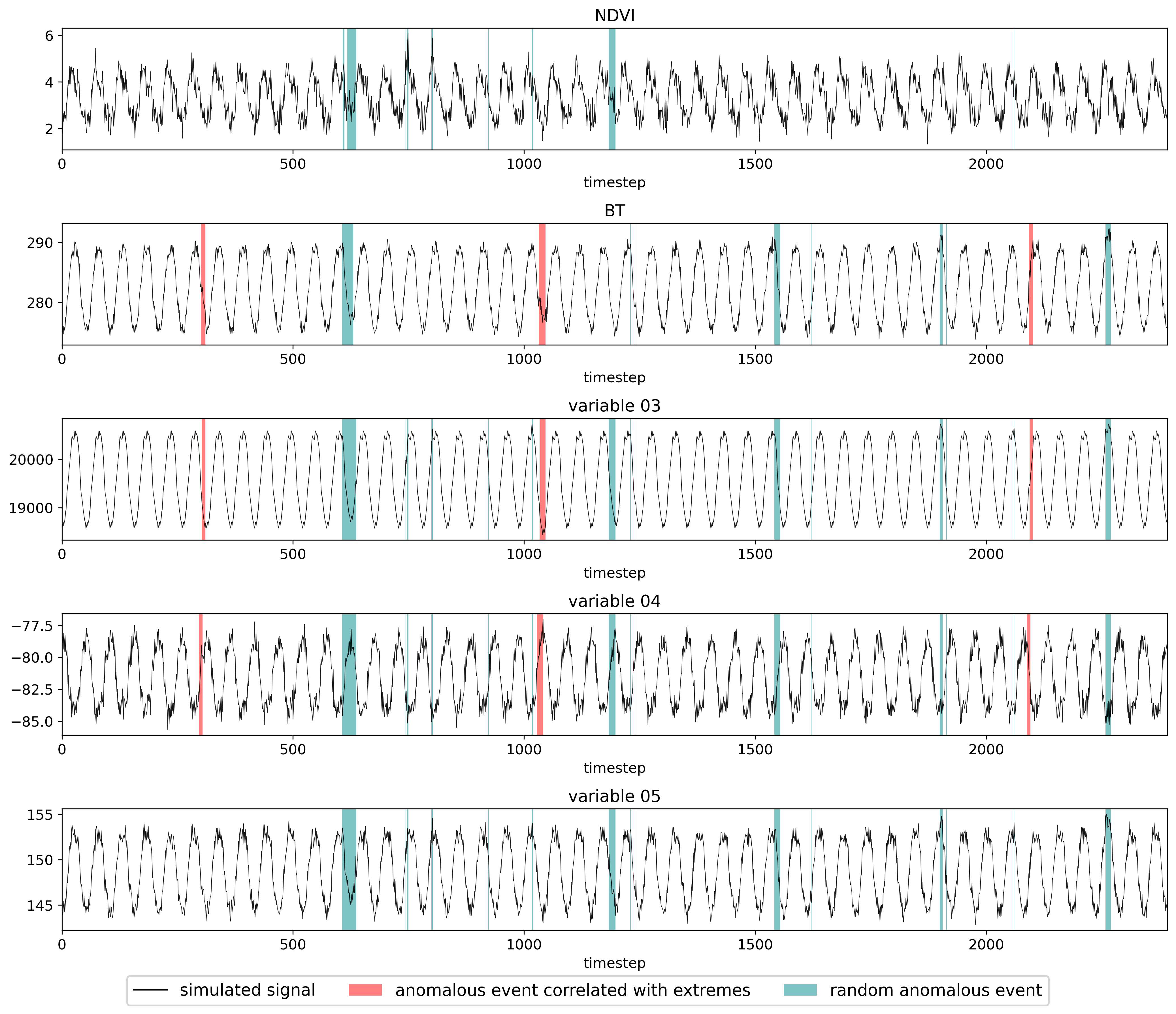}
  \caption{Visualization of the generated signals $\mathbf{\Phi}$ for $5$ different variables from the synthetic NOAA data described in Table \ref{table:4}. The time series are shown for the location ($\text{lat}=50, \text{lon}=50$). \legendbox{red} are the drivers/anomalies which are correlated with extremes, and \legendbox{teal} are random anomalies.}\label{fig:10}
\end{figure}

\begin{figure}[!h]
  \centering
  \includegraphics[draft=\draft,width=.7\textwidth]{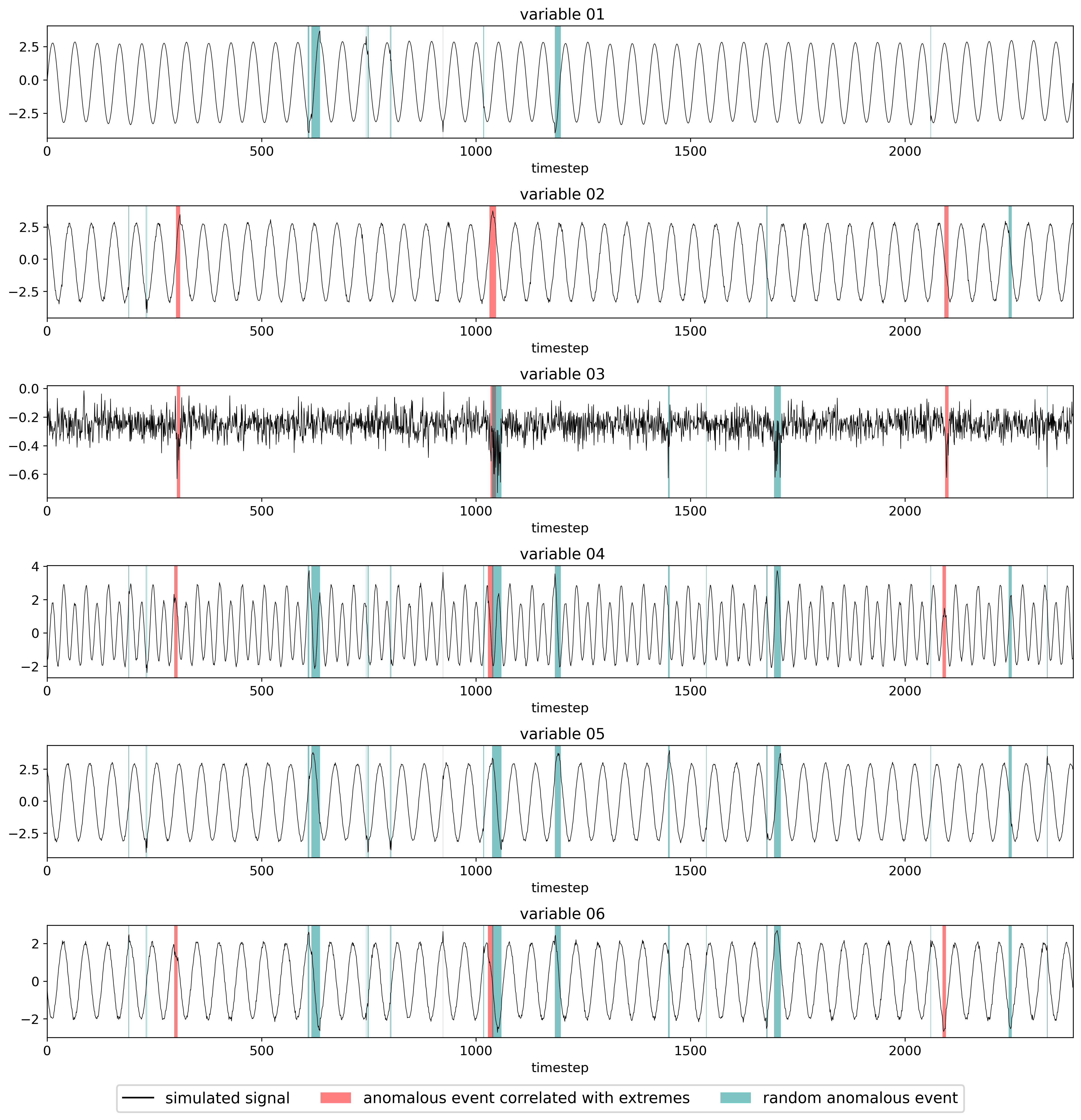}
  \caption{Visualization of the generated signals $\mathbf{\Phi}$ for $6$ different variables from the synthetic artificial data described in Table \ref{table:5}. The time series are shown for the location ($\text{lat}=50, \text{lon}=50$). \legendbox{red} are the drivers/anomalies which are correlated with extremes, and \legendbox{teal} are random anomalies.}\label{fig:11}
\end{figure}

The third dataset (Artificial) does not depend on real-world data but rather consists of basis signals with trigonometric functions where we also add a linear latitudinal gradient (latGrad). Finally, we mask out some regions to exhibit no anomalies i.e., pixels over water surface. Each generated dataset consists of $52\times46$ time steps corresponding to $46$ years of simulated data in $7$-day intervals. Examples of the synthesized dataset can be found in Figs.~\ref{fig:6}-\ref{fig:11}. The results on the synthetic CERRA reanalysis are reported in Table \ref{table:1} and the results on synthetic NOAA and the artificial data are shown Sec.~\ref{sec:9.2} in Tables \ref{table:6} and \ref{table:7}. 

\begin{table}[h!]
\tiny
  \caption{Configurations of the synthetic CERRA reanalysis.}
  \label{table:3}
  \centering
  \tabcolsep=0.6pt\relax
  \begin{tabular}{*{5}l c c c c}
    \toprule
    \multicolumn{9}{l}{Synthetic CERRA reanalysis} \\
    \midrule
    Ind. & Dep. & \multirow{1}{*}{Coupled variables} & Dimension & Extreme events &  $-\triangle T_{a}$ & $+\Delta T_{a}$ & \% Extreme & \% Correlated \\
    variables & variables & \multirow{1}{*}{with extreme} & & & & & & anomalous\\
    \midrule
     6 & 0 & 4 & \{lon=200,           &  \{CubeEvent(n=200, sx=35, sy=35, sz=25),   & 9 & 4 & 1.16 & 1.69 \\
       &   &   & lat=200,             &  RandomWalkEvent(n=1100, s=125),     &   &    &  &   \\
       &   &   & time=$52\times46$\}  &  LocalEvent(n=2600, sz=17),       &      &    &  & \% Random \\
       &   &   &                      &  GaussianEvent(n=340, sx=35, sy=35, sz=25), &  & & & anomalous\\
       &   &   &                      &  OnsetEvent(n=25, sx=17, sy=17, os=0.98)\} & &  &  & 1.32 \\
    \midrule
    Base   & Dependency & Weights & Noise & Random events & $\delta$ & kb & kn & ks \\
    \midrule
    Albedo           &  - & - & WhiteNoise( &  \{CubeEvent(n=320, sx=35, sy=35, sz=25), & - & 0.30 & 0.20 & 0.50 \\
               &   &   & meu=0, &  RandomWalkEvent(n=3000, s=125), & & & & \\
               &   &   & sigma=0.01) &  LocalEvent(n=4000, sz=17), & & & & \\
               &   &   &  &  GaussianEvent(n=300, sx=35, sy=35, sz=25)\} & & & & \\
    
      2m  &  - & -  & RedNoise( &  \{OnsetEvent(n=18, sx=17, sy=17, os=0.98), & +1 &  0.01 & 0.01 & 0.50 \\
        Temperature       &   &   & meu=0, &  RandomWalkEvent(n=1800, s=125),      & & & & \\
               &   &   &  sigma=0.90) &  LocalEvent(n=160, sz=17),      & & & & \\
               &   &   &  &  GaussianEvent(n=350, sx=35, sy=35, sz=25)\}      & & & & \\

      Total  & - & - & LaplaceNoise(& \{CubeEvent(n=300, sx=35, sy=35, sz=25),             & -1 & 0.03 & 0.08 & 0.50 \\
       cloud cover &  & & meu=0,         & RandomWalkEvent(n=2000, s=125),             & & & &  \\
                        &  & &  sigma=0.70,      &  LocalEvent(n=2800, sz=17),             & & & &  \\
                        &  & & lambda=1) & GaussianEvent(n=290, sx=35, sy=35, sz=25)\}              &  &  & &  \\

    Total            &  - & - & WhiteNoise( &  \{CubeEvent(n=320, sx=35, sy=35, sz=25), & -1 & 0.07 & 0.20 & 0.50 \\
     precipitation          &   &   & meu=0, &  RandomWalkEvent(n=3000, s=125), & & & & \\
               &   &   & sigma=0.04) &  LocalEvent(n=4000, sz=17), & & & & \\
               &   &   &  &  GaussianEvent(n=300, sx=35, sy=35, sz=25)\} & & & & \\
    
      Relative  &  - & -  & CauchyNoise( &  \{OnsetEvent(n=18, sx=17, sy=17, os=0.98), & - &  0.06 & 0.06 & 0.50 \\
        humedity       &   &   & meu=0, &  RandomWalkEvent(n=1800, s=125),      & & & & \\
               &   &   &  sigma=0.7) &  LocalEvent(n=160, sz=17),      & & & & \\
               &   &   &  &  GaussianEvent(n=350, sx=35, sy=35, sz=25)\}      & & & & \\

      Volumetric  & - & - & WhiteNoise(& \{CubeEvent(n=300, sx=35, sy=35, sz=25),             & -1 & 0.10 & 0.10 & 0.50 \\
                    soil moisture  &  & & meu=0.,         & RandomWalkEvent(n=2000, s=125),             & & & &  \\
                        &  & &  sigma=.017)      &  LocalEvent(n=2800, sz=17),             & & & &  \\
                        &  & & & GaussianEvent(n=290, sx=35, sy=35, sz=25)\}              &  &  & &  \\                              
                        
    \bottomrule
  \end{tabular}
\end{table}

\begin{table}
\tiny
  \caption{Configurations of the synthetic NOAA remote sensing.}
  \label{table:4}
  \centering
  \tabcolsep=0.6pt\relax
  \begin{tabular}{*{5}l c c c c}
    \toprule
    \multicolumn{9}{l}{Synthetic NOAA} \\
    \midrule
    Ind. & Dep. & \multirow{1}{*}{Coupled variables} & Dimension & Extreme events &  $-\triangle T_{a}$ & $+\triangle T_{a}$ & \% Extreme & \% Correlated \\
    variables & variables & \multirow{1}{*}{with extreme} & & & & & & anomalous\\
    \midrule
     2 & 3 & 4 & \{lon=200,           &  \{CubeEvent(n=200, sx=35, sy=35, sz=25),   & 9 & 4 & 0.79 & 1.02 \\
       &   &   & lat=200,             &  RandomWalkEvent(n=1100, s=125),            &  & \\
       &   &   & time=$52\times46$\}  &  LocalEvent(n=2600, sz=17),                 &  & & & \% Random\\
       &   &   &                      &  GaussianEvent(n=340, sx=35, sy=35, sz=25), &  & & & anomalous\\
       &   &   &                      &  OnsetEvent(n=25, sx=17, sy=17, os=0.98)\}  &  & & & 1.76\\
    \midrule
    Base   & Dependency & Weights & Noise & Random events & $\delta$ & kb & kn & ks \\
    \midrule
    NDVI &  - & - & WhiteNoise( &  \{CubeEvent(n=320, sx=35, sy=35, sz=25), & - & 0.25 & 0.10 & 0.50 \\
       &   &   & meu=0, &  RandomWalkEvent(n=3000, s=125), & & & & \\
               &   &   & sigma=0.30) &  LocalEvent(n=4000, sz=17), & & & & \\
           &   &   &  &  GaussianEvent(n=300, sx=35, sy=35, sz=25)\} & & & & \\

     BT &  - & -  & WhiteNoise( &  \{CubeEvent(n=300, sx=35, sy=35, sz=25), & +1 &  0.01 & 0.01 & 0.50 \\
         &   &   & meu=0, &  RandomWalkEvent(n=2000, s=125),      & & & & \\
            &   &   &  sigma=0.5) &  LocalEvent(n=2800, sz=17),      & & & & \\
            &   &   &  &  GaussianEvent(n=290, sx=35, sy=35, sz=25)\}      & & & & \\

     -     &  Quadratic & Norm & WhiteNoise( & - & -1 & 0.01 & 0.01 & 0.50 \\
               & Coupling()  & Weight()  & meu=0, & & & & & \\
               &   &   & sigma=0.65) &   & & & & \\
               &   &   &  &  & & & & \\
    
    -   &  Linear & Laplace  & WhiteNoise( &  - & - &  0.01 & 0.01 & 0.50 \\
               &  Coupling() &  Weight() & meu=0, &   & & & & \\
               &   &   &  sigma=.065) &     & & & & \\
               &   &   &  &   & & & & \\

    -   & Linear & Laplace & WhiteNoise(& - & +1 & 0.01 & 0.01 & 0.50 \\
                        & Coupling() & Weight() & meu=0.,  &  & & & &  \\
                        &  & &  sigma=.065)      &   & & & &  \\
                        &  & & &              &  &  & &  \\                              
                        
    \bottomrule
  \end{tabular}
\end{table}

\begin{table}
\tiny
  \caption{Configurations of the synthetic artificial data.}
  \label{table:5}
  \centering
  \tabcolsep=0.6pt\relax
  \begin{tabular}{*{5}l c c c c}
    \toprule
    \multicolumn{9}{l}{Synthetic artificial} \\
    \midrule
    Ind. & Dep. & \multirow{1}{*}{Coupled variables} & Dimension & Extreme events &  $-\triangle T_{a}$ & $+\triangle T_{a}$ & \% Extreme & \% Correlated \\
    variables & variables & \multirow{1}{*}{with extreme} & & & & & & anomalous\\
    \midrule
     3 & 3 & 4 & \{lon=200,           &  \{CubeEvent(n=200, sx=35, sy=35, sz=25),   & 9 & 4 & 1.24 & 1.81 \\
       &   &   & lat=200,             &  RandomWalkEvent(n=1100, s=125),            &  & \\
       &   &   & time=$52\times46$\}  &  LocalEvent(n=2600, sz=17),                 &  & & & \% Random\\
       &   &   &                      &  GaussianEvent(n=340, sx=35, sy=35, sz=25), &  & & & anomalous\\
       &   &   &                      &  OnsetEvent(n=25, sx=17, sy=17, os=0.98)\}  &  & & & 2.93\\
    \midrule
    Base   & Dependency & Weights & Noise & Random events & $\delta$ & kb & kn & ks \\
    \midrule
    SineBase( &  - & - & RedNoise( &  \{CubeEvent(n=320, sx=35, sy=35, sz=25), & - & 0.35 & 0.35 & 0.35 \\
    shift=0,    &   &   & meu=0, &  RandomWalkEvent(n=3000, s=125), & & & & \\
    amp=3,           &   &   & sigma=0.20) &  LocalEvent(n=4000, sz=17), & & & & \\
    nOsc=46,       &   &   &  &  GaussianEvent(n=300, sx=35, sy=35, sz=25)\} & & & & \\
        latGrad=True)     &   &   &  &  & & & & \\

     CosineBase( &  - & -  & LaplaceNoise( &  \{OnsetEvent(n=18, sx=17, sy=17, os=0.98), & +1 &  0.35 & 0.35 & 0.35 \\
      shift=0,    &   &   & meu=0, &  RandomWalkEvent(n=1800, s=125),      & & & & \\
      amp=3,       &   &   &  sigma=0.08, &  LocalEvent(n=160, sz=17),      & & & & \\
     nOsc=46,       &   &   & lambda=1) &  GaussianEvent(n=350, sx=35, sy=35, sz=25)\}      & & & & \\
     latGrad=True),       &   &   & & & & & & \\

      ConstantBase( & - & - & WhiteNoise(& \{CubeEvent(n=300, sx=35, sy=35, sz=25),             & -1 & 0.90 & 0.90 & 0.90 \\
       const=0,                  &  & & meu=0,         & RandomWalkEvent(n=2000, s=125),             & & & &  \\
       latGrad=True)                 &  & &  sigma=0.07)      &  LocalEvent(n=2800, sz=17),             & & & &  \\
                        &  & & & GaussianEvent(n=290, sx=35, sy=35, sz=25)\}              &  &  & &  \\

        -  &  Quadratic & Norm & WhiteNoise( & -  & -1 & 0.35 & 0.35 & 0.35 \\
               & Coupling()  & Weight()  & meu=0, &   & & & & \\
               &   &   & sigma=0.65) &   & & & & \\
               &   &   &  & & & & & \\
    
      - &  Linear & Laplace  & WhiteNoise( & - & - &  0.35 & 0.35 & 0.35 \\
               &  Coupling() &  Weight() & meu=0, &   & & & & \\
               &   &   &  sigma=.065) &       & & & & \\
               &   &   &  &       & & & & \\

     -  & Linear & Laplace & WhiteNoise(& -  & -1 & 0.35 & 0.35 & 0.35 \\
                        & Coupling() & Weight() & meu=0.,         &      & & & &  \\
                        &  & &  sigma=.065)      &              & & & &  \\
                        &  & & &             &  &  & &  \\                              
                        
    \bottomrule
  \end{tabular}
\end{table}



\clearpage

\clearpage

\section{Additional results on the synthetic data}
\label{sec:9.2}
In Tables \ref{table:6} and \ref{table:7}, we report the results on the synthetic NOAA remote sensing and artificial data described in Sec.~\ref{sec:9.1}. We noticed a drop of performance for all models for the synthetic artificial data (Table \ref{table:7}). This explained as the ratio of anomalies is higher than the other two datasets. Second, we used for this artificial dataset red noise and quadratic coupling to generate dependent base signals. This makes this dataset harder for training.
SimpleNet exhibits a dramatic dropped out of performance when it is tested on these two synthetic datasets. This illustrates that the model performance is highly dependent on the dataset and the backbone the model was trained on. Our model still outperforms all baselines on these datasets.

\begin{table}[h!]
  \caption{Driver detection results on the synthetic NOAA remote sensing. The best performance on each metric is highlighted in a bold text. \gray{($\pm$)} denotes the standard deviation for 3 runs.}
  \label{table:6}
  \centering
  \small
  \tabcolsep=2.5pt\relax
  \setlength\extrarowheight{2.5pt}
  \begin{tabular}{c r *{6}l}
    \toprule
    & & \multicolumn{3}{c}{Validation} & \multicolumn{3}{c}{Testing} \\
    \cmidrule{3-5} \cmidrule{6-8}
    & Algorithm & F1-score ($\uparrow$) & IoU ($\uparrow$) & OA ($\uparrow$) & F1-score ($\uparrow$) & IoU ($\uparrow$) & OA ($\uparrow$) \\
    \midrule
    & Naive & 47.47 & 31.12 & 99.17 & 51.07 & 34.29 & 98.95\\
    \midrule
   \multirow{3}{*}{\rotatebox[origin=c]{90}{\small{One-Class}}} & OCSVM \cite{OCSVM} & 37.94\gray{$\pm$13.11} & 24.21\gray{$\pm$9.93} & 98.60\gray{$\pm$0.30} & 39.25\gray{$\pm$12.57} &  25.17\gray{$\pm$9.67} & 98.26\gray{$\pm$0.37} \\
    & IF \cite{IF} & 44.14\gray{$\pm$1.18} & 28.34\gray{$\pm$0.97} & 98.68 \gray{$\pm$0.01} & 45.13\gray{$\pm$1.30} & 29.15\gray{$\pm$1.09} & 98.49\gray{$\pm$0.01}\\
   & SimpleNet \cite{SimpleNet} & 56.94\gray{$\pm$0.20} & 39.80\gray{$\pm$0.19} & 99.29\gray{$\pm$0.02} & 57.24\gray{$\pm$0.31} & 40.10\gray{$\pm$0.30} & 99.08\gray{$\pm$0.02} \\
    \midrule
    \multirow{2}{*}{\rotatebox[origin=c]{90}{\small{Rec.}}} & STEALNet \cite{STEALNet} & 56.97\gray{$\pm$0.86} & 39.84\gray{$\pm$0.84} & 99.06\gray{$\pm$0.01} & 58.65\gray{$\pm$0.88} & 41.50\gray{$\pm$0.88} & 98.83\gray{$\pm$0.02} \\
   & UniAD \cite{UniAD} & 45.24\gray{$\pm$3.58} & 29.30\gray{$\pm$2.94} & 98.65\gray{$\pm$0.21} & 46.99\gray{$\pm$4.21} & 30.81\gray{$\pm$3.54} & 98.36\gray{$\pm$0.29} \\
    \midrule
   \multirow{3}{*}{\rotatebox[origin=c]{90}{\small{MIL}}} & DeepMIL \cite{DeepMIL} & 71.77\gray{$\pm$0.38} &  55.97\gray{$\pm$0.46} & 99.55\gray{$\pm$0.01} & 72.02\gray{$\pm$0.31} & 56.28\gray{$\pm$0.38} & 99.42\gray{$\pm$0.01} \\
   & ARNet \cite{ARNet} & 71.06\gray{$\pm$0.46} & 55.11\gray{$\pm$0.56} & 99.53\gray{$\pm$0.01} & 71.43\gray{$\pm$0.51} & 55.56\gray{$\pm$0.62} & 99.40\gray{$\pm$0.01} \\
   
   & RTFM \cite{RTFM} & 60.30\gray{$\pm$0.26} & 43.16\gray{$\pm$0.27} & 98.92\gray{$\pm$0.02} & 61.91\gray{$\pm$0.21} & 44.83\gray{$\pm$0.23} & 98.70\gray{$\pm$0.02} \\
   \midrule
    & Ours$^*$~~~~~~~ & \textbf{81.93}\gray{$\pm$0.23} & \textbf{69.40}\gray{$\pm$0.36} & \textbf{99.69}\gray{$\pm$0.01} & \textbf{82.55}\gray{$\pm$0.25} & \textbf{70.28}\gray{$\pm$0.36} & \textbf{99.61}\gray{$\pm$0.01} \\
    & \logo Ours$^\dag$~~~~~~~ & 81.44\gray{$\pm$0.47} & 68.70\gray{$\pm$0.66} & \textbf{99.69}\gray{$\pm$0.01} & 82.07\gray{$\pm$0.45} & 69.59\gray{$\pm$0.65} & 99.60\gray{$\pm$0.01} \\
    \bottomrule    
\multicolumn{4}{l}{\footnotesize{$^*$Video Swin Transformer backbone \cite{liu2022video}}} & \multicolumn{3}{l}{\footnotesize{$^\dag$Mamba backbone \cite{gu2023mamba}}}\\
\end{tabular}
\end{table}

\begin{table}[h!]
  \caption{Driver detection results on the synthetic artificial data. The best performance on each metric is highlighted in a bold text. \gray{($\pm$)} denotes the standard deviation for 3 runs.}
  \label{table:7}
  \centering
  \small
  \tabcolsep=2.5pt\relax
  \setlength\extrarowheight{2.5pt}
  \begin{tabular}{c r *{6}l}
    \toprule
    & & \multicolumn{3}{c}{Validation} & \multicolumn{3}{c}{Testing} \\
    \cmidrule{3-5} \cmidrule{6-8}
    & Algorithm & F1-score ($\uparrow$) & IoU ($\uparrow$) & OA ($\uparrow$) & F1-score ($\uparrow$) & IoU ($\uparrow$) & OA ($\uparrow$) \\
    \midrule
    & Naive & 46.89 & 30.63 & 98.49 & 51.09 & 34.31 & 98.28 \\
    \midrule
   \multirow{3}{*}{\rotatebox[origin=c]{90}{\small{One-Class}}} & OCSVM \cite{OCSVM} & 27.96\gray{$\pm$1.76} & 16.27\gray{$\pm$1.20} & 97.77\gray{$\pm$0.11} & 33.05\gray{$\pm$1.29} & 19.80\gray{$\pm$0.93} & 97.42\gray{$\pm$0.12}\\
    & IF \cite{IF} & 28.55\gray{$\pm$1.21} & 16.66\gray{$\pm$0.82} & 97.63\gray{$\pm$0.02} & 34.15\gray{$\pm$1.25} & 20.60\gray{$\pm$0.91} & 97.19\gray{$\pm$0.03}\\
   & SimpleNet \cite{SimpleNet} & 34.57\gray{$\pm$0.54} & 20.90\gray{$\pm$0.39} & 97.56\gray{$\pm$0.04} & 41.18\gray{$\pm$0.35} & 25.93\gray{$\pm$0.28} & 97.94\gray{$\pm$0.09} \\
    \midrule
    \multirow{2}{*}{\rotatebox[origin=c]{90}{\small{Rec.}}} & STEALNet \cite{STEALNet} & 56.40\gray{$\pm$0.78} & 39.28\gray{$\pm$0.76} & 98.33\gray{$\pm$0.03} & 58.23\gray{$\pm$0.92} & 41.09\gray{$\pm$0.92} & 98.11\gray{$\pm$0.04} \\
   & UniAD \cite{UniAD} & 49.48\gray{$\pm$1.60} & 32.88\gray{$\pm$1.41} & 97.66\gray{$\pm$0.11} & 52.49\gray{$\pm$1.25} & 35.60\gray{$\pm$1.16} & 97.57\gray{$\pm$0.09} \\
    \midrule
   \multirow{3}{*}{\rotatebox[origin=c]{90}{\small{MIL}}} & DeepMIL \cite{DeepMIL} & 20.18\gray{$\pm$23.67} & 13.38\gray{$\pm$16.45} & 71.33\gray{$\pm$20.53} & 18.75\gray{$\pm$21.34} & 12.04\gray{$\pm$14.39} & 71.08\gray{$\pm$20.48} \\
   & ARNet \cite{ARNet} & 48.98\gray{$\pm$5.30} & 32.59\gray{$\pm$4.55} & 98.82\gray{$\pm$0.06} & 44.75\gray{$\pm$5.53} & 28.98\gray{$\pm$4.49} & 98.53\gray{$\pm$0.07} \\
   & RTFM \cite{RTFM} & 59.90\gray{$\pm$0.31} & 42.75\gray{$\pm$0.32} & 98.27\gray{$\pm$0.04} & 61.41\gray{$\pm$0.46} & 44.31\gray{$\pm$0.48} & 98.07\gray{$\pm$0.04} \\
   \midrule
    & Ours$^*$~~~~~~~ & \textbf{70.20}\gray{$\pm$0.43} & \textbf{54.08}\gray{$\pm$0.51} & \textbf{98.90}\gray{$\pm$0.03} & \textbf{70.33}\gray{$\pm$0.71} & \textbf{54.24}\gray{$\pm$0.84} & \textbf{98.74}\gray{$\pm$0.08} \\
    & \logo Ours$^\dag$~~~~~~~ & 66.63\gray{$\pm$5.41} & 50.19\gray{$\pm$5.91} & 98.81\gray{$\pm$0.12} & 67.64\gray{$\pm$5.95} & 51.39\gray{$\pm$6.60} & 98.71\gray{$\pm$0.18} \\
    \bottomrule
\multicolumn{4}{l}{\footnotesize{$^*$Video Swin Transformer backbone \cite{liu2022video}}} & \multicolumn{3}{l}{\footnotesize{$^\dag$Mamba backbone \cite{gu2023mamba}}}\\
  \end{tabular}
\end{table}

\section{Ablation studies}
\label{sec:9.3}
In this section, we do ablation analyses on different aspects of the proposed model. All experiments are done on the validation set of synthetic CERRA reanalysis and evaluated with the F1-scores for drivers anomalies and extreme events detection.

\subsection{Quantization layer}
\label{sec:9.3.1}
In Table \ref{table:8}, we study the performance of our model with different vector quantization algorithms. The first row \textbf{Threshold (Tanh)} represents a simple straight through estimator. For this quantization, we first map the input into a scalar value followed by a Tanh activation. Then we set positive values to be anomalies ($q=1$).
In \textbf{Random Quantization (RQ)} \cite{RQ} the input is projected with a randomly initialized weights and then compared with a randomly initialized codes. \textbf{Vector Quantization (VQ)} is the standard quantizer which uses an Euclidean distance \cite{VQ} or a cosine similarity \cite{VQ_cosine}. We further add an orthogonality loss for VQ similar to \cite{VQ_ortho}. \textbf{Finite Scalar Quantization (FSQ)} maps the input into a bounded scalar value. The code is then assigned based on the rounded value in the discrete space. As seen, \textbf{Lockup free quantization (LFQ)} \cite{LFQ} performs the best. We speculate that LFQ does not need to learn the code vectors which simplifies the task of quantization.

\begin{table}[!h]
  \caption{Ablation studies on the quantization layer. The metric is F1-score on the driver/\olive{extreme} detection.}
  \label{table:8}
  \centering
  \tabcolsep=6.pt\relax
  \setlength\extrarowheight{0pt}
  \begin{tabular}{l l}
      \toprule
      Quantization Layer & F1-score ($\uparrow$) \\
      \midrule
     Threshold (Tanh) & 70.07 / \olive{84.98} \\
      RQ (Euclidean distance) \cite{RQ} & 71.52 / \olive{87.23} \\
      VQ (Cosine similarity) \cite{VQ_cosine} & 78.83 / \olive{88.98} \\
      VQ (Cosine similarity + Orthogonality) \cite{VQ_cosine, VQ_ortho} & 78.54 / \olive{86.83} \\
      VQ (Euclidean distance) \cite{VQ} & 77.42 / \olive{87.27} \\
      VQ (Euclidean distance + Orthogonality) \cite{VQ, VQ_ortho} & 79.26 / \olive{88.28} \\
      FSQ \cite{FSQ} & 76.57 / \olive{88.65} \\
      LFQ \cite{LFQ} & \textbf{82.78 / \olive{92.45}} \\
      \bottomrule
  \end{tabular}
\end{table}

\subsection{Objective function for quantization}
\label{sec:9.3.2}
This part studies the loss function $\mathcal{L}_{(quantize)}$ used in Eq.~\eqref{eq:4}. We denote the term $\lVert \Z_l - \text{sg}(\sign(\Z_l)) \rVert_{2}^{2}$ as $\mathcal{L}_{(commit)}$, $H[\EE\bigl(\sign(\Z_l)\bigr)]$ as $\mathcal{L}_{(div)}$, and $\EE[H\bigl(\sign(\Z_l)\bigr)]$ as $\mathcal{L}_{(ent)}$. As seen from Table \ref{table:9}, $\mathcal{L}_{(commit)}$ is essential for training to prevent the input for the quantization layer from growing. While $\mathcal{L}_{(ent)}$ and $\mathcal{L}_{(div)}$ (rows 4, 5, and 7) improve the results compared to the model with $\mathcal{L}_{(commit)}$ (second row).

\begin{table}[!h]
  \caption{Ablation studies on the loss function $\mathcal{L}_{(quantize)}$ in Eq.~\eqref{eq:4}. The metric is F1-score on the anomaly/\olive{extreme} detection.}
  \label{table:9}
  \centering
  \tabcolsep=6.pt\relax
  \setlength\extrarowheight{0pt}
  \begin{tabular}{*{5}l}
      \toprule
      $\mathcal{L}_{(ent)}$ & $\mathcal{L}_{(commit)}$ & $\mathcal{L}_{(div)}$ & F1-score ($\uparrow$) \\
      \midrule
      \cmark & \xmark & \xmark & 00.00 ~/ \olive{00.00}  \\
      \xmark & \cmark & \xmark & 81.86 / \olive{91.18} \\
      \xmark & \xmark & \cmark & 00.00 ~/ \olive{00.00} \\
      \cmark & \cmark & \xmark & 81.87 / \olive{91.48} \\
      \xmark & \cmark & \cmark & 82.26 / \olive{92.30} \\
      \cmark & \xmark & \cmark & 00.00 ~/ \olive{00.00} \\
      \cmark & \cmark & \cmark & \textbf{82.78 / \olive{92.45}} \\
      \bottomrule
  \end{tabular}
\end{table}

\subsection{Objective function for extreme events prediction.}
\label{sec:9.3.3}

Due to class imbalanced, we add class weighting applied to the binary cross entropy loss $\mathcal{L}_{(extreme)}$ to predict extremes. The weighting is based on the logarithm of the inverse square roots of the relative class frequencies in the batch. The weighted loss function achieves better results as shown in Table \ref{table:10}.

\begin{table}[!h]
  \caption{Ablation studies on the weighted loss function $\mathcal{L}_{(extreme)}$ in Eq.~\eqref{eq:3}. The metric is F1-score on the driver/\olive{extreme} detection.}
  \label{table:10}
  \centering
  \tabcolsep=6.pt\relax
  \setlength\extrarowheight{0pt}
  \begin{tabular}{*{2}l}
      \toprule
      $\mathcal{L}_{(extreme)}$ & F1-score ($\uparrow$) \\
      \midrule
      unweighted $\mathcal{L}_{(extreme)}$ & 81.14 / \olive{90.91} \\
      weighted $\mathcal{L}_{(extreme)}$ & \textbf{82.78 / \olive{92.45}} \\
      \bottomrule
  \end{tabular}
\end{table}

\subsection{Feature extractor ($f_{\theta}$)}
\label{sec:9.3.4}

We evaluate three types of feature extractors, 3D CNN, Video Swin Transformer \cite{liu2022video}, and SSM Vision Mamba \cite{gu2023mamba} with local scans. All models are trained from scratch. To build the 3D CNN model, we replaces the attention blocks in Swin Transformer with 3D convolutions and keep the overall architecture. 
For Mamba, we use local scans. Vision Mamba backbones replace the attention module with a linear selective state space model which provides an efficient alternative for Video Swin Transformer \cite{zhu2024vision, pei2024efficientvmamba, chen2024video, li2024videomamba, huang2024localmamba, park2024videomamba, hu2024zigma}.
From Table \ref{table:11} we notice that Swin Transformer achieves better results compared to 3D CNN. The parameters are also less compared to 3D CNN.
The results also indicate that using Mamba instead of Swin Transformer achieves similar or better results when less parameters are used. In contrast to Swin Transformer, Mamba can be scaled to the global scale. However, we have used Swin Transformer in all experiments to have a fair comparison with the baselines due to its commonality as a backbone.

\begin{table}[!h]
  \caption{Ablation study on the backbone $f_{\theta}$ used for feature extraction. The metric is F1-score on the driver/\olive{extreme} detection.}
  \label{table:11}
  \centering
  \tabcolsep=6.pt\relax
  \setlength\extrarowheight{0pt}
  \begin{tabular}{*{4}l}
      \toprule
      Backbone $f_{\theta}$ & Hidden dimension ($K$) & Parameters & F1-score ($\uparrow$) \\
      \midrule
      3D CNN & 8 & 63k & 57.15 / \olive{91.21} \\
      Video Swin Transformer & 8 & 19k & 81.22 / \olive{91.16} \\
      Mamba & 8 & 15k & 82.15 / \olive{90.18} \\
      \midrule
      3D CNN & 16 & 250k & 70.93 / \olive{93.75} \\
      Video Swin Transformer & 16 & 62k& 82.78 / \olive{92.45} \\
      Mamba & 16 & 56k & 83.45 / \olive{93.12} \\
      \midrule
      3D CNN & 32 & 998k & 84.95 / \olive{93.43} \\
      Video Swin Transformer & 32 & 230k & 84.14 / \olive{93.12} \\
      Mamba & 32 & 214k & 84.00 / \olive{93.43} \\
      \bottomrule
  \end{tabular}
\end{table}

Table \ref{table:11} shows that increasing the model parameters still does not show a sign of overfitting.

\subsection{Lossy vs lossless driver detection}
\label{sec:9.3.5}
We trained the model without the quantization layer. In other words, we remove the driver detection step and trained the model to predict extreme droughts directly from the extracted features with one classification head. The reason behind this experiment is to check the information loss through the quantization/driver detection step. Table \ref{table:12} shows the improvement in performance on both the synthetic and real-world datasets. Adding the driver detection step identifies the related drivers to the events. However, at the cost of a slight decrease in accuracy on extreme events prediction.

\begin{table}[!h]
  \caption{Ablation study on the driver detection step. The metric is F1-score on \olive{extreme} detection where  ($\Delta\text{F1}= \text{F1}_{(\text{without~quantization})} - \text{F1}_{(\text{with~quantization})}$).}
  \label{table:12}
  \centering
  \tabcolsep=6.pt\relax
  \setlength\extrarowheight{0pt}
  \begin{tabular}{*{2}l}
      \toprule
      Dataset & $\Delta$F1-score ($\downarrow$) \\
      \midrule
      Synthetic CERRA Reanalysis & \olive{+0.96\%} \\
      ERA5-Land (EUR-11) & \olive{+1.93\%} \\
      \bottomrule
  \end{tabular}
\end{table}

\subsection{Weighting parameters in the main objective function}
\label{sec:9.3.6}
      
\begin{table}[!h]
  \caption{Sensitivity studies on the weighting parameters used in the main loss function in Eq.~\eqref{eq:6}. The metric is mean F1-score on the driver/\olive{extreme} detection for $3$ random seeds.}
  \label{table:13}
  \centering
  \tabcolsep=2.2pt\relax
  \setlength\extrarowheight{0pt}
  \begin{tabular}{*{11}l}
  \multicolumn{2}{l}{(a)} & & \multicolumn{2}{l}{(b)} & & \multicolumn{2}{l}{(c)} & & \multicolumn{2}{l}{(d)} \\
      \cmidrule[\heavyrulewidth]{1-2} \cmidrule[\heavyrulewidth]{4-5} \cmidrule[\heavyrulewidth]{7-8} \cmidrule[\heavyrulewidth]{10-11}
      $\lambda_{(driver)}$ & F1-score ($\uparrow$) & & $\lambda_{(commit)}$ & F1-score ($\uparrow$) & & $\lambda_{(ent)}$ & F1-score ($\uparrow$) & & $\lambda_{(div)}$ & F1-score ($\uparrow$) \\
      \cmidrule[\heavyrulewidth]{1-2} \cmidrule[\heavyrulewidth]{4-5} \cmidrule[\heavyrulewidth]{7-8} \cmidrule[\heavyrulewidth]{10-11}
      1 & 02.45 / \olive{58.07} & & 1.0 & 79.81 / \olive{92.02} & & 0.01 & 82.52 / \olive{92.61} & & 0.01 & 77.51 / \olive{91.71}\\
      10 & 03.34 / \olive{75.73} & & 1.5 & 81.43 / \olive{91.96} & & 0.1 & \textbf{82.78 / \olive{92.45}} & & 0.1 & \textbf{82.78 / \olive{92.45}}\\
      100 & \textbf{82.78 / \olive{92.45}} & & 2.0 & 82.27 / \olive{92.11} & & 1.0 & 82.72 / \olive{92.46} & & 0.5 & 82.48 / \olive{92.21} \\
      200 & 80.85 / \olive{91.69} & & 3.0 & \textbf{82.78 / \olive{92.45}} \\
      1000 & 79.94 / \olive{86.44} & & 4.0 & 82.76 / \olive{91.89} \\
      \cmidrule[\heavyrulewidth]{1-2} \cmidrule[\heavyrulewidth]{4-5} \cmidrule[\heavyrulewidth]{7-8} \cmidrule[\heavyrulewidth]{10-11}
  \end{tabular}
\end{table}

We study the role of the weighting parameters in the main loss function in Eq.~\eqref{eq:6}. We can observe from Table \ref{table:13} that $\lambda_{(driver)}$ plays a crucial role in anomaly detection. Having $\lambda_{(driver)}$ small makes the model less constrained and reduces supervision on where extremes were reported in the training data. Bigger values of $\lambda_{(driver)}$ make the model more constrained to identify drivers near or in overlap with the extremes and thus reduce the overall performance. We can also observe that the model is less sensitive to $\lambda_{(ent)}$ compared to other parameters. We selected the final default weighting parameters in the experiments based on the average performance on both driver and extreme events prediction. 

\subsection{Objective functions}
\label{sec:9.3.7}

\begin{figure}[h!]
  \centering
  \includegraphics[draft=\draft, width=.9\textwidth]{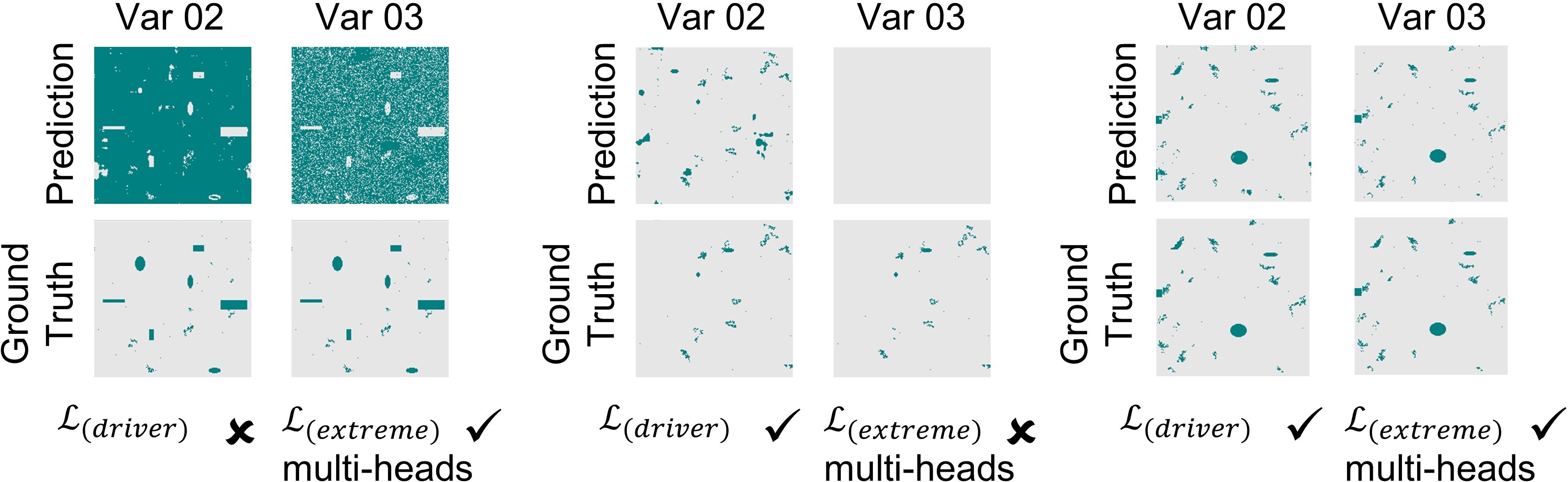}
  \caption{
  Supplementary to the ablation study in Table \ref{table:2}.
  }\label{fig:30}
\end{figure}

Without the loss $\mathcal{L}_{(driver)}$, the detection of drivers/anomalies is not reliable since pixels at regions and intervals where no extreme event occurred can be assigned to $z_{q=1}$ (driver) as well. In case of $\mathcal{L}_{(extreme)}$ multi-heads, we observe that anomalies are identified in a small subset of variables because the network omits some variables if there is a correlation with other variables. Please see Fig.~\ref{fig:30}. In case of a single head such flips occur less often, but they can occur. If the loss $\mathcal{L}_{(driver)}$ is used, such flips cannot occur and the multi-head improves both F1-scores by a large margin. When comparing rows 2 and 4 in Table \ref{table:2} (a), there is a slight decrease in extreme prediction but a large improvement in anomaly detection. Note that there is always a trade-off between extreme and anomaly detection. The anomalies generate a bottleneck of information. When more information goes through the bottleneck the better the extreme prediction gets. Without any anomaly detection, the extreme prediction is best as shown in Table \ref{table:12}, but the increase in F1 score is only moderate. This is also visible in rows 2 and 3 in Table \ref{table:2} (b). Cross-attention improves extreme prediction, but it hurts the detection of anomalies since the information is propagated between the variables.

\section{Results on the real-world reanalysis data}
\label{sec:9.4}

\subsection{Quantitative results}
\label{sec:9.4.1}

It is difficult to quantify the quality of the predicted drivers and anomalies on real-world data. Therefore, we hypothesize that the model can predict extreme agricultural droughts from the identified drivers and anomalies, only if those drivers and anomalies are causally correlated with the extreme events. To verify this, we report in Table \ref{table:14} the prediction accuracy on extreme agricultural droughts. The results show that using the identify drivers as input, the model can predict extreme agricultural droughts across different regions and datasets.

\begin{table}[!h]
  \caption{Quantitative results for \olive{extreme droughts} detection on real-world data. \gray{($\pm$)} denotes the standard deviation for 3 runs.}
  \label{table:14}
  \centering
  \small
  \tabcolsep=2.3pt\relax
  \setlength\extrarowheight{0pt}
  \begin{tabular}{l l *{6}l}
    \toprule
    & & \multicolumn{3}{c}{Validation} & \multicolumn{3}{c}{Testing} \\
    \cmidrule{3-5} \cmidrule{6-8}
    Dataset & Region & F1-score ($\uparrow$) & IoU ($\uparrow$) & OA ($\uparrow$) & F1-score ($\uparrow$) & IoU ($\uparrow$) & OA ($\uparrow$) \\
    \midrule
    CERRA & Europe & \olive{22.31}\gray{$\pm$0.74} & \olive{12.56}\gray{$\pm$0.47} & \olive{90.45}\gray{$\pm$1.05} & \olive{28.63}\gray{$\pm$1.45} & \olive{16.71}\gray{$\pm$0.98} & \olive{89.13}\gray{$\pm$1.44}\\
    \midrule
    ERA5-Land & Europe & \olive{31.87}\gray{$\pm$0.39} & \olive{18.96}\gray{$\pm$0.28} & \olive{95.45}\gray{$\pm$0.25} & \olive{21.52}\gray{$\pm$0.86} & \olive{12.06}\gray{$\pm$0.54} & \olive{95.84}\gray{$\pm$0.21} \\
    & Africa & \olive{22.49}\gray{$\pm$0.42} & \olive{12.67}\gray{$\pm$0.27} & \olive{85.59}\gray{$\pm$1.48} & \olive{18.53}\gray{$\pm$0.48} & \olive{10.21}\gray{$\pm$0.29} & \olive{78.63}\gray{$\pm$1.43} \\
    & North America & \olive{27.39}\gray{$\pm$0.63} & \olive{15.87}\gray{$\pm$0.42} & \olive{93.84}\gray{$\pm$0.13} & \olive{31.74}\gray{$\pm$0.48} & \olive{18.86}\gray{$\pm$0.34} & \olive{89.95}\gray{$\pm$0.42} \\
    & South America & \olive{29.30}\gray{$\pm$1.52} & \olive{17.17}\gray{$\pm$1.04} & \olive{89.31}\gray{$\pm$0.33} & \olive{28.96}\gray{$\pm$1.67} & \olive{16.94}\gray{$\pm$1.13} & \olive{83.99}\gray{$\pm$0.23} \\
    & Central Asia & \olive{20.99}\gray{$\pm$0.43} & \olive{11.73}\gray{$\pm$0.27} & \olive{95.38}\gray{$\pm$0.04} & \olive{25.01}\gray{$\pm$0.11} & \olive{14.29}\gray{$\pm$0.07} & \olive{94.51}\gray{$\pm$0.04}\\
    & East Asia & \olive{18.82}\gray{$\pm$0.72} & \olive{10.39}\gray{$\pm$0.44} & \olive{93.10}\gray{$\pm$0.42} & \olive{25.35}\gray{$\pm$0.20} & \olive{12.58}\gray{$\pm$0.12} & \olive{93.73}\gray{$\pm$0.38}\\
    \bottomrule
  \end{tabular}
\end{table}

The performance depends on the type and ratio of extremes, spatio-temporal resolution, the quality and consistency between the remote sensing and the reanalysis data. F1 scores substantially increase when the threshold on VHI is increased (see Table.~\ref{table:20} in Sec.~\ref{sec:9.4.3}). Note that it is not required to predict all extremes in order to learn specific relations from the predicted events.

\subsection{Robustness}
\label{sec:9.4.2}

To further check the model robustness, we train the model on the same EUR-11 region with $6$ different combinations of physical variables. We anticipate that if a variable is relevant to extremes over specific region i.e., Europe it should appear in all identified sets of variables i.e., we expect soil moisture (swvl1) and evaporation (e) to be always presented as explanatory variables for extremes over Europe. Our experimental results in Table \ref{table:15} confirms this assumption.

\begin{table}[!h]
  \caption{Quantitative results from ERA5-Land EUR-11 data for different combination of physical variables. The metric is F1-score on the \olive{extreme droughts} detection for the validation/test sets.}
  \label{table:15}
  \centering
  \tabcolsep=6.pt\relax
  \setlength\extrarowheight{0pt}
  \begin{tabular}{*{3}l}
      \toprule
      Input variables & Selected variables & F1-score($\uparrow$)\\
      \midrule
      \{t2m, fal, e, tp, stl1, swvl1\} & \{e, swvl1\} & \olive{31.87} / \olive{21.52}  \\
      \{d2m, t2m, sp, e, tp, stl1\} & \{e\} & \olive{32.72} / \olive{23.48}  \\
      \{d2m, fal, sp, e, skt, swvl1\} & \{e, swvl1\} & \olive{32.92} / \olive{22.41}  \\
      \{t2m, sp, e, skt, stl1, swvl1\} & \{e, swvl1\} & \olive{32.33} / \olive{21.61} \\
      \{t2m, fal, sp, skt, stl1, swvl1\} & \{swvl1\} & \olive{24.34} / \olive{16.05}  \\
      \{t2m, e, tp, swvl1\} & \{e, swvl1\} & \olive{30.38} / \olive{21.59}  \\
      \bottomrule
  \end{tabular}
\end{table}

\subsection{Scientific validity}
\label{sec:9.4.3}
Surely, soil temperature is a key factor in the drought processing and soil moisture–temperature feedback \cite{garcia2023soil}.
State variable of the land surface such as albedo (fal/al) and soil temperature (stl1/sot) should be very related to reflectance on the ground and consequently to VHI from remote sensing.
However, our approach indicates that these variables are not informative enough for the model to identify drivers and predict extremes.
To investigate this issue, we conducted 4 more experiments on both CERRA and ERA5-Land where we trained models that take only one variable al/fal or stl as input and predict the extreme events directly without the driver/anomaly detection step (similar to Sec.~\ref{sec:9.3.5}). In all of these experiments, the F1-score was very low. In the next experiment, we increased the threshold for VHI and trained new models to predict extremes directly. The results for the validation set are shown below in Table \ref{table:20}:

\begin{table}[!h]
  \caption{The relation between the definition of extremes from VHI and the model prediction. The metric is F1-score ($\uparrow$) on the \olive{extreme droughts} detection for the validation sets.}
  \label{table:20}
  \centering
  \tabcolsep=6.pt\relax
  \setlength\extrarowheight{0pt}
  \begin{tabular}{*{6}l}
      \toprule
      Input variables & Dataset & Domain & VHI < 26 & VHI < 40 & VHI < 50 \\
      \midrule
      \{stl1\} & ERA5-Land & EUR-11 & \olive{05.67} & \olive{31.53} & \olive{58.36} \\
      \{t2m, fal, e, tp, stl1, swvl1\} & ERA5-Land & EUR-11 & \olive{33.80} & \olive{46.72} & \olive{68.71} \\
      \bottomrule
  \end{tabular}
\end{table}

The first potential reason to consider is that some land surface variables might deviate from the reality. ERA5-Land does not use data assimilation directly. The evolution and the simulated land fields are controlled by the ERA5 atmospheric forcing. Another reason might be that when training only on extremes (VHI < 26), there are not enough samples to learn the relations. Note that VHI is a combination of both TCI and VCI. Most extremes (VHI < 26) might result from a deficiency in both stl/t2m and vsw. This might also explain why stl and albedo cannot be that informative to predict very extreme events. Last row in Table \ref{table:15} shows the result when we discard albedo and soil temperature as input variables from ERA5-Land.

Moreover, state variables of the hydrological cycle in ERA5-land like volumetric soil water variable (swvl1/vsw) has biases \cite{9957057}. One solution to improve the validity of investigation is to use satellite observations for the top layer \cite{5460980}. However, the experiments showed that the model relates soil moisture anomalies with the extremes in VHI and provides reasonable predictions.

\subsection{Spatial Resolution}
\label{sec:9.4.4}

\begin{table}[!h]
  \caption{The impact of spatial resolution on the model prediction. The metric is F1-score on the \olive{extreme droughts} detection for the validation sets.}
  \label{table:21}
  \centering
  \tabcolsep=6.pt\relax
  \setlength\extrarowheight{0pt}
  \begin{tabular}{*{4}l}
      \toprule
      Dataset & Domain & Spatial resolution & F1-score($\uparrow$)\\
      \midrule
      ERA5-Land & EUR-11 & $0.1^\circ$ & \olive{31.87} \\
      ERA5-Land & EUR-11 & $0.2^\circ$ & \olive{30.09} \\
      \bottomrule
  \end{tabular}
\end{table}

\section{Baselines}
\label{sec:9.5}
We evaluate our approach with an interpretable forecasting approach using integrated gradients \cite{Integrated_Gradient} and $8$ baselines from $3$ main related categories in anomaly detection; one-class unsupervised \cite{OCSVM, IF, SimpleNet}, reconstruction-based \cite{STEALNet, UniAD}, and multiple instance learning \cite{DeepMIL, ARNet, RTFM}. Note that these baselines are not directly applicable to the task we addressed in this paper.
We modified and trained all baselines from scratch. For this, we relied on the officially released codes and started from the default hyperparameters.

\textbf{Integrated Gradients \cite{Integrated_Gradient}}~ is an axiomatic attribution method which can be used as a post-hoc method to explain the model prediction. We trained two models that predict extreme events directly from the input variables and then we applied a post-hoc integrated gradients from Pytorch Captum \cite{Captum}. Both models use the same Swin Transformer backbone as our model but without the anomaly detection step. For Integrated Gradients II, we added a cross attention. For these baselines, we compute the gradient only with respect to predicted extremes and computed a different threshold for each variable separately. 

\textbf{Isolated Forest (IF) \cite{IF}}~ is an ensemble of random trees that isolate input instances by randomly selecting abounded splits for features. A shorter averaged path for a recursive partitioning implies an anomalous input. We empirically set the number of estimators to $100$.

\textbf{OCSVM \cite{OCSVM}}~ is a one-class support vector machine solved using stochastic gradient descent. We use a radial basis kernel with $100$ components.

For IF and OCSVM, we extracted multivariate features as the distances between the input variables and then train a model on each input variable separately. We sampled 400k normal data points (locations without extreme flags) for training and defined the expected ratio of anomalous to be roughly equivalent to the ratio of extreme events in the data.

\textbf{SimpleNet \cite{SimpleNet}}~ is categorized as an embedding-based one-class algorithm for anomaly detection. In SimpleNet, a feature extractor is first used to extract local features from normal data followed by a feature adaptor. Then an anomaly feature generator induces anomalous features by adding a Gaussian noise in the feature space. Finally, a discriminator is trained to distinguish between the normal and anomalous generated features. During inference, the feature generator is discarded and the trained discriminator is expected to separate anomalous from normal input features. During inference, we compute the median anomaly score for both normal pixels and pixels with extreme flags. Then, we set a threshold for anomalous pixels based on the average of the two former median anomaly scores. We trained SimpleNet with our pretrained model as a feature extractor and set the feature dimension for feature adaptor to $512$. The extracted features are scaled by $10^{-2}$ and the anomalous feature generator uses a Gaussian noise $\bepsilon\sim\mathcal{N}(0,\sigma=1.5)$. The discriminator composes of two linear layers with a hidden dimension of $96$. We set $th^+$ and $th^{-}$ to $1.0$ and trained with AdamW optimizer for $48$ epochs with a batch size of $2$ and a learning rate of $3 \times 10^{-4}$ with $1\times10^{-5}$ weight decay. 

\textbf{STEALNet \cite{STEALNet}}~ or synthetic temporal anomaly guided end-to-end video anomaly detection. This is a reconstruction-based algorithm. STEALNet was trained to maximize the reconstruction loss for locations with extreme flags and to minimize the reconstruction loss otherwise. We set the dimensions for the auto-encoder as $96$, $128$, and $256$. We trained with a batch size of $8$ for $100$ epochs using Adam optimizer with $1 \times 10^{-3}$ weight decay and a learning rate of $3 \times 10^{-4}$.

For STEALNet and UniAD, we compute during inference the mean reconstruction loss for both normal pixels and pixels with extreme flags. Then, we set a threshold for anomalous pixels based on the average of these two values.

\textbf{DeepMIL \cite{DeepMIL}}~ is a multiple instance learning for anomaly detection in surveillance videos. In MIL, each video is represented as a bag of snippets (instances). We define positive instances as pixels with extreme event flags where the exact information (which variable is anomalous) within the positive instances is unknown. In the original implementation, the ranking is enforced only on the top instance with the highest anomaly score in each bag. We modified the ranking loss to top-k and set $K=100$.

\textbf{ARNet \cite{ARNet}}~ or anomaly regression net is another MIL-based approach. The ranking loss is based on a top-k binary cross entropy. In addition, there is a center loss to reduce the intra-class discriminative features of normal instance. We set $\alpha=400$ and $\lambda_{center}=20$.

\textbf{RTFM \cite{RTFM}}~ or robust temporal feature magnitude learning is a MIL-based algorithm which learns to distinguish between the normal and anomalous scores by selecting the top-k snippets with the highest feature magnitudes. We modified the multi-scale temporal network (MTN) to capture the local and global spatial dependencies between pixels. We set the dimension for MTN to $32$, $K=100$, the margin to separate features to $\times10^{2}$ and $\alpha=1\times10^{-4}$.

We use the same Video Swin Transformer backbone as our model to train DeepMIL, ARNet, and RTFM. During training, we noticed that the ranking loss becomes biases toward one variable when the ranking is computed among all variables, so we computed a loss for each variable and then average the losses. RTFM only worked when we added a cross attention between the variables. We suppose this to be related to the feature magnitude learning. We trained with a batch size of $2$ using Adam optimizer and a learning rate of $6\times10^{-4}$ for 100 epochs with a weight decay of $1\times10^{-3}$. We also added an instance dropout of $0.5$.

\section{Computational efficiency}
\label{sec:9.6}
In Table \ref{table:16}, we report the inference time with a fixed input of $6$ variables, $8$ days and $200\times200$ spatial resolution. STEALNet is based on a 3D CNN auto-encoder without a self-attentions which explains its efficiency but on the cost of accuracy. In the last row, we show the estimated time when we discard the classification head to detect extreme events and only use the model for anomaly detection.

\begin{table}[!h]
 \caption{Inference time in seconds for our model and other DL baselines.}\label{table:16}
 \centering
 \setlength\tabcolsep{6pt}
 \begin{tabular}{l *{2}l}
 \toprule
 Algorithm & GPU$^1$ (sec) & Parameters (M) \\
\midrule
SimpleNet & 0.156 \gray{$\pm0.003$} & 0.203  \\

STEALNet  & 0.003 \gray{$\pm0.000$} & 6.005  \\
UniAD     & 3.733 \gray{$\pm0.012$} & 3.674  \\

DeepMIL   & 0.193 \gray{$\pm0.008$} & 0.285  \\
ARNet     & 0.191 \gray{$\pm0.003$} & 0.285  \\
RTFM      & 0.257 \gray{$\pm0.001$} & 0.319  \\

Ours      & 0.132 \gray{$\pm0.000$} & 0.479  \\
Ours$^2$     & 0.122 \gray{$\pm0.000$} & 0.145  \\
\bottomrule
\multicolumn{3}{l}{\footnotesize{$^1$NVIDIA GeForce RTX 3090 GPU}}\\
\multicolumn{3}{l}{\footnotesize{$^2$Our model is used only for driver/anomaly detection}}
\end{tabular}
\end{table}

The training on the real-world data for EUR-11 took about $\sim21$ hours with a Swin Transformer model, $K=16$, and $4$ NVIDIA A$100$ GPUs. Table \ref{table:22} gives a rough estimation for training on the synthetic CERRA for $1$ epoch. SimpleNet was trained with a pretrained backbone. The training time includes some postprocessing to compute metrics on the training set. The time might also differ depending on the I/O during training and the number of available workers.

\begin{table}[!h]
 \caption{Training time on the synthetic CERRA for 1 epoch.}\label{table:22}
 \centering
 \setlength\tabcolsep{6pt}
 \begin{tabular}{l *{2}l}
 \toprule
 Algorithm & Time (min) & GPU \\
\midrule
SimpleNet & $\sim2$ & A100 \\

STEALNet  & $\sim1$ & A100 \\
UniAD     & $\sim11$ & 4 $\times$ A100 \\

DeepMIL   & $\sim13$ & A40 \\
ARNet     & $\sim13$ & A40 \\
RTFM      & $\sim20$ & A40 \\

Ours      & $\sim8$ & A40 \\
\bottomrule
\end{tabular}
\end{table}

\section{Implementation details}
\label{sec:9.7}

The training was done mainly on clusters with NVIDIA A100 80GB and NVIDIA A40 48GB GPUs. In the following, we highlight the main technical implementations and hyperparameters for training:

\textbf{Synthetic data}~
For the synthetic data we set the embedding dimension $K=16$. We use one layer Video Swin Transformer with \{depth=[2, 1], heads=[2, 2], window size=[[2, 4, 4], [6, 1, 1]]\}.

For the quantization layer, we use two sequential 3D CNN layers on each variable with: \{[kernel=(3, 3, 3), stride=(1, 1, 1), padding=(1, 1, 1)]\} and followed by a shared linear layer.

The classifier $g_{\psi}$ consists of the following layers: \{[3D CNN, kernel=(3, 3, 3), stride=(3, 1, 1), padding=(0, 1, 1)], [3D CNN, kernel=(2, 3, 3), stride=(2, 1, 1), padding=(0, 1, 1)]\}.
We set $\lambda_{(ent)}=\lambda_{(div)}=0.1$, $\lambda_{(anomaly)}=100$, and $\lambda_{(commit)}=3$. The models were trained with Adam optimizer \cite{kingma2014adam} for $100$ epochs with a batch size of $4$. We use a linear warm up of $2$ epochs and a cosine decay with an initial learning rate of $2\times10^{-3}$ and a weight decay of $3\times10^{-3}$.

\textbf{Reanalysis data}~
We set the embedding dimension $K=24$ for CERRA and CAS-11 datasets and $K=16$ for the rest of ERA5-Land datasets. We use one layer Video Swin Transformer with \{depth=[2, 1], heads=[2, 2], window size=[[2, 4, 4], [8, 1, 1]]\}.
The quantization layer is similar to the one for the synthetic data.
The classifier $g_{\psi}$ consists of the $3$ layers each has \{[D CNN, kernel=(2, 3, 3), stride=(2, 1, 1), padding=(0, 1, 1)\}.
We set $\lambda_{(ent)}=\lambda_{(div)}=0.1$, and $\lambda_{(anomaly)}=100$ by defaults and $\lambda_{(commit)}=1.0$ for CAS-11. For CERRA, we set $\lambda_{(ent)}=\lambda_{(div)}=0.01$. Due to the high resolution on the reanalysis data, we use gradient checkpoint during training. For CERRA reanalysis, we also cut the boundaries between low and high latitudes focusing on the central region. This results in a final grid with 512×832 cells.

To handle missing data and temporal gaps in the input reanalysis data, we first normalize the data using the pre-computed statistics and then replace the invalid pixels with zero values.

\section{Reanalysis data}
\label{sec:9.8}
\begin{figure}[!h]
  \centering
  \includegraphics[draft=\draft, width=.9\textwidth]{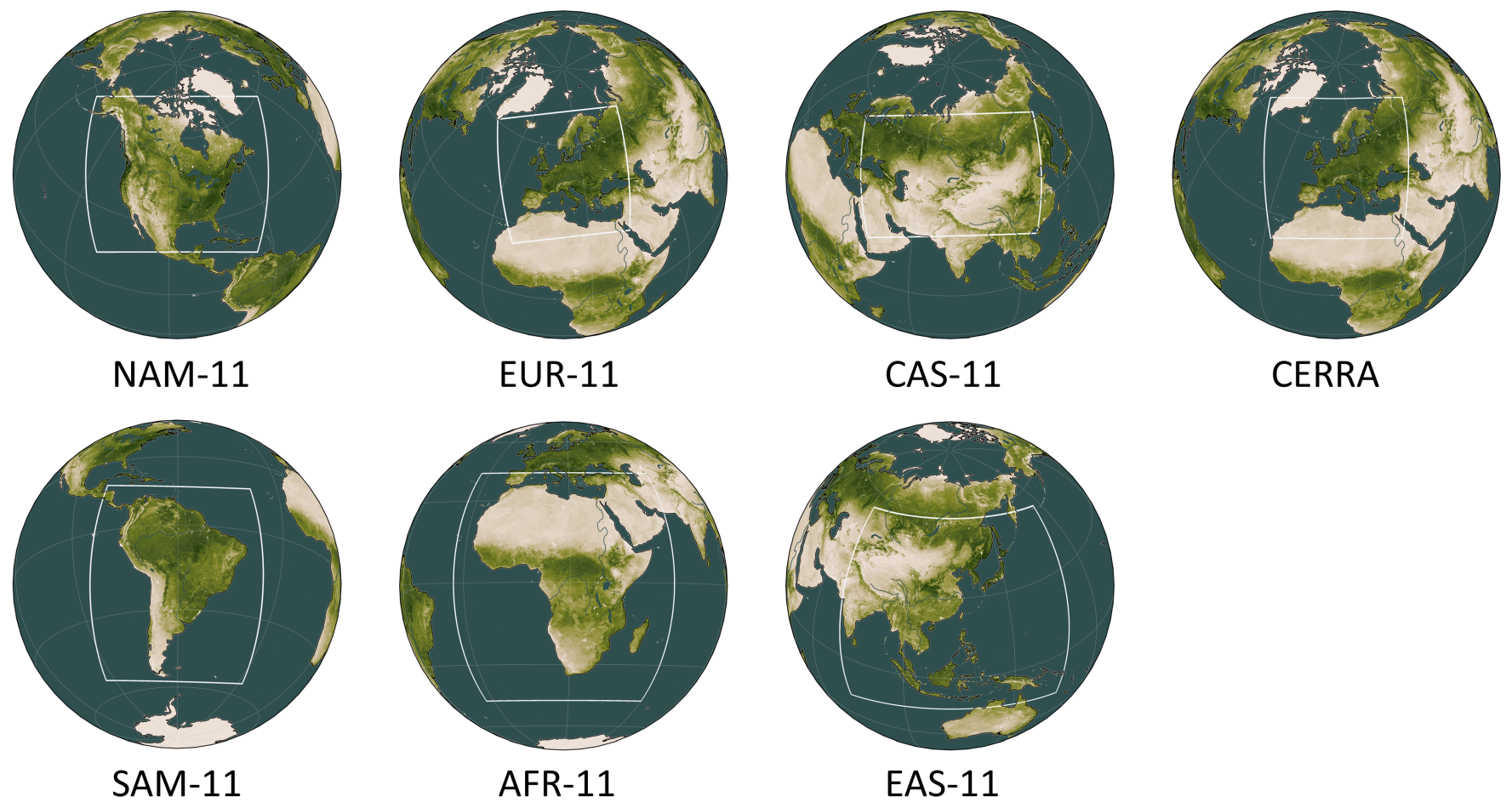}
  \caption{The definition of the domains used in the study. ERA5-Land reanalysis is mapped onto the CORDEX domains \cite{CORDEX}. CERRA has its own domain definition \cite{CERRA}.}\label{fig:12}
\end{figure}

The raw reanalysis data were provided by the Climate Data Store (CDS) \cite{ERA5-Land, CERRA}. Technical details regarding the reanalysis datasets are provided in Tables \ref{table:17} and \ref{table:18}. CORDEX regions \cite{CORDEX} used in this study are shown in Fig.~\ref{fig:12}.
For all regions on ERA5-Land, we used the following variables: \{"t2m", "fal", "e", "tp", "swvl1", "stl1"\}.
For CERRA, we did the experiments with: \{"t2m", "al", "tcc", "tp", "vsw", "r2"\}.

\begin{table}[h!]
  \caption{Datasets used in the experiments on real-world data. CORDEX domains are defined based on \cite{CORDEX}.}
  \label{table:17}
  \centering
  \tabcolsep=3pt\relax
  \begin{tabular}{l *{7}l}
    \toprule
    Dataset       & Region & CORDEX  & Resolution & Train     & Val       & Test \\
    \midrule
    CERRA     & Europe         & -      & 1069$\times$1069   & 1984-2015 & 2016-2018 & 2019-2021 \\
    \midrule
    ERA5-Land  & Europe         & EUR-11 & 412$\times$424     & 1981-2017 & 2018-2020 & 2021-2024 \\
    ERA5-Land  & Africa         & AFR-11 & 804$\times$776     & 1981-2017 & 2018-2020 & 2021-2024 \\
    ERA5-Land  & North America  & NAM-11 & 520$\times$620     & 1981-2017 & 2018-2020 & 2021-2024 \\
    ERA5-Land  & South America  & SAM-11 & 668$\times$584     & 1981-2017 & 2018-2020 & 2021-2024 \\
    ERA5-Land  & Central Asia   & CAS-11 & 400$\times$612     & 1981-2017 & 2018-2020 & 2021-2024 \\
    ERA5-Land  & East Asia      & EAS-11 & 668$\times$812     & 1981-2017 & 2018-2020 & 2021-2024 \\
    \bottomrule
  \end{tabular}
\end{table}

\begin{table}[h!]
\small
 \caption{Details regarding the processed variables from ERA5-Land \cite{ERA5-Land} and CERRA \cite{CERRA} reanalysis.}\label{table:18}
 \centering
 \setlength\tabcolsep{3pt}
 \begin{tabular}{*{5}l}
    \toprule
        Dataset & Variable name & Long name & Unit & Height \\
    \midrule
        CERRA & al     & albedo & \% & surface \\
              & hcc    & high cloud cover & \% & above 5000m \\
              & lcc    & low cloud cover & \% & surface-2500m \\
              & mcc    & medium cloud cover & \% & 2500m-5000m \\
              & liqvsm & liquid volumetric soil moisture & m$^{3}$/m$^{3}$ & top layer of soil \\
              & msl    & mean sea level pressure & Pa & surface \\
              & r2     & 2 metre relative humidity & \% & 2m \\
              & si10   & 10 metre wind speed & m/s & 10m\\
              & skt    & skin temperature & K & surface \\
              & sot    & soil temperature & K & top layer of soil \\
              & sp     & surface pressure & Pa & surface \\
              & sr     & surface roughness & m & surface \\
              & t2m    & 2 metre temperature & K & 2m \\
              & tcc    & total Cloud Cover & \% & above ground \\
              & tciwv  & total column integrated water vapour & kg/m$^{2}$ & surface \\
              & tp     & total Precipitation & kg/m$^{2}$ & surface \\
              & vsw    & volumetric soil moisture & m$^{3}$/m$^{3}$ & top layer of soil\\
              & wdir10 & 10 metre wind direction & $\circ$ & 10m \\
    \midrule
        ERA5-Land & d2m   & 2m dewpoint temperature & K & 2m \\
                  & t2m   & 2m temperature & K & 2m \\
                  & fal   & forecast albedo & \% & surface \\
                  & skt   & skin temperature & K & surface \\
                  & stl1  & soil temperature & K &  soil layer (0 - 7 cm) \\
                  & sp    & surface pressure & Pa & surface \\
                  & e     & total evaporation & m of water & above ground\\
                  &      &  & equivalent & \\

                  & tp    & total precipitation & m & surface \\
                  & swvl1 & volumetric soil water & m$^{3}$/m$^{3}$ & soil layer (0 - 7 cm) \\
\bottomrule
\end{tabular}
\end{table}

\section{Agricultural drought definition and remote sensing data}
\label{sec:9.9}
\subsection{Satellite-derived agricultural drought}
\label{sec:9.9.1}
It is generally challenging to define what exactly constitutes an extreme.
Extremes can be categorized from the perspective of their impacts. For instance, extreme drought can be categorized into 4 types based on their impacts \cite{liu2016agricultural, Hao_2018}; meteorological or climatological drought, agricultural drought \cite{meza_2020}, hydrological drought, and socioeconomic drought \cite{pedro2015review}. Meteorological drought is mainly related to the dryness and can be defined based on a deficiency in temperature or precipitation. Agricultural drought measures the impact of stress on vegetation and usually defined as soil water deficits. It is also widely conceived that drought originates and progresses from meteorological into agricultural drought \cite{HUANG201545, ZHANG2017141}. Hydrological drought is related to the water storage. While socioeconomic drought can be measured based on supply and demand related to weather and deficit in water supply.
In this paper, we are interested in extreme agricultural droughts.

\begin{figure}[!h]
  \centering
  \includegraphics[draft=\draft, width=.9\textwidth]{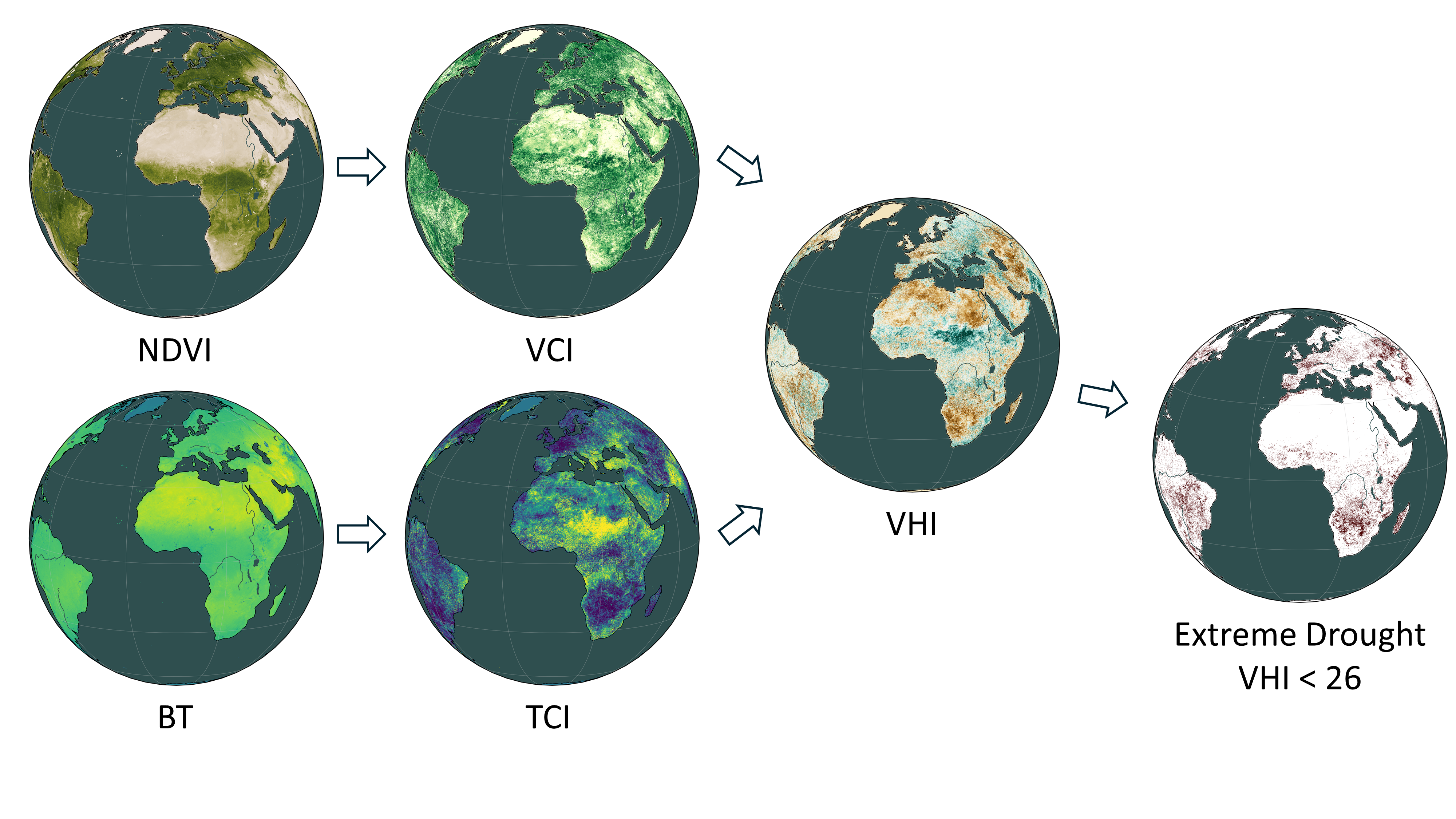}
  \caption{An overview of the extreme agricultural droughts definition from remote sensing.}\label{fig:13}
\end{figure}

Worldwide satellite observations allow for almost real-time monitoring of drought and vegetation conditions. In practice, vegetation states can be estimated from land surface reflectances acquired from satellites. As a result, the reflectances on the ground can be employed as agricultural drought indicators and as proxies for vegetation health.
To define extreme agricultural drought events, we processed satellite-based vegetation health dataset \citep{Blended} from the National Oceanic and Atmospheric Administration (NOAA) (\url{https://www.star.nesdis.noaa.gov/smcd/emb/vci/VH/index.php} [last access: 22 May 2024)]).
This dataset consists of long-term remote sensing data acquired from a system of NOAA satellites: the Advanced Very-High-Resolution Radiometer (AVHRR) which starts from 1981 until 2012 and the new system the Visible Infrared Imaging Radiometer Suite (VIIRS) from 2013 onwards.
The dataset has a global coverage with $\sim0.05^\circ$ ($\sim$4km) spatial resolution.
The normalized difference vegetation index (NDVI) \cite{NDVI} and brightness temperature (BT) \cite{Kogan1995} are the two key products of the dataset. The BT is an infrared (IR)-based calibrated spectrum radiation. While the NDVI is a combination of the near-infrared (NIR) and red (R) bands.
To remove the effects of clouds, atmospheric disturbance, and other error sources, the data were aggregated temporally into a smoothed product on a weekly basis. 
The weekly temporal coverage is needed for outliers and discontinuities removal and is suitable to study the phenological phases of vegetation and consequently to define agricultural drought \citep{Blended, Kogan_2011}.
Based on the long-term upper and lower bounds of the ecosystem (maximum and minimum values of the NDVI and BT), agricultural drought indicators such as vegetation condition index (VCI), thermal condition index (TCI), and vegetation health index (VHI) can be derived \citep{Kogan1995, Kogan1990}.
VHI is a combination of VCI and TCI (Fig.~\ref{fig:13}) and it fluctuates between 0 (unfavourable condition) and 100 (favourable condition). Values outside the range are clipped. Based on this definition of vegetation health, extreme agricultural drought can be defined when VHI < 26.
Please note that vegetation stress detected by VHI could not be necessarily caused by a drought event i.e., a change in the land cover can change the signal as well \cite{nhess-12-3519-2012, van2016drought}. Thus, VHI should be interpreted carefully.

\subsection{Pre-processing of the remote sensing data}
\label{sec:9.9.2}

The remote sensing dataset is provided on the Plate Carr\'{e}e projection (geographic latitude and longitude). The target agricultural drought data and reanalysis data have to be aligned in the same coordinate systems and over the same regions. To realize this, we mapped the remote sensing data onto the Lambert conformal conical grid for CERRA and onto the rotated coordinate systems over the different CORDEX domains for ERA5-Land. For the mapping, we use the first-order conservative mapping using the software from Zhuang et.~al \cite{xESMF}. To calculate the spatial averaged, we excluded coastal lines, invalid, and water body pixels. Furthermore, we combined the dataset with masks obtained from the quality assurance metadata for pixels over no vegetation and very cold areas. 
As mentioned in Sec.~\ref{sec:9.9.1}, a temporal decomposition was conducted to remove some discontinuity and aggregate the data into a weekly product. However, some pixels will still be empty. To tackle this issue, we first checked if the pixel was covered by another satellite and averaged the measurements of the satellites. If it was not the case, we flagged the pixel as invalid and discard it from the training and evaluation. 
This remote sensing dataset serves as a reference of extreme agricultural drought events to train and evaluate the performance of the model. Table ~\ref{table:19} shows the ratio of extreme events in the datasets. Please note that there is no ground truth for drivers or anomalies in our real-world dataset. We only report the ratio of extreme agricultural drought events, which can be detected using remote sensing data.

\begin{table}[h!]
 \caption{Details regarding the ratio of extreme events in the pre-processed NOAA remote sensing data.}\label{table:19}
 \centering
 \begin{tabular}{*{4}l}
    \toprule
    Region & Domain & \multicolumn{2}{c}{Extremes (\%)} \\
    \cmidrule{3-4}
        & & Val & Test \\
    \midrule
    Europe & CERRA & 4.34 & 5.32 \\
    \midrule
    Europe & EUR-11 & 3.20 & 2.86 \\
    Africa & AFR-11 & 6.41 & 6.87 \\
    North America & NAM-11 & 3.68 & 6.61 \\
    South America & SAM-11 & 5.16 & 6.53 \\
    Central Asia & CAS-11 & 3.60 & 4.38 \\
    East Asia & EAS-11 & 3.16 & 3.05 \\
\bottomrule
\end{tabular}
\end{table}

\clearpage

\section{Additional results}
\label{sec:9.10}


\begin{figure}[!h]
  \centering
  \includegraphics[width=.75\textwidth]{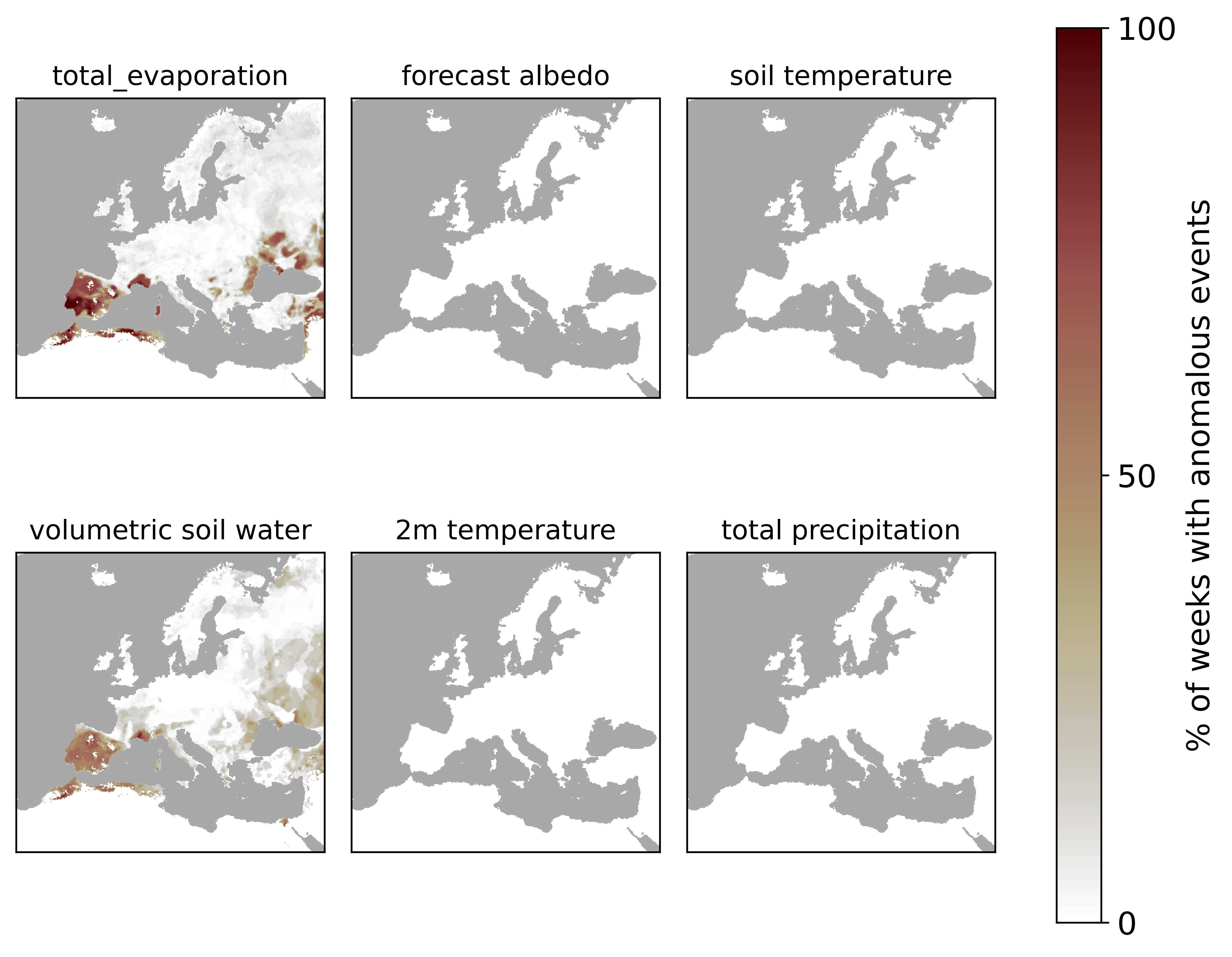}
  \caption{The averaged spatial distribution of drivers and anomalies related to Portugal in Europe. For this experiment, we use prediction on EUR-11 from ERA5-Land and select frames (times) within the period 2018-2024 where there were extreme drought of at least 25\% of the pixels in the Portugal. Then we normalize the identified drivers and anomalies by the total number of frames to obtain the final map. As can be seen drivers are spatially centered around where extremes were reported.}\label{fig:14}
\end{figure}

\begin{figure}[!h]
  \centering
  \includegraphics[width=.75\textwidth]{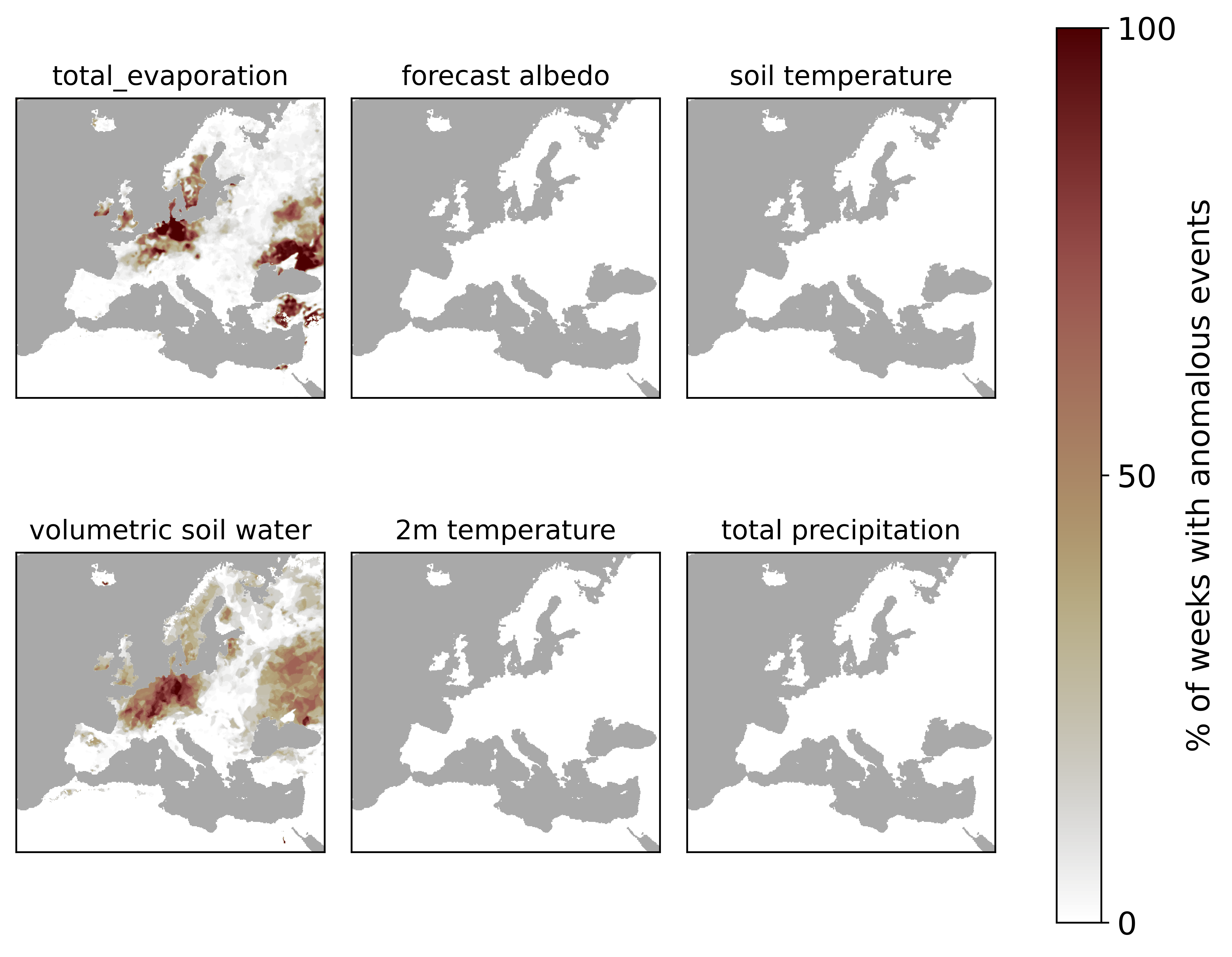}
  \caption{The averaged spatial distribution of drivers and anomalies related to a specific place in Europe (North Rhine-Westphalia). For this experiment, we use prediction on EUR-11 from ERA5-Land and select frames (times) within the period 2018-2024 where there were extreme drought of at least 25\% of the pixels in the North Rhine-Westphalia. Then we normalize the identified drivers and anomalies by the total number of frames to obtain the final map. As can be seen drivers are spatially centered around where extremes were reported.}\label{fig:15}
\end{figure}


\begin{figure}[!h]
  \centering
  \includegraphics[width=.99\textwidth]{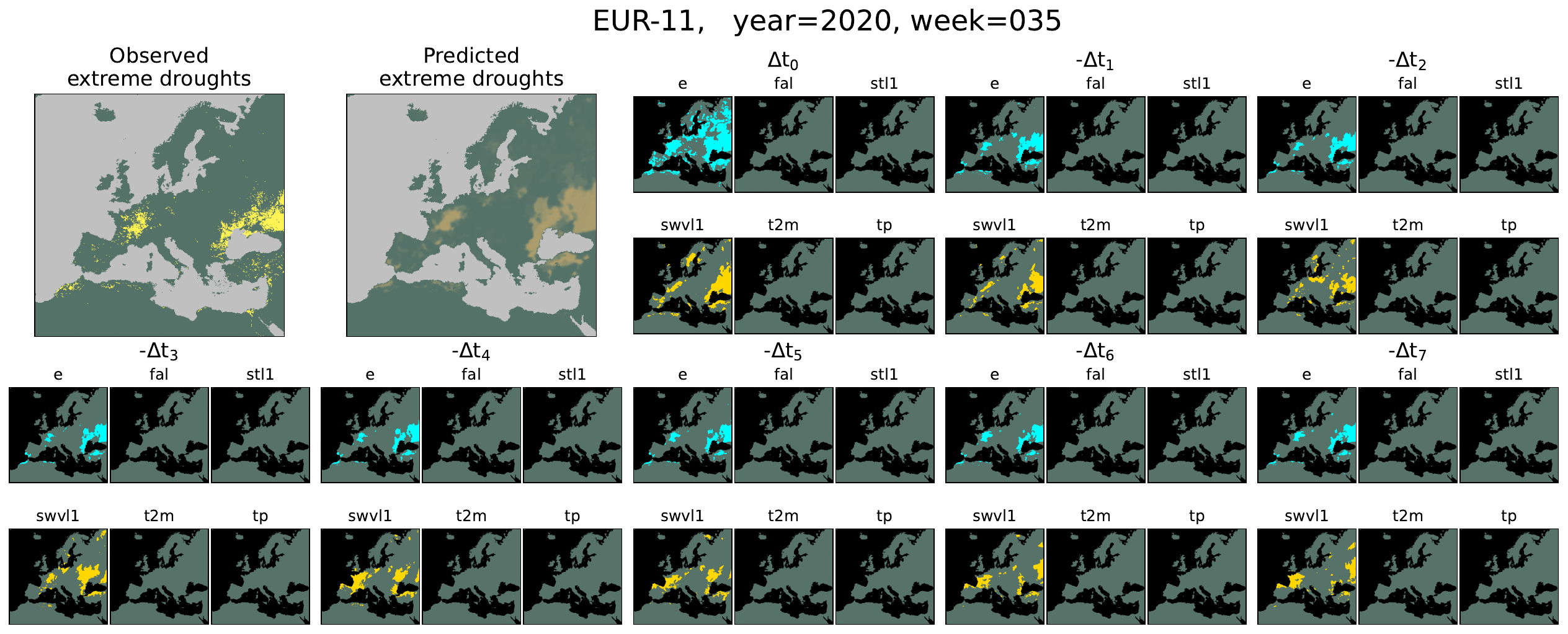}
  \caption{Qualitative results on ERA5-Land for Europe (EUR-11). Shown are the identified drivers and anomalies for each variable along with the prediction of extreme agricultural droughts on the top left.}\label{fig:16}
\end{figure}

\begin{figure}[!h]
  \centering
  \includegraphics[width=.99\textwidth]{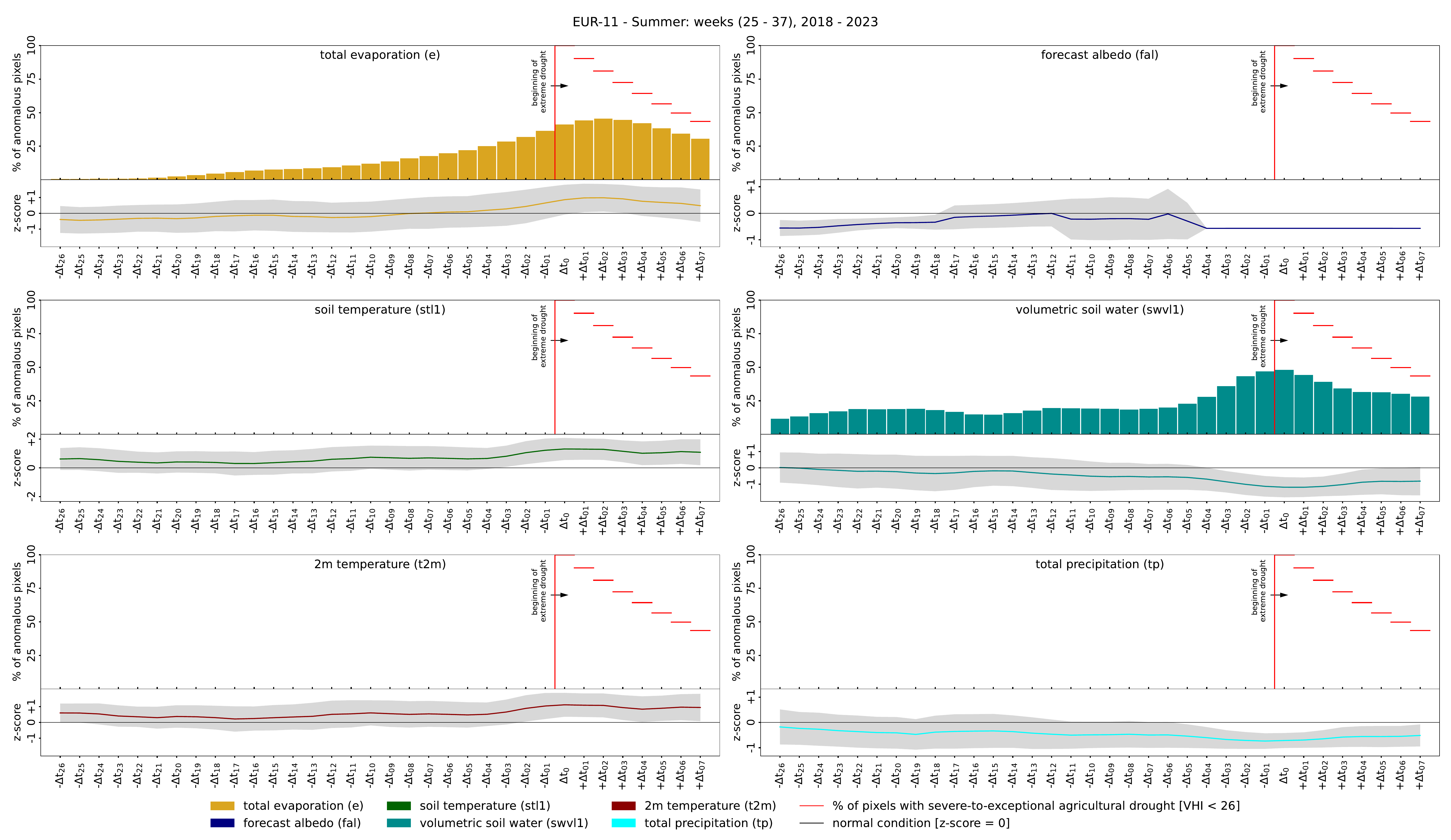}
  \caption{Temporal evolution of drivers and anomalies related to the extremes in ERA5-Land for Europe (EUR-11). For this experiments, we select pixels with extreme events during summer (weeks $25$-$38$) for the years 2018-2023 and compute the average distribution of drivers and anomalies with time. The red line at $\delta t_0$ indicates the beginning of the extreme droughts. $Z_{score}$ in the underneath curve represents the deviation from the mean computed from the ERA5-Land climatology.}\label{fig:17}
\end{figure}

\clearpage


\begin{figure}[!h]
  \centering
  \includegraphics[width=.99\textwidth]{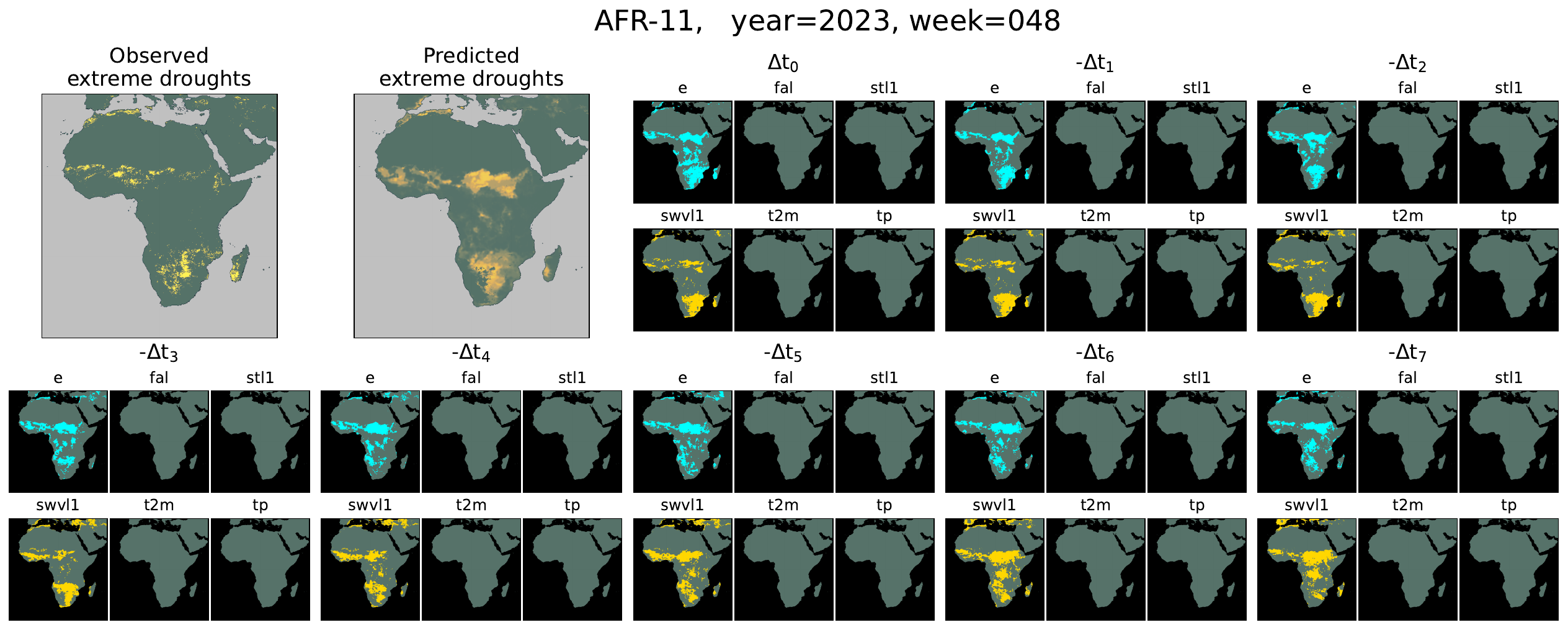}
  \caption{Qualitative results on ERA5-Land for Africa (AFR-11). Shown are the identified drivers and anomalies for each variable along with the prediction of extreme agricultural droughts on the top left.}\label{fig:18}
\end{figure}

\begin{figure}[!h]
  \centering
  \includegraphics[width=.99\textwidth]{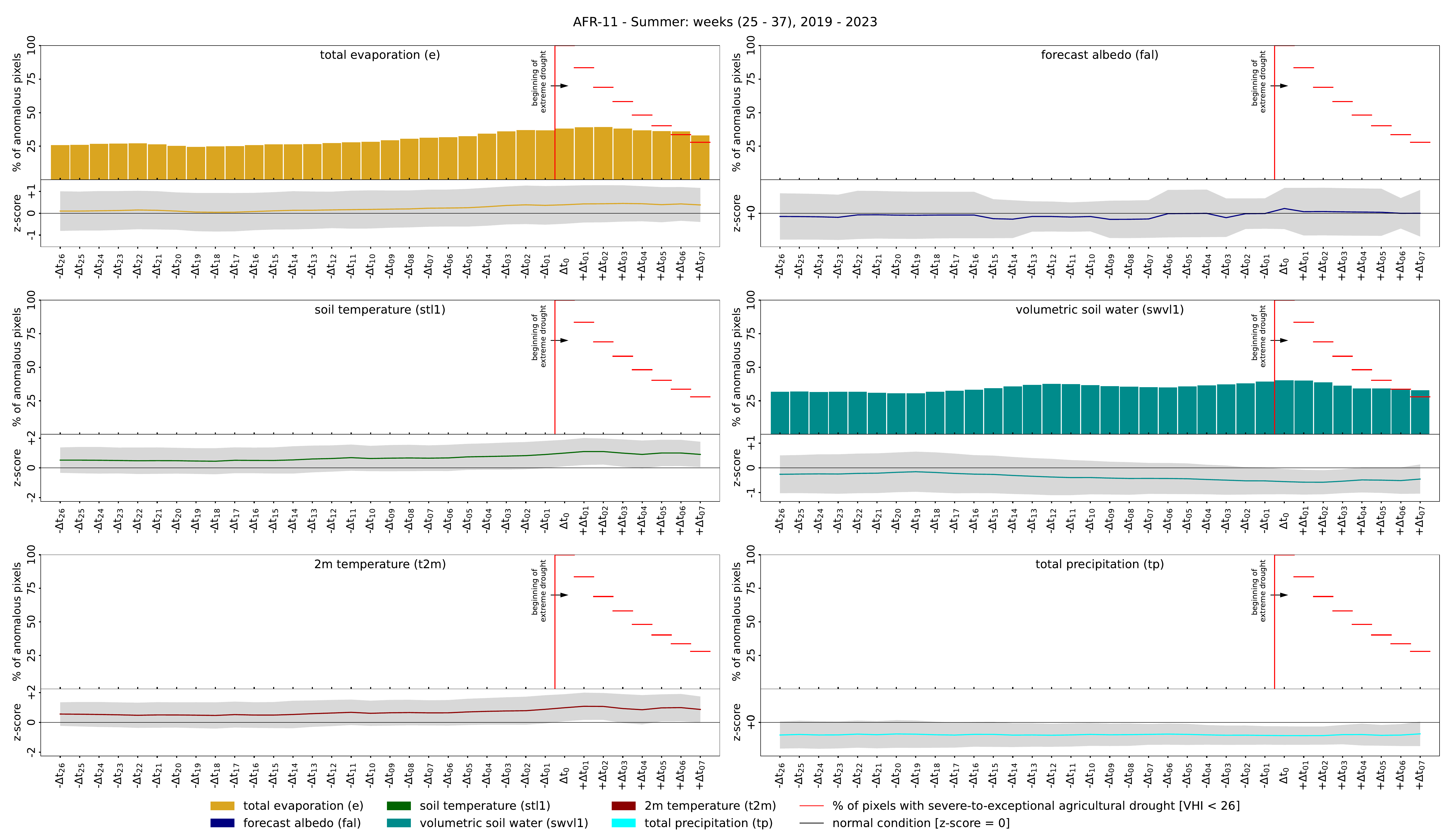}
  \caption{Temporal evolution of drivers and anomalies related to the extremes in ERA5-Land for Africa (AFR-11). For this experiments, we select pixels with extreme events during summer (weeks $25$-$38$) for the years 2019-2023 and compute the average distribution of drivers and anomalies with time. The red line at $\delta t_0$ indicates the beginning of the extreme droughts. $Z_{score}$ in the underneath curve represents the deviation from the mean computed from the ERA5-Land climatology.}\label{fig:19}
\end{figure}

\clearpage


\begin{figure}[!h]
  \centering
  \includegraphics[width=.99\textwidth]{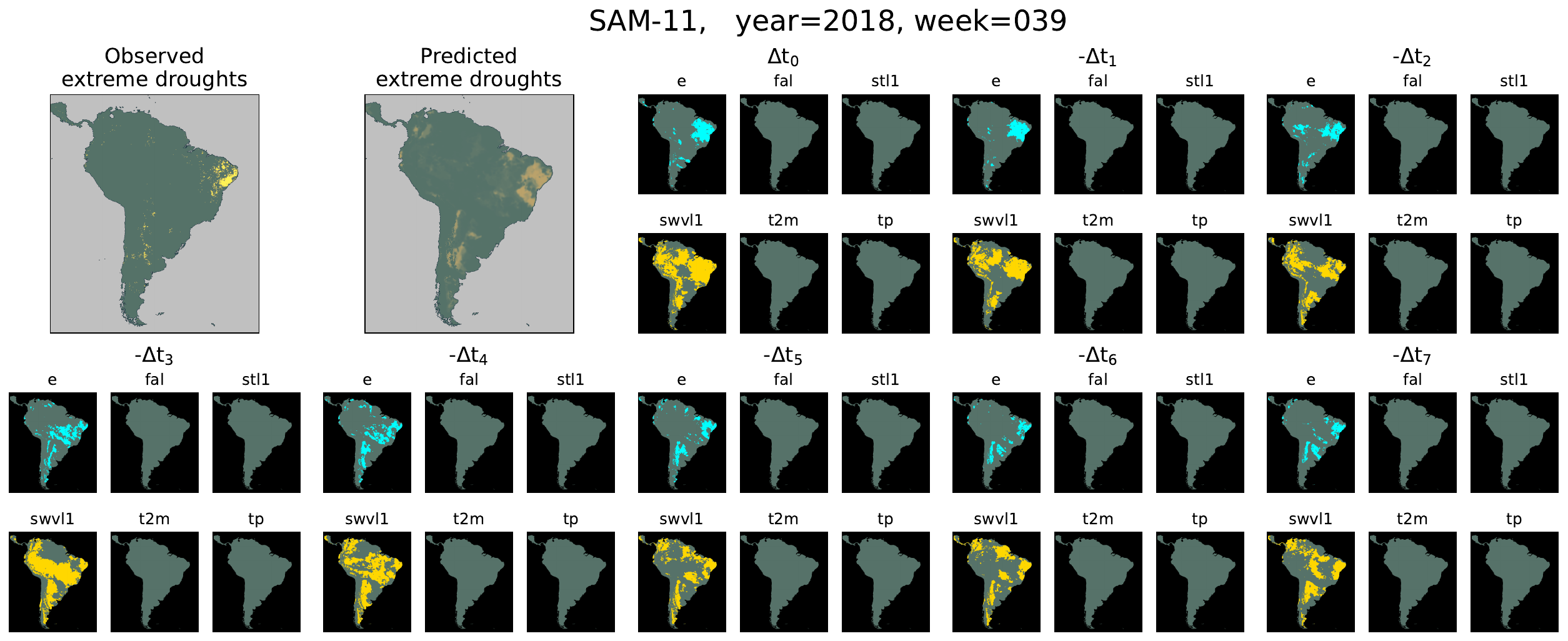}
  \caption{Qualitative results on ERA5-Land for South America (SAM-11). Shown are the identified drivers and anomalies for each variable along with the prediction of extreme agricultural droughts on the top left.}\label{fig:20}
\end{figure}

\begin{figure}[!h]
  \centering
  \includegraphics[width=.99\textwidth]{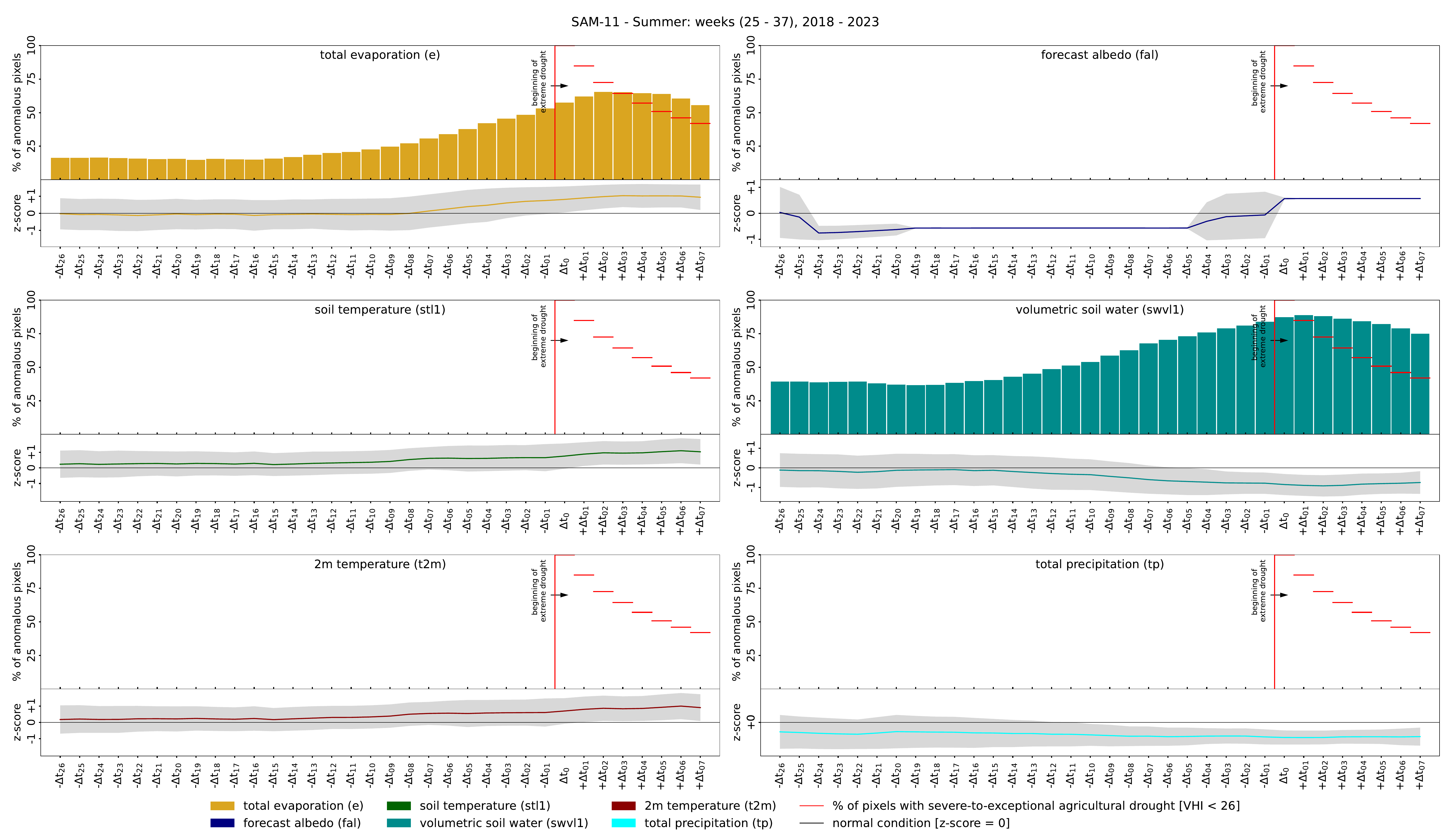}
  \caption{Temporal evolution of drivers and anomalies related to the extremes in ERA5-Land for South America (SAM-11). For this experiments, we select pixels with extreme events during summer (weeks $25$-$38$) for the years 2018-2023 and compute the average distribution of drivers and anomalies with time. The red line at $\delta t_0$ indicates the beginning of the extreme droughts. $Z_{score}$ in the underneath curve represents the deviation from the mean computed from the ERA5-Land climatology.}\label{fig:21}
\end{figure}

\clearpage


\begin{figure}[!h]
  \centering
  \includegraphics[width=.99\textwidth]{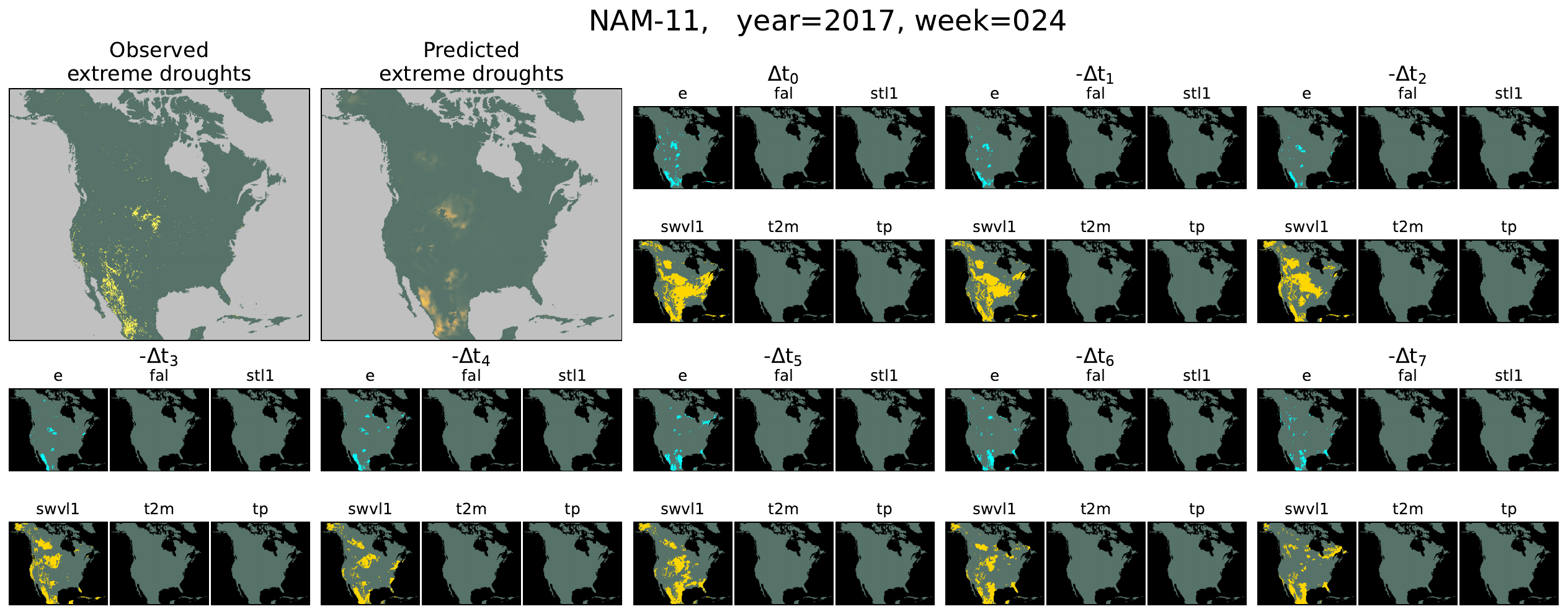}
  \caption{Qualitative results on ERA5-Land for North America (NAM-11). Shown are the identified drivers and anomalies for each variable along with the prediction of extreme agricultural droughts on the top left.}\label{fig:22}
\end{figure}

\begin{figure}[!h]
  \centering
  \includegraphics[width=.99\textwidth]{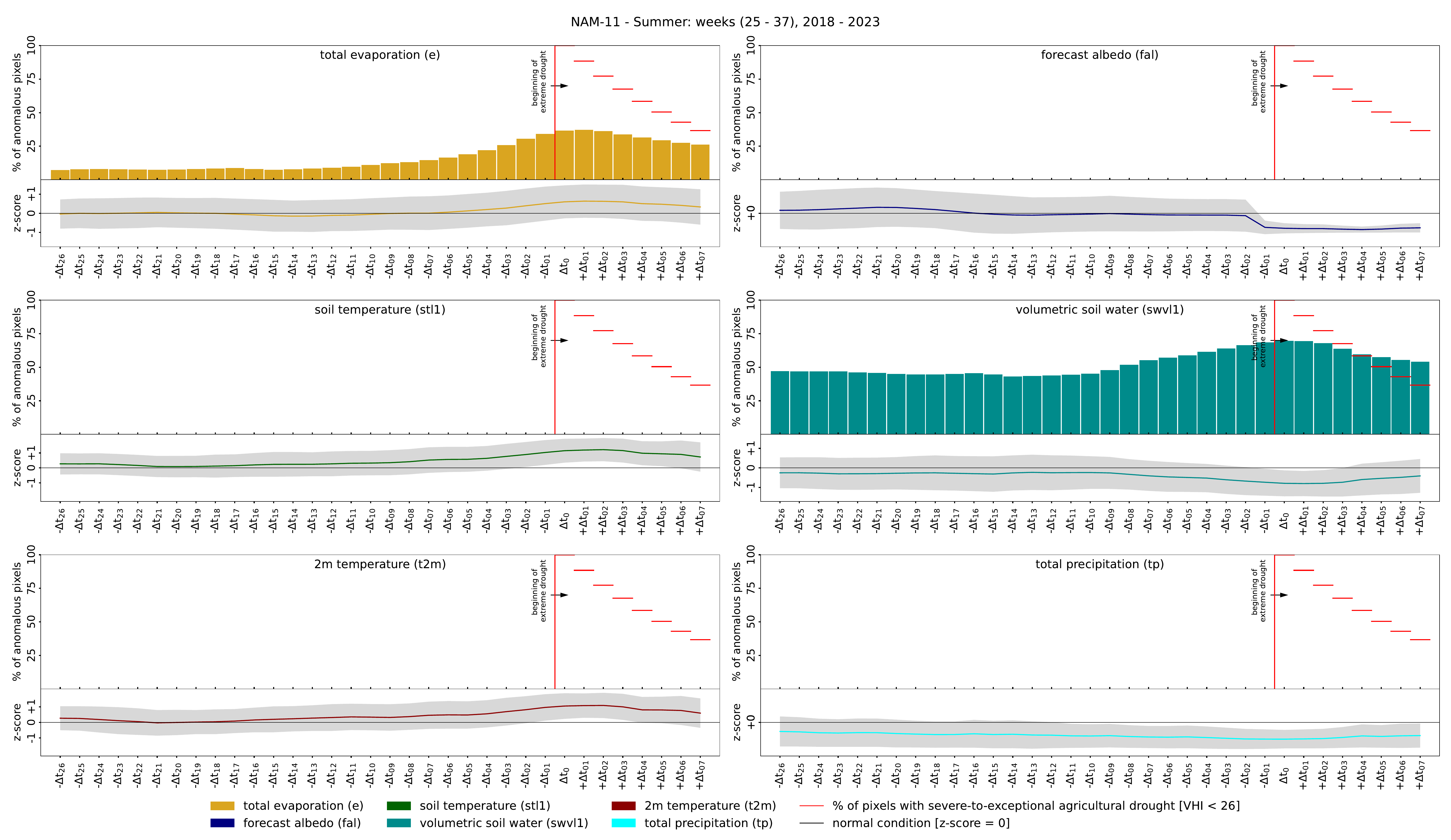}
  \caption{Temporal evolution of drivers and anomalies related to the extremes in ERA5-Land for North America (NAM-11). For this experiments, we select pixels with extreme events during summer (weeks $25$-$38$) for the years 2018-2023 and compute the average distribution of drivers and anomalies with time. The red line at $\delta t_0$ indicates the beginning of the extreme droughts. $Z_{score}$ in the underneath curve represents the deviation from the mean computed from the ERA5-Land climatology.}\label{fig:23}
\end{figure}

\clearpage


\begin{figure}[!h]
  \centering
  \includegraphics[width=.99\textwidth]{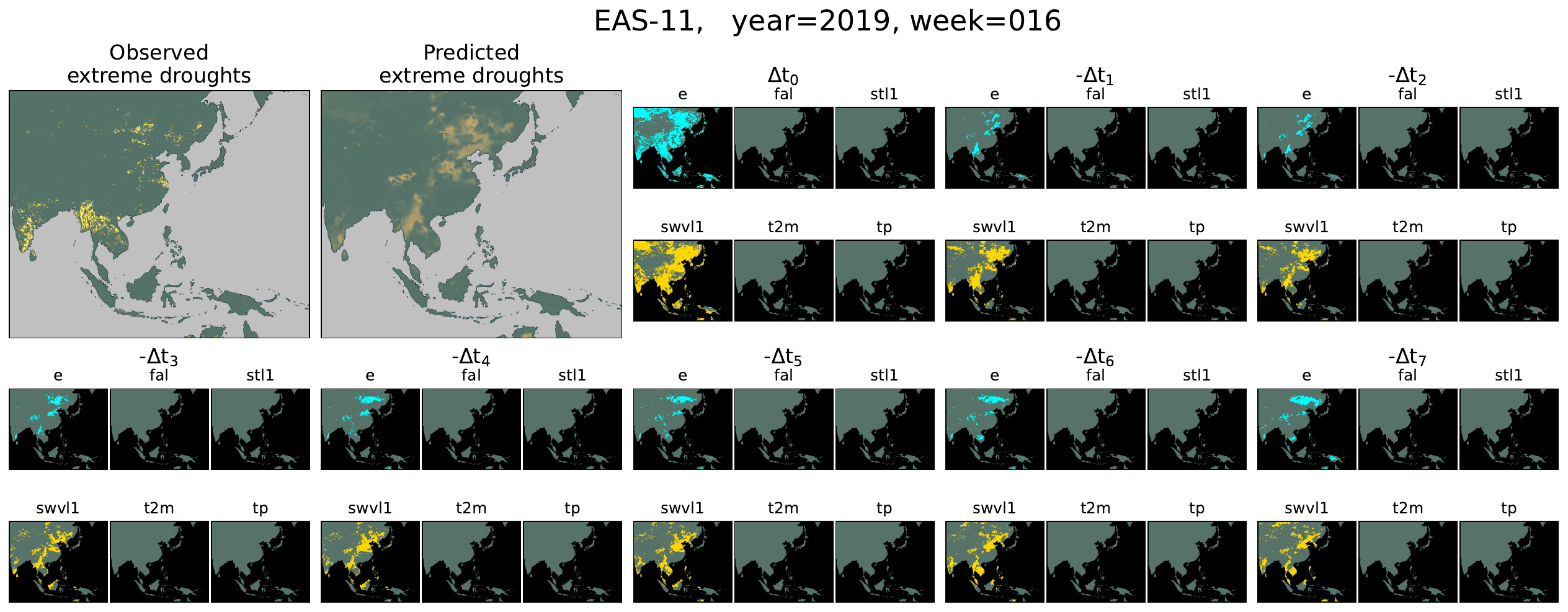}
  \caption{Qualitative results on ERA5-Land for East Asia (EAS-11). Shown are the identified drivers and anomalies for each variable along with the prediction of extreme agricultural droughts on the top left.}\label{fig:24}
\end{figure}

\begin{figure}[!h]
  \centering
  \includegraphics[width=.99\textwidth]{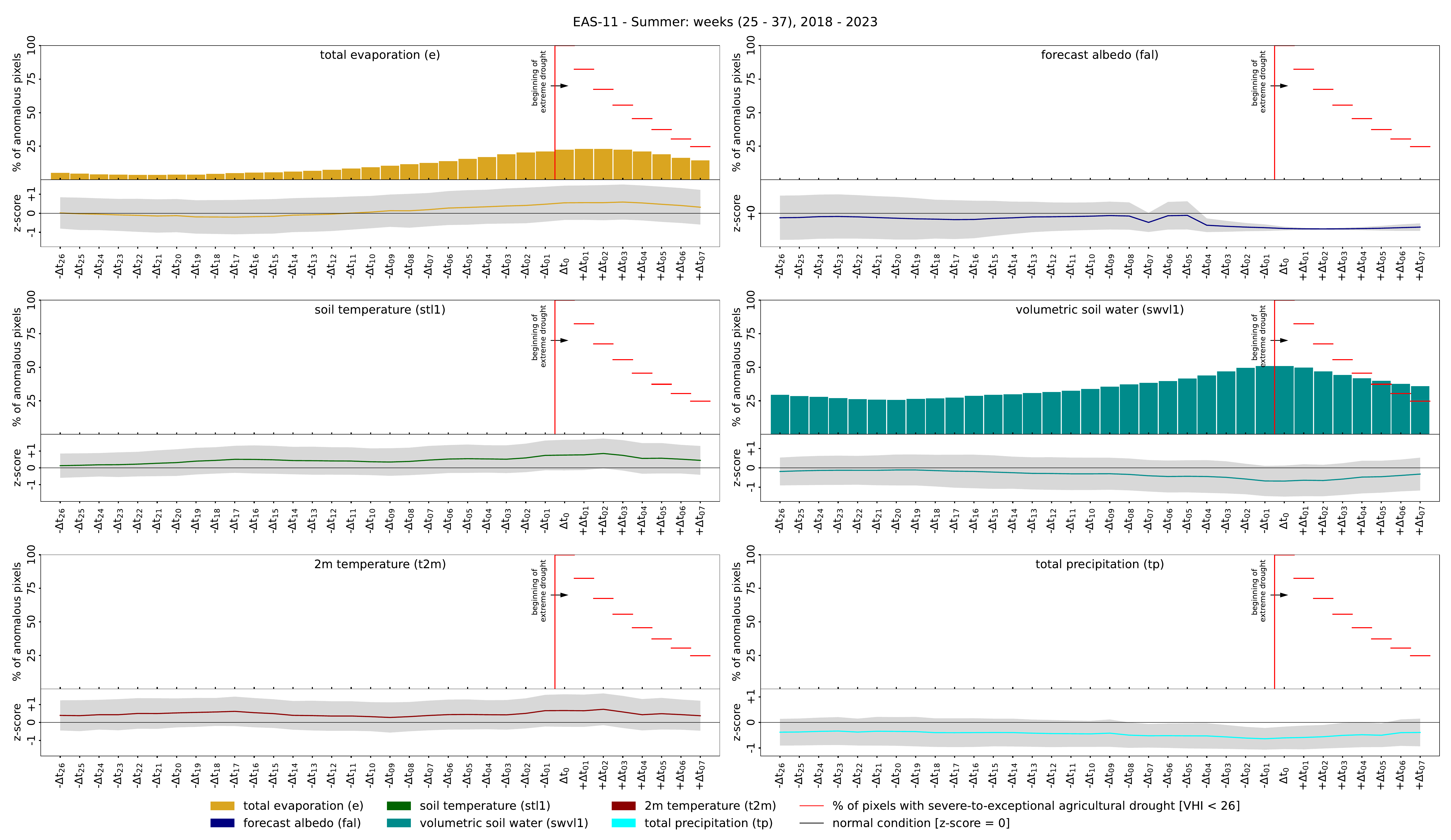}
  \caption{Temporal evolution of drivers and anomalies related to the extremes in ERA5-Land for East Asia (EAS-11). For this experiments, we select pixels with extreme events during summer (weeks $25$-$38$) for the years 2018-2023 and compute the average distribution of drivers and anomalies with time. The red line at $\delta t_0$ indicates the beginning of the extreme droughts. $Z_{score}$ in the underneath curve represents the deviation from the mean computed from the ERA5-Land climatology.}\label{fig:25}
\end{figure}

\clearpage


\begin{figure}[!h]
  \centering
  \includegraphics[width=.99\textwidth]{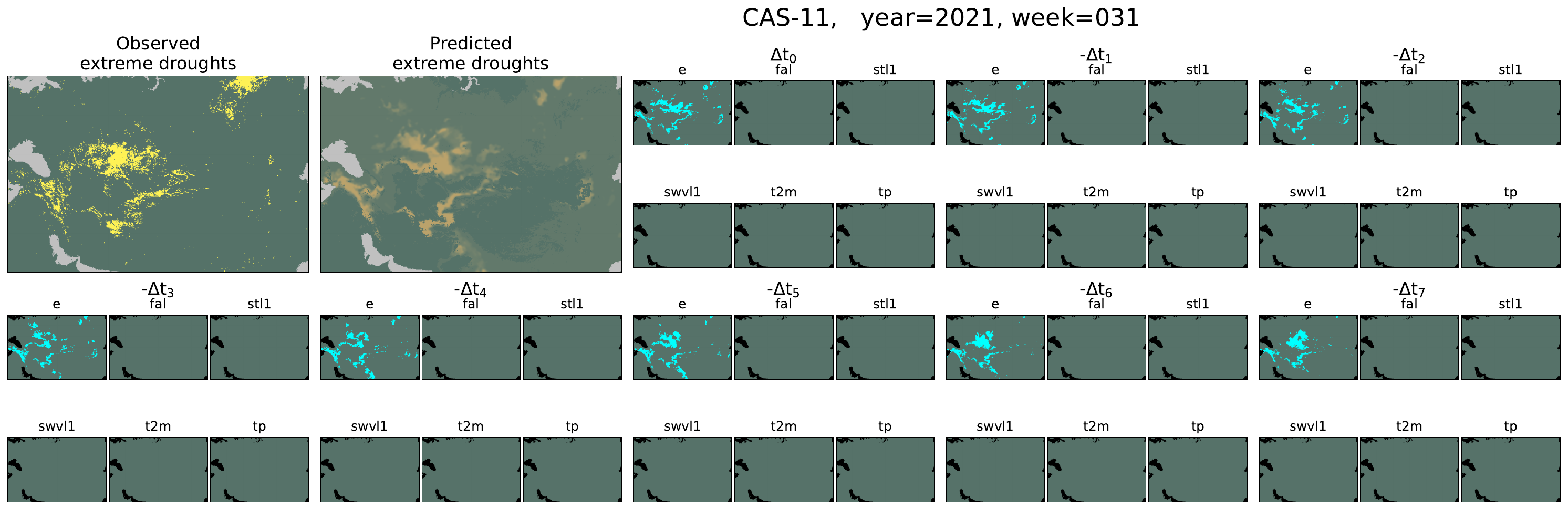}
  \caption{Qualitative results on ERA5-Land for Central Asia (CAS-11). Shown are the identified drivers and anomalies for each variable along with the prediction of extreme agricultural droughts on the top left.}\label{fig:26}
\end{figure}

\begin{figure}[!h]
  \centering
  \includegraphics[width=.99\textwidth]{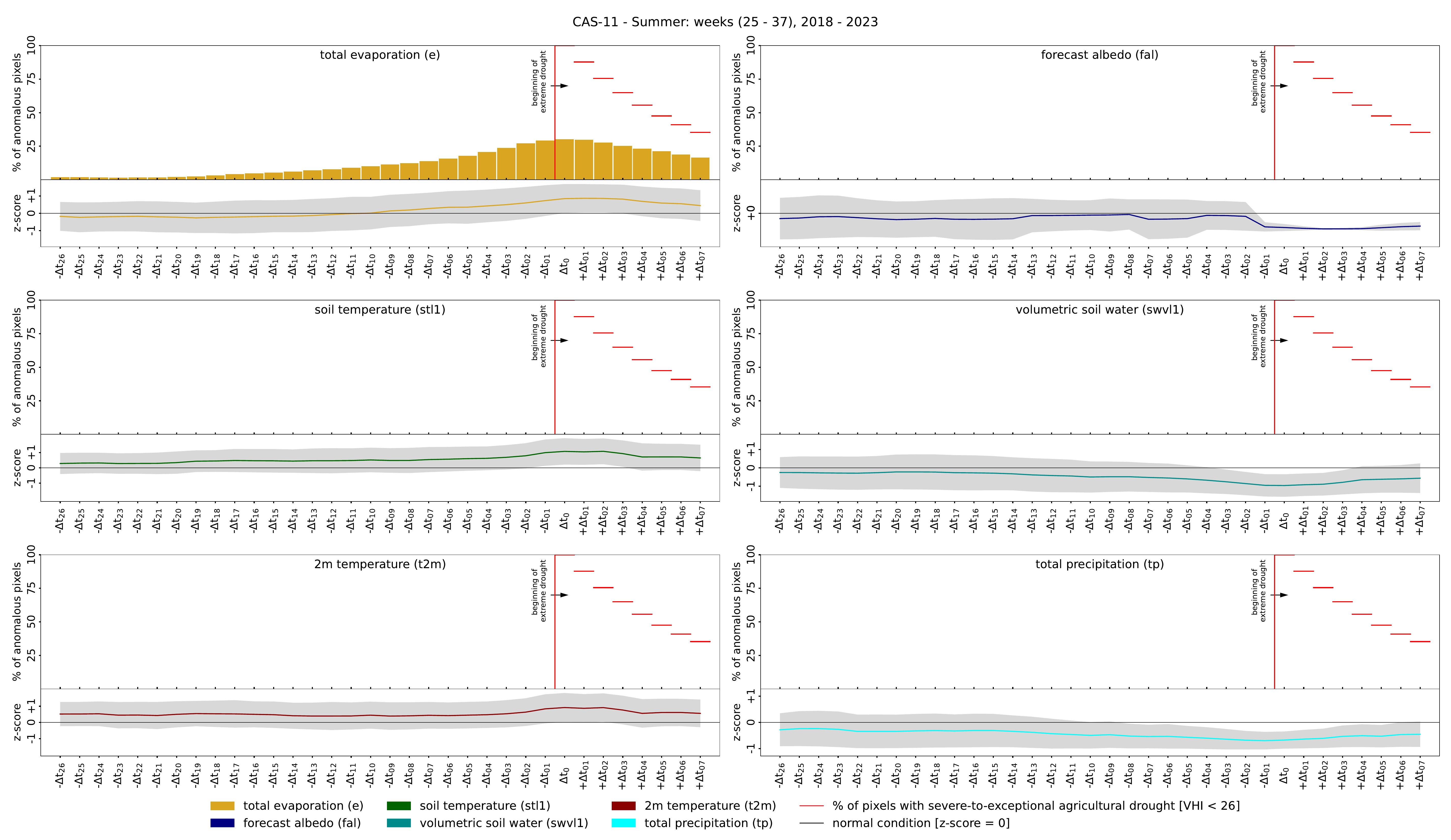}
  \caption{Temporal evolution of drivers and anomalies related to the extremes in ERA5-Land for Central Asia (CAS-11). For this experiments, we select pixels with extreme events during summer (weeks $25$-$38$) for the years 2018-2023 nd compute the average distribution of drivers and anomalies with time. The red line at $\delta t_0$ indicates the beginning of the extreme droughts. $Z_{score}$ in the underneath curve represents the deviation from the mean computed from the ERA5-Land climatology.}\label{fig:27}
\end{figure}

\clearpage


\begin{figure}[!h]
  \centering
  \includegraphics[width=.99\textwidth]{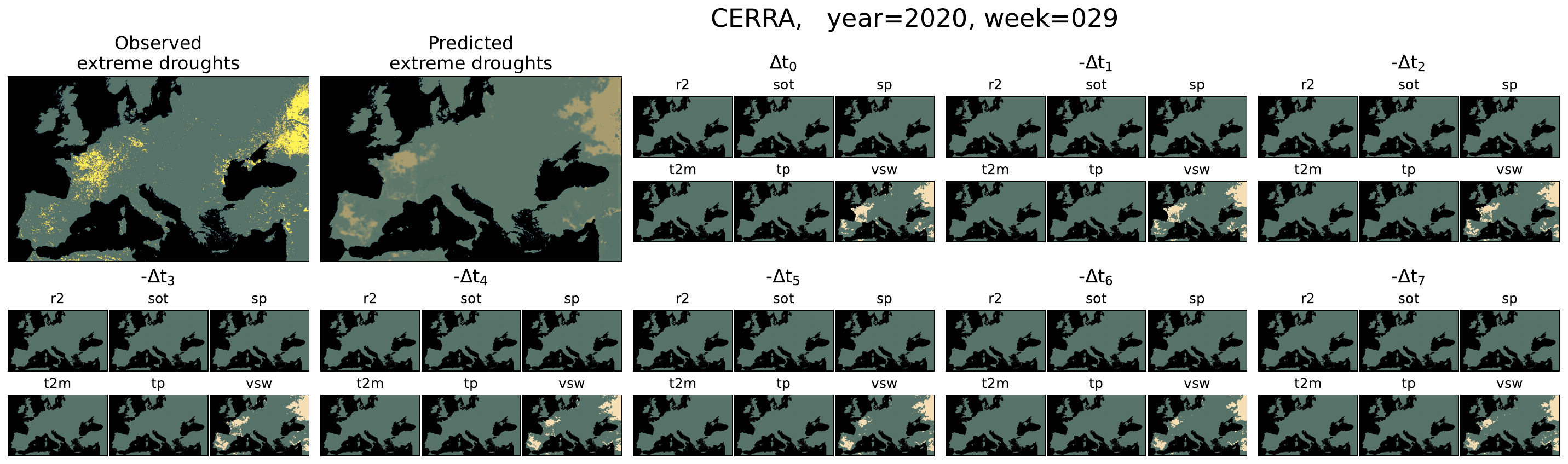}
  \caption{Qualitative results on CERRA reanalysis for Europe. Shown are the identified drivers and anomalies for each variable along with the prediction of extreme agricultural droughts on the top left.}\label{fig:28}
\end{figure}

\begin{figure}[!h]
  \centering
  \includegraphics[width=.99\textwidth]{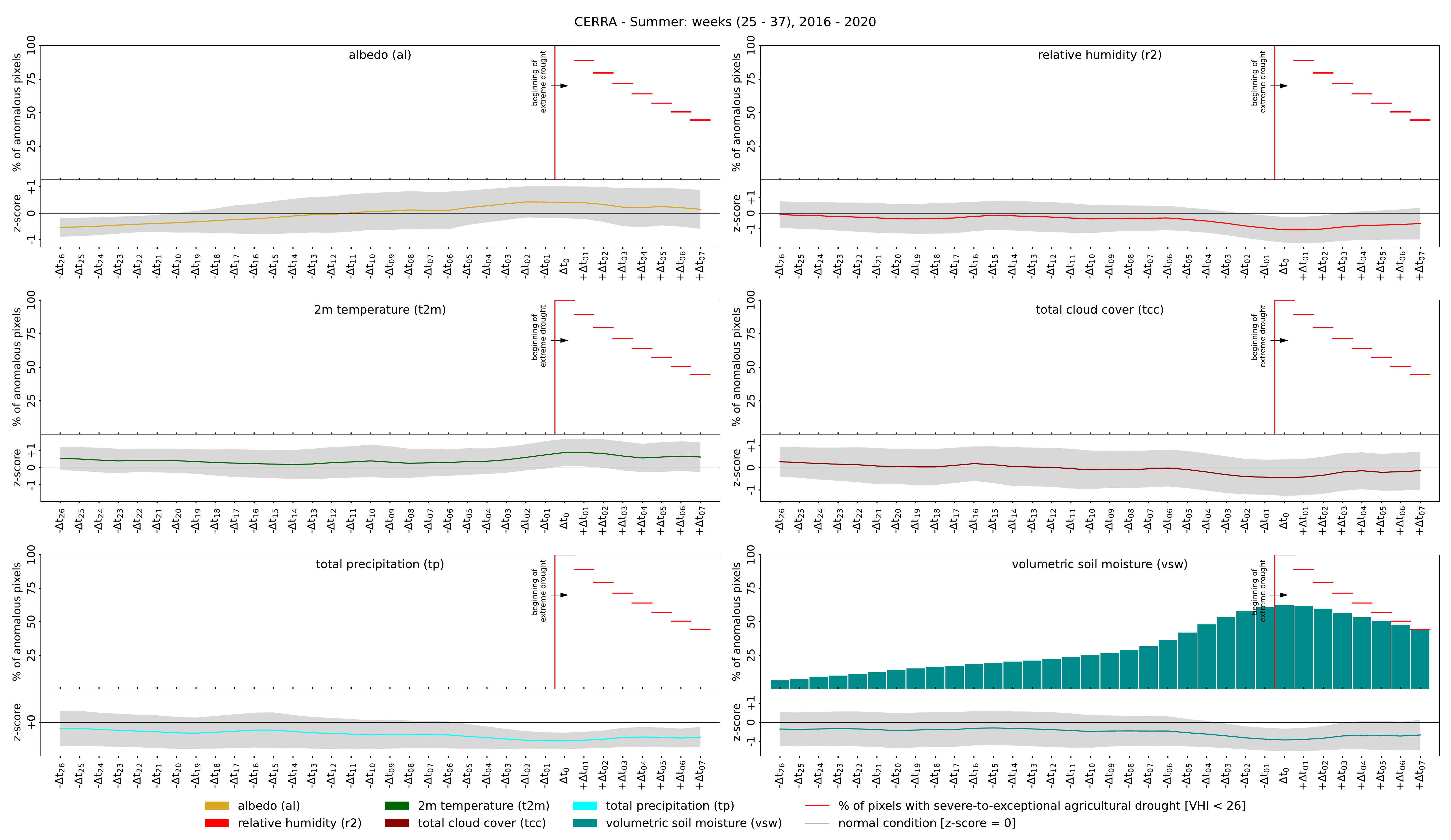}
  \caption{Temporal evolution of drivers and anomalies related to the extremes in CERRA reanalysis for Europe. For this experiments, we select pixels with extreme events during summer (weeks $25$-$38$) for the years 2016-2020 and compute the average distribution of drivers and anomalies with time. The red line at $\delta t_0$ indicates the beginning of the extreme droughts. $Z_{score}$ in the underneath curve represents the deviation from the mean computed from the CERRA climatology.}\label{fig:29}
\end{figure}

\section{Code and data availability}
\label{sec:9.12}
The source code to reproduce the results is available on GitHub at \url{https://github.com/HakamShams/IDEE}. The source code for the synthetic data generation is also available on GitHub at \url{https://github.com/HakamShams/Synthetic_Multivariate_Anomalies}. The pre-processed data used in this study are available at \url{https://doi.org/10.60507/FK2/RD9E33} \cite{shams_data}. 

\section{Broader impacts}
\label{sec:9.11}
There are generally no direct negative social impacts for conducting climate science researches.
However, anomaly detection algorithms in general could be adapted for video surveillance and might infringe privacy considerations. Although this is the risk of developing anomaly detection algorithms.

\end{document}